\documentclass[11pt]{article}

\usepackage[margin=1in]{geometry}
\usepackage{natbib}

\usepackage[utf8]{inputenc} 
\usepackage[T1]{fontenc}    
\usepackage{hyperref}       
\hypersetup{hidelinks}
\usepackage{url}            
\usepackage{booktabs}       
\usepackage{amsfonts}       
\usepackage{amsmath}
\usepackage{amssymb}
\usepackage{nicefrac}       
\usepackage{microtype}      
\usepackage{xcolor}         

\usepackage{graphicx}
\usepackage{multirow}
\usepackage{makecell}
\usepackage{colortbl}
\usepackage{subcaption}
\usepackage{pifont}
\usepackage{float}
\usepackage{placeins}       
\usepackage{enumitem}       
\usepackage{titletoc}       
\usepackage{setspace}       
\usepackage{fancyhdr}
\usepackage{fontawesome5}
\definecolor{HimmelBlue}{HTML}{1E88E5}
\definecolor{HimmelLightBlue}{HTML}{F6FBFF}
\newsavebox{\abstractcontentbox}
\renewenvironment{abstract}{%
  \par\begingroup
  \setlength{\parskip}{0pt}%
  \noindent\colorbox{HimmelBlue}{%
    \parbox{\dimexpr\linewidth-2\fboxsep\relax}{%
      \vspace{3pt}\hspace{5pt}{\color{white}\large\bfseries Abstract}\vspace{3pt}}}%
  \par\nointerlineskip
  \begin{lrbox}{\abstractcontentbox}%
    \begin{minipage}{\dimexpr\linewidth-16pt-2\fboxrule\relax}%
      \setlength{\parindent}{0pt}\noindent\ignorespaces
}{%
    \end{minipage}%
  \end{lrbox}%
  \noindent\setlength{\fboxsep}{8pt}%
  \fcolorbox{HimmelBlue}{HimmelLightBlue}{\usebox{\abstractcontentbox}}%
  \par\endgroup\vspace{0.7em}
}
\newcommand{\corrmark}{\textsuperscript{\ensuremath{\dagger}}}
\newcommand{\answerYes}[1][]{\textcolor{blue}{[Yes]#1}}
\newcommand{\answerNo}[1][]{\textcolor{orange}{[No]#1}}
\newcommand{\answerNA}[1][]{\textcolor{gray}{[N/A]#1}}
\newenvironment{ack}{\section*{Acknowledgments and Disclosures}}{}

\hypersetup{
  pdftitle={HIMMEL: Hierarchical Interleaved Multi-stream Motion Encoding for Long Video Understanding},
  pdfauthor={Haopeng Jin, Wenlong Zhao, Tao Yu, Tiankun Yang, Zhenyu Guan, ShiQuan Dong, Yujia Yang, Shani Ye, Chubin Chen, Hongzhu Yi, Jinwen Luo},
  pdfsubject={Revised technical report; a conference version was accepted at NeurIPS 2026}
}

\begin{document}
\thispagestyle{plain}
\noindent
\includegraphics[width=0.16\linewidth]{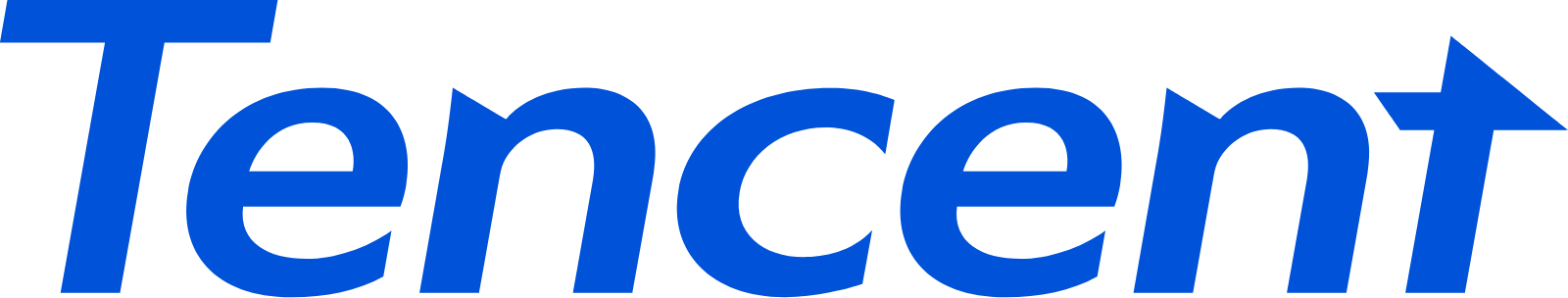}
\hfill
{\small\href{https://github.com/Sunbeam23333/HIMMEL}{\faGithub\enspace\texttt{github.com/Sunbeam23333/HIMMEL}}}
\par\vspace{0.35cm}
\begin{center}
\setlength{\parskip}{0pt}
{\LARGE\bfseries HIMMEL: Hierarchical Interleaved\\[0.2em]
Multi-stream Motion Encoding\\[0.2em]
for Long Video Understanding\par}

\vspace{0.45cm}
{\large
Haopeng Jin\textsuperscript{1} \quad
Wenlong Zhao\textsuperscript{1} \quad
Tao Yu\textsuperscript{2}\\[0.2em]
Tiankun Yang\textsuperscript{2} \quad
Zhenyu Guan\textsuperscript{2} \quad
ShiQuan Dong\textsuperscript{3} \quad
Yujia Yang\textsuperscript{2}\\[0.2em]
Shani Ye\textsuperscript{1} \quad
Chubin Chen\textsuperscript{4}\corrmark \quad
Hongzhu Yi\textsuperscript{2}\corrmark \quad
Jinwen Luo\textsuperscript{1}\corrmark\par}
\vspace{0.3cm}
{\small
\textsuperscript{1}Tencent \qquad
\textsuperscript{2}University of Chinese Academy of Sciences\\[0.15em]
\textsuperscript{3}Beijing Forestry University \qquad
\textsuperscript{4}Tsinghua University\par}
\vspace{0.25cm}
{\small
Haopeng Jin: \href{mailto:king.ham.roc@gmail.com}{king.ham.roc@gmail.com}\\[0.15em]
\corrmark~Corresponding authors: Chubin Chen: \href{mailto:trubeeen@gmail.com}{trubeeen@gmail.com}\\[0.15em]
Hongzhu Yi: \href{mailto:yihongzhu23@mails.ucas.ac.cn}{yihongzhu23@mails.ucas.ac.cn}
\quad Jinwen Luo: \href{mailto:jamsluo@tencent.com}{jamsluo@tencent.com}\par}
\end{center}

\begin{figure}[H]
  \centering
  \includegraphics[width=\linewidth]{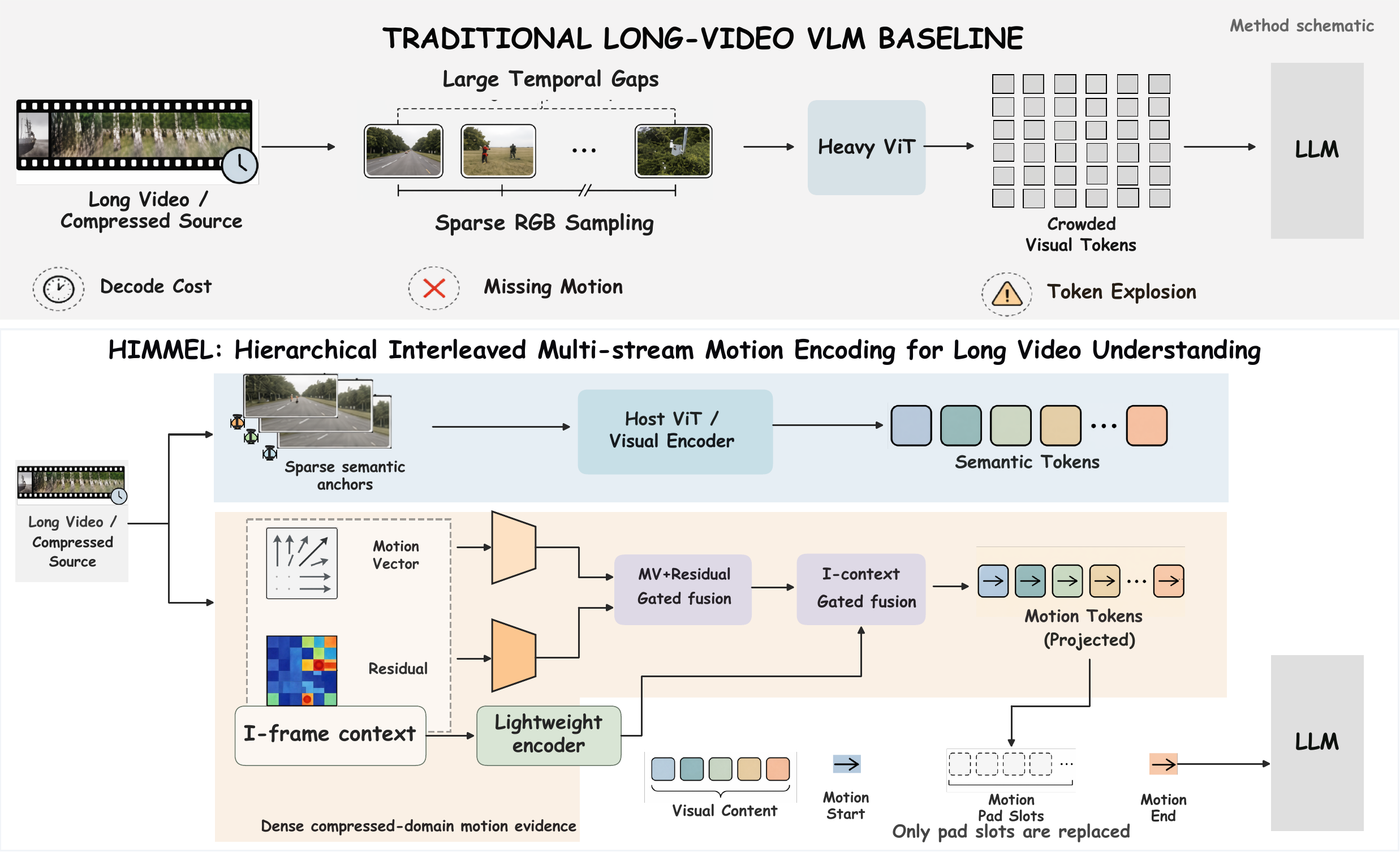}
  \caption{HIMMEL overview. Top: an RGB-frame baseline. Bottom: HIMMEL sends
    sparse semantic anchors to the frozen host ViT and dense inter-frame
    signals to a lightweight tri-stream adapter. MV and residual features
    are fused first; separately encoded I-frame context enters the second
    fusion stage. The output $H_k^{\mathrm{fused}}$ is projected by
    $\pi_{\mathrm{proj}}$ to aligned motion tokens $M_k$. These replace only
    the reserved motion-pad slots, not the motion-start/end embeddings,
    and are interleaved with the anchor's visual tokens for the LLM.}
  \label{fig:overview}
\end{figure}
\clearpage

\begin{abstract}
Long video understanding with multimodal language models suffers from three compounding
bottlenecks: heavy decode cost to obtain dense RGB frames, token growth with
frame count, and weak motion perception under sparse keyframe sampling.
Existing remedies either prune visual tokens after decoding, which leaves the
expensive RGB pipeline untouched, or they discretise codec motion vectors with a
generative tokeniser, which couples motion modelling to a heavy pre-training stage.
In this paper we present \textbf{HIMMEL}, a hierarchical video-language framework
that allocates semantic and motion capacity along separate paths.
A small set of sparse \emph{semantic anchor frames} is routed to the expensive host ViT to
ground object identity and scene layout, while the far denser inter-frame intervals
are encoded by a lightweight \emph{compressed-domain tri-stream adapter} that distils
motion evidence from motion-vector maps, residual maps, and an I-frame context branch
into aligned motion tokens.  These tokens are injected into the LLM via a
differentiable placeholder mechanism, after a dedicated Stage~1 contrastive alignment
that places the motion representation in a geometry compatible with the frozen visual
backbone.  We further show that an InfoNCE alignment objective beats MSE regression by
$+1.5$\,pp.
On Video-MME, HIMMEL surpasses the dense 32-frame baseline by $+2.3$\,pp
($61.2\to 63.5$\%) while using $\mathbf{3.6\times}$ fewer context tokens and running
end-to-end in $1.03$\,s per question instead of $2.75$\,s.  Switching to a stronger
Qwen3-VL-8B host pushes HIMMEL to $64.9$\%, narrowing the gap to much larger
proprietary systems while keeping inference on a single NVIDIA H20 GPU.  Extensive
ablations across stream composition, motion-encoder family, fusion mode, alignment
objective, anchor count, LoRA rank, and video duration confirm that the full
tri-stream contributes to the observed gains, and that the
benefit grows with video length.
\end{abstract}

\section{Introduction}
Long-video multimodal language models (MLLMs) typically encode sampled RGB frames
with a vision transformer and pass their tokens to an LLM. This has three
bottlenecks: \emph{decode cost}, as dense extraction adds latency and memory use;
\emph{context growth}, as additional frame tokens increase full-attention cost
quadratically; and \emph{blind motion}, as sparse sampling discards the dynamics
needed to distinguish what happened from what a static image shows.

HIMMEL spends the expensive ViT budget on sparse \emph{semantic anchors} and
recovers dynamics from compressed-domain signals. Motion vectors encode
codec-selected block displacements; residual maps encode differences after motion
compensation. Neither is ground-truth optical flow. Extraction costs depend on
the backend (Appendix~\ref{app:codec-compat}); the central saving is that dense
intervals bypass the host ViT. A tri-stream adapter fuses motion, residual, and
I-frame context into compact tokens, aligns them to the host visual space in
Stage~1, and injects them into the multimodal sequence. Expensive semantic
processing thus stays sparse while lightweight motion processing stays dense.

\paragraph{Contributions.}
\begin{itemize}[leftmargin=*]
  \item We propose a hierarchical semantic/motion decomposition for long-video
        MLLMs that combines sparse host-ViT processing with dense codec-derived
        motion tokens.
  \item We design a tri-stream adapter with separate branches and gated fusion,
        and demonstrate the benefit of I-frame context through controlled stream
        ablations (Section~\ref{sec:ablation}).
  \item We show that Stage~1 contrastive alignment with InfoNCE outperforms MSE
        feature regression by $1.5$\,pp on Video-MME under the same training budget.
  \item We evaluate HIMMEL across four host backbones and use controlled ablations
        and baseline comparisons to characterize its accuracy--efficiency
        trade-offs and the contributions of its components.
\end{itemize}


\section{Related Work}
\paragraph{MLLMs for long videos.}
Long-video MLLMs reduce temporal cost through frame selection, compact visual
representations, and streaming memory~\citep{llamavid,longvila,flashvstream,malmm}.
Frame-selection systems such as AKS, QuoTA, BOLT, and Static-or-Dynamic
select or allocate observations before downstream reasoning
\citep{aks,quota,bolt,staticdynamic}. HIMMEL instead allocates sparse host-ViT
processing to appearance and learns interval representations from motion and
residual inputs. These approaches can be complementary, but selection may
retain temporal evidence and should not be characterized as intrinsically
incapable of motion reasoning.

\paragraph{Compressed-domain video understanding.}
CoViAR~\citep{coviar} first showed that motion vectors and residuals carry strong
action cues at a fraction of decode cost, but it attached a classification head
rather than a language model.  Recent codec-aware video MLLMs extend this line to
generative video understanding.  Video-LaVIT~\citep{videolavit} tokenises MPEG-4
motion vectors with a VQ module and combines them with keyframe appearance tokens,
while CoPE-VideoLM~\citep{cope} adds a similar MV+residual adapter on top of a
frozen host VLM. ReMoRa~\citep{remora} combines I-frame features and refined
motion vectors with GOP-level state-space modeling and external flow supervision.
HIMMEL uses an explicit residual branch, I-frame-grounded staged fusion, and
host-visual-delta alignment without an external flow teacher.
InfoNCE~\citep{cpc}, LoRA~\citep{lora}, and gating are established components;
our contribution is their codec-specific representation and interface.
Appendices~\ref{app:codec-compat} and~\ref{app:videolavit} detail backend
boundaries and the Video-LaVIT comparison.

\paragraph{Efficient long-context reasoning.}
TimeSuite~\citep{timesuite} and LongVU~\citep{longvu} address the long-context
challenge via temporal token compression and adaptive frame selection.
Training-free approaches such as LOOK-M~\citep{lookm} and HERMES~\citep{hermeskv}
manage the KV cache without learning a motion representation.
Appendix~\ref{app:training-free} compares HIMMEL with these methods and
codec-guided pruning in CodecSight~\citep{codecsight}.
SCOPE-Router~\citep{yu2026scoperoutercostawareopensetvlm} selects among VLMs
to balance cost and quality; HIMMEL instead reallocates visual computation
within a chosen host.


\section{Method}

\subsection{Hierarchical video decomposition}\label{sec:decomp}

Given a video of $T$ frames, HIMMEL partitions the temporal axis into a sparse set of
\emph{semantic anchors} and a dense set of \emph{motion intervals}.  Let
$\mathcal{A} = \{t_1, t_2, \dots, t_{N_a}\}$ be the index set of anchor positions, with
$N_a \ll T$, chosen uniformly over $[1, T]$ at inference time.  A semantic anchor
is a sampled appearance frame; uniform temporal sampling does not imply that it is
an intra-coded frame in the original bitstream.  We retain ``I-frame context'' as
the name of the adapter's appearance branch.  The decomposition is
\begin{equation}
\label{eq:decomp}
  \underbrace{\mathbf{V} = \{x_1, x_2, \dots, x_T\}}_{\text{raw frames}}
  \;\;\longrightarrow\;\;
  \underbrace{\{x_t : t \in \mathcal{A}\}}_{\text{semantic path}}
  \;\cup\;
  \underbrace{\{I_k : k \in [1, K]\}}_{\text{motion path}},
\end{equation}
where $I_k$ is a temporal interval with endpoint indices $(a_k,b_k)$ and an associated
appearance anchor.  The interval count $K$ is determined by the temporal grouping;
if only intervals between consecutive anchors are used, $K=N_a-1$.
Anchors carry appearance and layout context; interval maps encode inter-frame
dynamics for the lightweight adapter.  This
decomposition turns long-video encoding into a resource allocation problem:
high-cost semantic processing stays sparse, low-cost motion processing stays dense.

\paragraph{Codec side-channel per interval.}
For each interval $I_k$, the preprocessing backend constructs three input streams.
Block-level displacement records are rasterized to a two-channel motion field;
the image-like adapter interface encodes its horizontal component, vertical
component, and magnitude as
$\mathbf{F}^{\text{mv}}_k \in \mathbb{R}^{H \times W \times 3}$.
A motion-compensated pixel-difference map
$\mathbf{F}^{\text{res}}_k \in \mathbb{R}^{H \times W \times 3}$ capturing unmodelled
appearance changes and an appearance-context patch $\mathbf{F}^{\text{ifr}}_k$
complete the three inputs.  Clipping, quantization, and resizing of these
image-like maps are model preprocessing operations, not lossless representations
of the encoded motion syntax.  The residual denotes a difference between a
decoded image and a motion-warped reference image, rather than raw transform
coefficients; its exact definition and the provenance of the motion field depend
on the extraction backend (Appendix~\ref{app:codec-compat}).
Because $I_k$ typically spans $T_k > 1$ frames, each branch receives in fact a short
temporal sequence $\{\mathbf{F}^{\star}_{k,t}\}_{t=1}^{T_k}$; we hide this axis in the
notation by assuming the per-branch encoder $\phi_{\star}$ reduces the temporal
dimension internally (either by temporal averaging, as in our default MotionSSM and
ResNet configurations, or by temporal attention, as in the SigLIP-ViT variant; see
Appendix~\ref{app:backbone}).
Extraction may require reconstructed reference frames or decoded luma, especially
when vectors are estimated rather than exported by the decoder.  HIMMEL's
architectural saving is that dense interval maps enter the lightweight adapter
instead of being processed as dense RGB inputs by the expensive host ViT.

\paragraph{Token-budget accounting.}
If anchor $i$ produces $n_i$ host-visual tokens, the visual input length is
$N_{\mathrm{vis}}=\sum_{i=1}^{N_a}n_i+K K_m$.
Our default uses $N_a=8$ and $K_m=64$; the measured mean visual-context sizes
are $12.2$\,k for HIMMEL and $44.7$\,k for the dense 32-frame baseline.
These measured totals depend on the host's visual preprocessing and interval
grouping; they are not obtained by identifying the anchor and interval counts.

\subsection{Tri-stream motion adapter}\label{sec:adapter}
\begingroup
\setstretch{1.18}

Figure~\ref{fig:tristream-example} illustrates what the three streams look like on a
fast-motion sports clip: the appearance anchors provide semantic context (players, hoop, court
lines), the motion-vector maps capture directional block-level flow (jumping,
dribbling, arm swings) as vivid polarised colours, and the residuals highlight
high-frequency appearance changes around object boundaries where motion compensation
is imperfect.  The stream ablations test whether these different representations
provide complementary information for downstream video understanding.

\begin{figure}[!t]
  \centering
  \includegraphics[width=\linewidth]{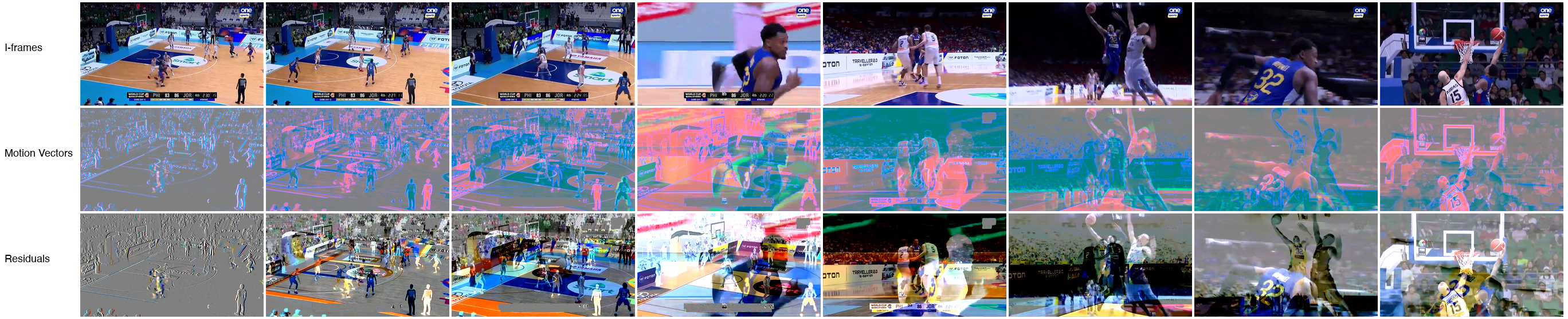}
  \caption{Tri-stream visualisation on Video-MME \#143 (basketball dunk).  Row~1:
    uniformly sampled appearance anchors.  Row~2: per-interval motion-vector maps capturing
    directional block-level flow.  Row~3: residual maps highlighting
    appearance changes that motion compensation cannot predict.  HIMMEL feeds
    all three streams to the tri-stream adapter and lets gated fusion decide which
    signal to trust per interval.}
  \label{fig:tristream-example}
\end{figure}

As summarised in Figure~\ref{fig:overview}, the compressed-domain adapter has three
parallel branches for I-frame context, motion vectors, and residuals.  Each branch is
a lightweight encoder $\phi_\star$ (either a state-space block or a small convolutional
tokeniser; see Appendix~\ref{app:backbone}) that maps the raw codec map to an
interval-level hidden-state sequence:
\begin{equation}
\label{eq:branches}
  H^{\text{mv}}_k   = \phi_{\text{mv}}(\mathbf{F}^{\text{mv}}_k),\quad
  H^{\text{res}}_k  = \phi_{\text{res}}(\mathbf{F}^{\text{res}}_k),\quad
  H^{\text{ifr}}_k  = \phi_{\text{ifr}}(\mathbf{F}^{\text{ifr}}_k),
\end{equation}
where each hidden state $H^{\star}_k \in \mathbb{R}^{K_m \times d}$ contains
$K_m$ tokens of width $d$.  Fusion operates on these token sequences; pooling is
applied subsequently for the Stage-1 alignment loss.

\paragraph{Staged gated fusion.}
MV and residual describe displacement and prediction error within an interval,
whereas I-frame context supplies appearance grounding.  This grouping motivates
two-stage fusion. For token sequences $P,S\in\mathbb{R}^{K_m\times d}$, let
$\bar P=\mathrm{LN}_P(P)$ and $\bar S=\mathrm{LN}_S(S)$. Each fusion module uses
normalized inputs, their difference, and their element-wise product:
\begin{align}
\label{eq:fuse-mr}
  Z &= [\bar P;\bar S;\bar P-\bar S;\bar P\odot\bar S],\qquad
  G = \sigma(ZW_g^\top+b_g), \\
  \mathcal F(P,S) &= \mathrm{LN}_{o}\!\left(G\odot P+(\mathbf1-G)\odot S+D(Z)\right),
\end{align}
where feature-wise concatenation gives $Z\in\mathbb{R}^{K_m\times4d}$,
$W_g\in\mathbb{R}^{d\times4d}$, and $D$ is a $4d\!\to\!d\!\to\!d$
MLP with GELU activation. The gate mixes the original inputs; the learned
correction $D(Z)$ and output LayerNorm complete the fusion. Two independently
parameterized instances first combine motion streams, then appearance:
\begin{equation}
  H^{\text{mr}}_k=\mathcal F_{\mathrm{mr}}(H^{\text{mv}}_k,H^{\text{res}}_k),\qquad
  H^{\text{fused}}_k=\mathcal F_{\mathrm{tri}}(H^{\text{mr}}_k,H^{\text{ifr}}_k).
\end{equation}
Thus both stages preserve the $K_m\times d$ token shape.  For Stage~1 only, we
form $h^{\text{fused}}_k=\mathrm{Pool}(H^{\text{fused}}_k)\in\mathbb{R}^d$.
The two gates use separate instances of \texttt{CompressedStreamFusion}; their
activation statistics and interpretation are discussed in
Appendix~\ref{app:mechanism}.
\endgroup

\subsection{Stage-1: motion-space alignment}\label{sec:stage1}

The host ViT, host multimodal projector, and LLM are frozen during Stage~1;
the tri-stream adapter and alignment heads are trainable.  An \texttt{AlignmentHead}
$\pi_{\text{align}}: \mathbb{R}^d \to \mathbb{R}^{d_v}$ projects each fused motion
embedding $h^{\text{fused}}_k$ into the host visual space so that it can be compared
to a visual target.  Let $\psi$ denote the frozen host ViT followed by its
multimodal projector.  The target $v_k \in \mathbb{R}^{d_v}$ is the difference
between pooled endpoint representations for interval $I_k$:
\begin{equation}
\label{eq:vdelta}
  v_k = \mathrm{Pool}\!\left(\psi(x_{b_k})\right)
        -\mathrm{Pool}\!\left(\psi(x_{a_k})\right),
\end{equation}
where the target is taken \emph{after} the host multimodal projector, so
$d_v=d_{\mathrm{llm}}$ for each host.  This feature difference supplies a
host-space summary of appearance change; it is not identified with a codec's
pixel prediction error.  Let $m_k = \pi_{\text{align}}(h^{\text{fused}}_k)$.  The alignment
loss combines a bidirectional InfoNCE~\citep{cpc} term with a cosine regulariser;
with a slight overload of notation, we let $k$ index minibatch \emph{examples}
(each example being a single interval $I_k$ drawn from some training video)
and $B$ denote the minibatch size:
\begin{align}
\label{eq:infonce}
  \mathcal{L}_{\text{InfoNCE}} &= -\tfrac{1}{2B}\sum_{k=1}^{B}\!\left[
    \log\frac{\exp(\mathrm{sim}(m_k, v_k)/\tau)}{\sum_{j=1}^{B}\exp(\mathrm{sim}(m_k, v_j)/\tau)}
   \;+\; \log\frac{\exp(\mathrm{sim}(v_k, m_k)/\tau)}{\sum_{j=1}^{B}\exp(\mathrm{sim}(v_k, m_j)/\tau)}
  \right], \\
\label{eq:align}
  \mathcal{L}_{\text{align}} &= \mathcal{L}_{\text{InfoNCE}}
  \;+\; \tfrac{\lambda_{\cos}}{B}\sum_{k=1}^{B}\bigl(1 - \cos(m_k, v_k)\bigr),
\end{align}
where $\mathrm{sim}(\cdot,\cdot)$ is cosine similarity, $\tau$ is a learned temperature,
and $\lambda_{\cos}$ balances the cosine regulariser against the contrastive term.  The
InfoNCE term enforces a \emph{relative} alignment: the motion code must be closer to
its own visual delta than to any other interval in the batch, while the cosine term
encourages agreement in direction.  In the
compressed tri-stream setting we additionally maintain branch-specific projections so
that the MV, residual, and I-frame sub-streams are individually forced to respect the
geometry of the host visual space.  This auxiliary supervision is intended to
support branch alignment; it does not guarantee a particular gate distribution.

\subsection{Stage-2: differentiable motion-token injection}\label{sec:stage2}

Stage~2 freezes the host ViT, host multimodal projector, and base LLM weights and
trains LoRA~\citep{lora} adapters on the LLM attention and MLP projections, together with the
motion adapter and token projector $\pi_{\text{proj}}$.
During Stage~2 we retain the unpooled sequence $H^{\text{fused}}_k$
of $K_m$ per-interval tokens (that is, we do \emph{not} feed the
Stage-1 contrastive pooled vector $m_k$ into the LLM; the pooled $m_k$ is only used
for the Stage-1 InfoNCE loss).  Concretely, the adapter's hidden state is already a
length-$K_m$ sequence $H^{\text{fused}}_k \in \mathbb{R}^{K_m \times d}$ after
staged fusion, and a lightweight projector $\pi_{\text{proj}}$,
initialised from $\pi_{\text{align}}$, maps each row of $H^{\text{fused}}_k$ from the
adapter width $d$ to the LLM embedding width $d_{\text{llm}}$, yielding
$M_k \in \mathbb{R}^{K_m \times d_{\text{llm}}}$.  Both projectors use the same
two-layer MLP shape, allowing an exact parameter copy at initialization because
$d_v=d_{\mathrm{llm}}$.  Their roles differ: $\pi_{\text{align}}$ is applied to a
pooled interval under the contrastive objective, whereas $\pi_{\text{proj}}$ is
applied token-wise and updated under the autoregressive objective.
All $K \cdot K_m$ motion tokens are
concatenated into $M \in \mathbb{R}^{(K K_m) \times d_{\text{llm}}}$ and injected into
the multimodal embedding stream at reserved \texttt{<|motion\_pad|>} positions.

\paragraph{One-hot scatter injection.}
We use out-of-place injection to avoid modifying tensors reused during autograd
or gradient-checkpoint recomputation.  Let
$E\in\mathbb{R}^{S\times d_{\mathrm{llm}}}$ be the multimodal embedding sequence
and $J=K K_m$ the total motion-token count.  Exactly $J$ reserved
\texttt{<|motion\_pad|>} positions occur at distinct ordered indices
$p_1,\ldots,p_J$.  Define $\Pi\in\{0,1\}^{S\times J}$ by
$\Pi[s,j]=\mathbb{1}[s=p_j]$.  Each column has one nonzero, and each row has
at most one.  The injection is
\begin{equation}
\label{eq:inject}
  E' \;=\; \mathrm{sg}(E) \odot \bigl(\mathbf{1} - \Pi \mathbf{1}_{J}\bigr)
          \;+\; \Pi\,M,
\end{equation}
where $\mathrm{sg}(\cdot)$ stops gradients through the frozen host embedding
path and the length-$S$ mask is broadcast along the hidden dimension.
Consequently, $E'[p_j]=M_j$ and all other rows retain their host embeddings.
The integer indices are fixed, while the operation is differentiable with
respect to $M$: $\partial\mathcal{L}/\partial M
=\Pi^\top(\partial\mathcal{L}/\partial E')$.
Gradients therefore reach $\pi_{\text{proj}}$ and the tri-stream adapter, as well
as the LLM LoRA weights.  The one-hot matrix defines the operation mathematically;
it can be implemented as indexed scatter without materializing a dense matrix.

\begin{figure}[H]
  \centering
  \includegraphics[width=\linewidth]{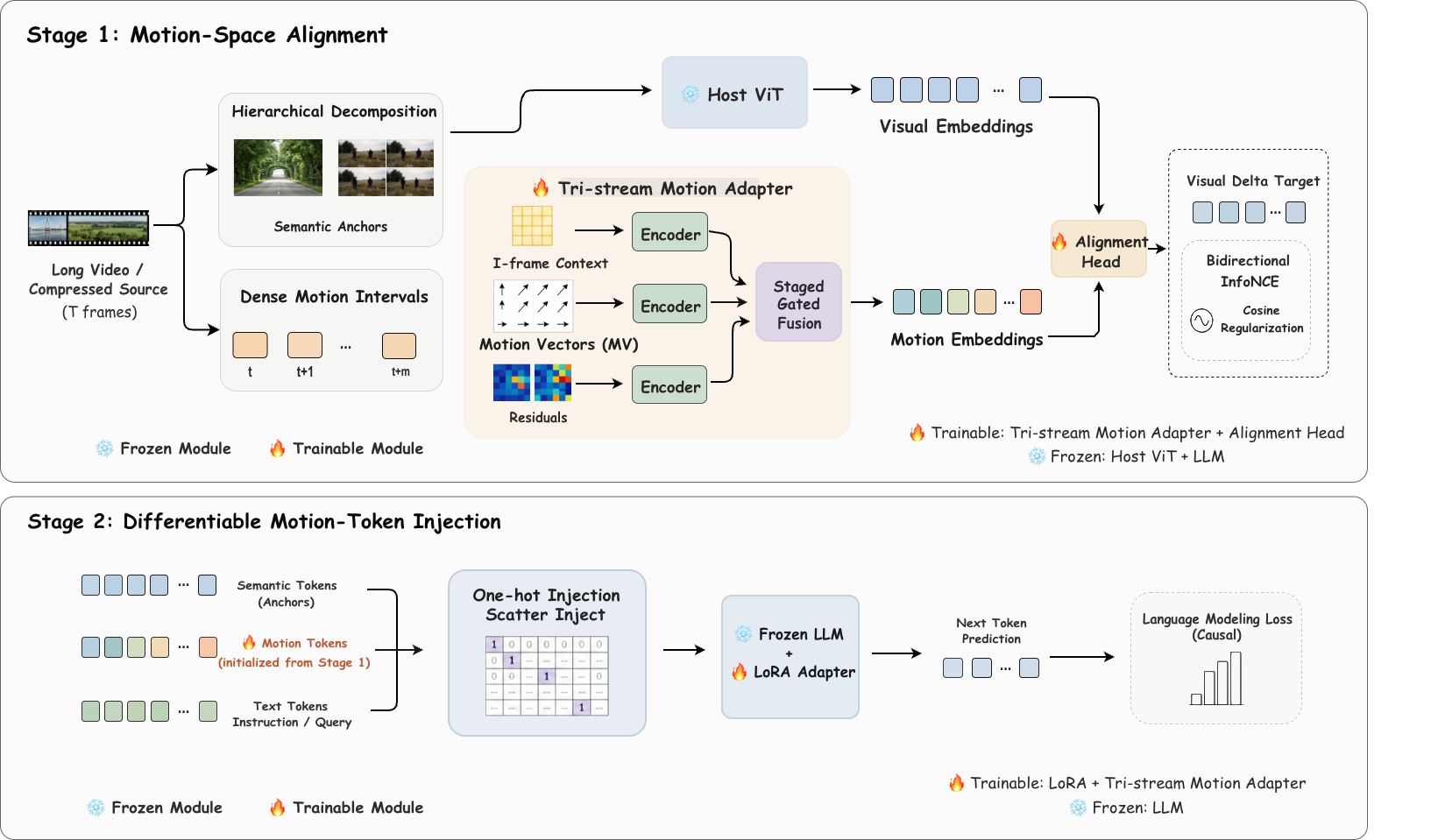}
  \caption{Two-stage training.  The branch encoders $\phi_{\mathrm{mv/res/ifr}}$
    and staged fusion produce $H_k^{\mathrm{fused}}$.  Stage~1 pools this sequence
    and applies $\pi_{\mathrm{align}}$ to match the host-projected endpoint
    difference $v_k$ via Eq.~\ref{eq:align}.  Stage~2 applies
    $\pi_{\mathrm{proj}}$ to each unpooled token to obtain $M_k$, injects $M_k$
    through Eq.~\ref{eq:inject}, and trains the adapter, token projector, and LLM
    LoRA while retaining the frozen host ViT, multimodal projector, and base LLM.}
  \label{fig:training}
\end{figure}

\paragraph{Training objective.}
Stage~2 uses the standard language-modelling loss on the target answer conditioned on
anchor tokens, motion tokens, and the text prompt:
\begin{equation}
\label{eq:stage2}
  \mathcal{L}_{\text{SFT}} \;=\;
  -\sum_{n=1}^{|y|} \log p_\theta\!\bigl(y_n \mid y_{<n},\, E'\bigr),
\end{equation}
with $\theta$ collecting the LoRA parameters, adapter weights, and
$\pi_{\text{proj}}$ parameters.
Figure~\ref{fig:training} summarises the overall two-stage optimisation schedule.

\section{Experiments}
\subsection{Setup}
\paragraph{Benchmarks.}
We evaluate video understanding on Video-MME~\citep{videomme} (2,700 questions),
MVBench~\citep{videochat2} (4,000 questions, 20 tasks), and
MLVU~\citep{mlvu} (multiple-choice M-Avg). MathVista-testmini~\citep{mathvista}
(1,000 questions) and GSM8K-test~\citep{gsm8k} (1,319 questions) assess retention
of visual and text-only mathematics; MathVision/MATH-V~\citep{mathvision}
is reported separately in the appendix.

\paragraph{Baseline and HIMMEL configuration.}
Our dense-frame baseline is Qwen2.5-VL-7B~\citep{qwen25vl} with 32 uniformly sampled frames
(our local evaluation gives 61.2\% on Video-MME).
HIMMEL uses the same host backbone with 8 semantic anchors plus the full compressed
tri-stream adapter (MotionSSM encoder, gated fusion, $r{=}32$ LoRA).  This LoRA-based
Stage-2 variant is the default HIMMEL configuration reported throughout the paper.
Unless stated, all ablations vary only the listed dimension and keep everything else fixed.

\subsection{Main results}\label{sec:main}

Table~\ref{tab:main} reports the stream-composition ablation with the Qwen2.5-VL-7B
backbone. The adapter variants share the two-stage training recipe and differ
in the streams fed into the adapter (see
Appendix~\ref{app:ablation-full} for further details).  We add one simple RGB-frame
baseline, ``\emph{+ RGB-diff tokens}'', which replaces the compressed streams with
per-interval pixel-difference images $|x_{t_{k+1}}-x_{t_k}|$ tokenised by the same
motion encoder, so as to isolate the contribution of codec-derived signals from
naive RGB motion proxies.
The results tell a clear hierarchical story: (1)~dropping from 32 dense
frames to 8 anchor frames alone costs $-3.2$~pp on Video-MME; (2)~adding MV or residual
tokens \emph{in isolation} hurts further in this configuration;
(3)~joining MV and residual \emph{without} I-frame context remains below the
anchor-only configuration under the evaluated alignment recipe
(Appendix~\ref{app:ablation-full});
(4)~the full tri-stream with the I-frame context branch in the adapter surpasses the
dense baseline by $+2.3$~pp while using $3.6\times$ fewer context tokens.
The matched retention evaluation gives 682 versus 681 correct MathVista answers
for the untouched host and video-only HIMMEL, respectively. This small difference
does not by itself establish statistical equivalence (Appendix~\ref{app:cr-math}).

\begin{table}[!htbp]
  \centering
  \caption{Main results on Video-MME (2700~Q) and MVBench (4000~Q).  All results are
    based on Qwen2.5-VL-7B as the host backbone.  We report Video-MME numbers from our
    own evaluation, while the MVBench baseline is taken from the official Qwen2.5-VL
    technical report~\citep{qwen25vl}.
    Ctx = average visual context tokens (k) used by each configuration
    (anchor ViT tokens plus motion-branch tokens). Exact-count mathematical
    retention is evaluated separately in Appendix~\ref{app:cr-math}.}
  \label{tab:main}
  \setlength{\tabcolsep}{4pt}
  \begin{tabular}{lcccc}
    \toprule
    Method & Video-MME & MVBench & MLVU & Ctx (k) \\
    \midrule
    Dense 32-frame baseline & 61.2 & 69.6 & 70.2 & 44.7 \\
    \midrule
    8 anchors only            & 58.0 & 67.0 & 66.8 & 11.2 \\
    + RGB-diff tokens         & 60.0 & 68.0 & 68.4 & 13.2 \\
    + MV tokens only          & 55.9 & 65.0 & 52.1 & 12.2 \\
    + Residual tokens only    & 54.9 & 64.0 & 49.7 & 12.2 \\
    + MV+Residual (no I-frame ctx.) & 50.8 & 60.5 & 46.5 & 12.2 \\
    + Full tri-stream (HIMMEL)      & \textbf{63.5} & \textbf{69.9} & \textbf{71.6} & \textbf{12.2} \\
    \bottomrule
  \end{tabular}
\end{table}

\paragraph{Controlling tuning and context budget.}
The untouched dense host, dense host with the same video SFT, and HIMMEL obtain
61.22, 62.04, and 63.52 on Video-MME, respectively. Thus the improvement over
same-data dense SFT is 1.48\,pp, rather than the full 2.30\,pp over the untouched
host. At the matched 12.2k visual-token budget, the RGB-only control obtains
61.74 versus HIMMEL's 63.52. Appendix~\ref{app:cr-controls} reports the
equal-budget comparisons separately from published-model reference results.

\subsection{Comparison with state of the art}\label{sec:sota}

Table~\ref{tab:sota} places HIMMEL in the landscape of published 7--8\,B models and
large-scale proprietary systems.  On Qwen2.5-VL, HIMMEL outperforms the same-backbone
baseline by $+2.3$~pp while using $3.6\times$ fewer context tokens.  Switching the host
to Qwen3-VL~\citep{qwen3vl} pushes HIMMEL to 64.9\% ($+2.2$~pp over its own 32-frame baseline), further
illustrating transfer to another host. Published reference rows use their original
protocols, which can differ in frame count, resolution, and modalities; they are
not controlled comparisons of architectural quality. Within-protocol
comparisons on other backbones are in Appendix~\ref{app:multibench}.

\begin{table}[!htbp]
  \centering
  \caption{Video-MME (w/o subtitles) comparison.
    Rows marked $^{\dagger}$ are our local
    evaluation so the numbers differ from the vendor-reported headline
    figures (which use higher resolution, more frames, or subtitles).
    \emph{Bottom}: proprietary / large-scale reference models (scores from
    official reports or leaderboards).
    Unmarked rows are published reference scores, not matched reruns.
    Ctx = mean visual context tokens (k); reduction is in token count, not latency.}
  \label{tab:sota}
  \setlength{\tabcolsep}{5pt}
  \begin{tabular}{lcccc}
    \toprule
    Method & Params & Video-MME & Ctx (k) & Token reduction \\
    \midrule
    \multicolumn{5}{l}{\textit{Open-weight 7--8\,B models}} \\
    VideoChat2~\citep{videochat2}             & 7B  & 54.1 & 32.0 & --- \\
    Video-LLaVA~\citep{videollava}            & 7B  & 56.3$^{\dagger}$ &  8.0 & --- \\
    LLaVA-Video~\citep{llavavideo}$^{\dagger}$ & 7B & 61.3 & 44.7 & --- \\
    InternVL3-8B$^{\dagger}$~\citep{internvl3} & 8B & 62.3 & 44.7 & --- \\
    \midrule
    Qwen2.5-VL (32 fr., ours)$^\dagger$       & 7B  & 61.2 & 44.7 & 1.0$\times$ \\
    HIMMEL (Qwen2.5-VL)$^\dagger$             & 7B  & 63.5 & 12.2 & \textbf{3.6$\times$} \\
    HIMMEL (Qwen3-VL)$^\dagger$               & 8B  & \textbf{64.9} & 12.2 & \textbf{3.6$\times$} \\
    \midrule
    \multicolumn{5}{l}{\textit{Proprietary / large-scale reference models}} \\
    GPT-4o~\citep{gpt4o}                      & --- & 71.9 & --- & --- \\
    Qwen2.5-VL-72B~\citep{qwen25vl}           & 72B & 73.3 & --- & --- \\
    \bottomrule
  \end{tabular}
\end{table}

\subsection{Comparison with codec-aware and token-pruning baselines}\label{sec:codec-pruning}

To position HIMMEL among methods that explicitly target the context-token-efficiency
frontier, Table~\ref{tab:codec-pruning} groups two representative families:
(i)~codec-aware compressed-domain VLMs that learn a motion representation, and
(ii)~training-free visual-token pruning, KV-cache management, or codec-guided pruning that
operates on visual tokens or caches. CoPE and HIMMEL use different token budgets
in this summary; it is not an equal-budget ranking. The matched CoPE/RGB/HIMMEL
control and additional frame-selection and learned-compression baselines are in
Appendix~\ref{app:cr-controls}. Some selection methods achieve higher absolute
accuracy with substantially larger contexts, so the relevant comparison is the
accuracy--cost trade-off. Extended timing and per-category results are in
Appendix~\ref{app:training-free}.

\begin{table}[!htbp]
  \centering
  \caption{HIMMEL vs.\ codec-aware learners and token-pruning methods on Video-MME
    (2700~Q). All learnable methods use 7--8\,B hosts, but this summary does not
    match host, training data, and token budget across every method. The
    reduction column refers to tokens, not measured wall-clock speed.}
  \label{tab:codec-pruning}
  \setlength{\tabcolsep}{4pt}
  \renewcommand{\arraystretch}{1.15}
  \begin{tabular}{@{}llcccc@{}}
    \toprule
    Method & Family & V-MME & Ctx (k) & Token red. & Trains? \\
    \midrule
    Dense 32-fr.\ (Qwen2.5-VL)   & \makecell[l]{baseline}                & 61.2 & 44.7 & 1.0$\times$ & --- \\
    \midrule
    \multicolumn{6}{l}{\textit{(i) Codec-aware learnable (compressed-domain VLMs)}} \\
    CoPE-VideoLM~\citep{cope}    & \makecell[l]{MV+Res\\adapter}         & 61.9 & $\sim$3.1 & --- & \checkmark \\
    \textbf{HIMMEL (Qwen2.5-VL)} & \makecell[l]{tri-stream\\+ InfoNCE}   & \textbf{63.5} & 12.2 & 3.6$\times$ & \checkmark \\
    \textbf{HIMMEL (Qwen3-VL)}   & \makecell[l]{tri-stream\\+ InfoNCE}   & \textbf{64.9} & 12.2 & 3.6$\times$ & \checkmark \\
    \midrule
    \multicolumn{6}{l}{\textit{(ii) Training-free token pruning / KV management / codec-guided}} \\
    Panel ($3{\times}3$ grid)           & \makecell[l]{spatial tile\\merge}     & 61.5 & 11.1 & --- & \ding{55} \\
    LOOK-M (ratio 0.25)~\citep{lookm}   & \makecell[l]{KV cache\\eviction}      & 60.9 & 44.7 & --- & \ding{55} \\
    HERMES~\citep{hermeskv}             & \makecell[l]{hierarchical\\KV}        & 60.9 & 44.7 & --- & \ding{55} \\
    CodecSight~\citep{codecsight}       & \makecell[l]{codec-guided\\pruning}   & 60.7 & $\sim$17.3 & --- & \ding{55} \\
    \bottomrule
  \end{tabular}
\end{table}

\subsection{Efficiency analysis}\label{sec:efficiency}

HIMMEL allocates host visual tokens to sparse anchors and a fixed number of
motion tokens to each interval. In general,
$N_{\mathrm{visual}}=\sum_{i=1}^{N_a}T_{\mathrm{host},i}+K K_m$,
where $T_{\mathrm{host},i}$ is the token count after the host's spatial merging
and $K$ is the actual interval count. Token counts depend on preprocessing and
the host interface; they must not be inferred solely from an image's pixel size.
The reported average visual contexts are 44.7k for the dense baseline and 12.2k
for HIMMEL (approximately $3.6\times$ reduction).
End-to-end wall-clock inference per Video-MME question drops from $2.75$\,s (dense)
to $1.03$\,s ($\approx 2.7\times$ speedup; see Appendix~\ref{app:training-free} for
the full per-method breakdown against training-free baselines). The corresponding
single-H20 measurements are 0.364 versus 0.974 samples/s and 20.09 versus
17.75\,GiB peak allocated memory. Appendix~\ref{app:cr-efficiency} distinguishes
the latency and memory workloads and the allocator-based memory definition.
Results on PerceptionTest~\citep{perceptiontest}, ActivityNet-QA~\citep{activitynetqa},
and the Video-MME long sub-split are in Appendix~\ref{app:per-benchmark}.

\FloatBarrier

\subsection{Ablation study}\label{sec:ablation}
\begingroup
\setstretch{1.18}
The full ablation across all design dimensions is presented in
Appendix~\ref{app:ablation-full}.  Here we highlight the most important findings.

\textbf{Alignment stage.}
Removing Stage~1 and training Stage~2 directly from a random motion initialisation drops
Video-MME from 63.5 to 62.0\% ($-1.5$~pp).  A randomly initialised motion module with no
training at all yields 55.0\%, confirming that both stages contribute and neither can be
omitted without a performance cost in this setting.

\textbf{Anchor frame count.}
The sweet spot is $N_a = 8$: 4 anchors are insufficient for scene grounding (59.0\%),
and adding more increases accuracy (64.0\% at 12, 65.0\% at 16,
66.0\% at 32) at higher context cost. Eight is a practical operating point,
not a universally optimal anchor count.

\textbf{Video duration.}
HIMMEL's gain grows with video duration: $+1.3$~pp on short clips ($<2$\,min),
$+2.0$~pp on medium clips ($2$--$15$\,min), and $+3.6$~pp on long content ($>15$\,min);
see Appendix~\ref{app:duration} for the exact per-bucket counts.
This is the expected behaviour of dense motion tokens: they matter more when temporal
extent exceeds what a few anchor frames can cover.

\textbf{Aligning without the I-frame branch.}
Removing the adapter's I-frame context branch makes Stage-1 optimization
less effective in our recipe and lowers downstream accuracy below the anchor-only
reference. This is an empirical failure of the evaluated configuration, not a
claim that all motion-only representations are unlearnable. Complementary controls
give 61.38 for MV+I and 60.89 for residual+I, versus 63.52 for the full tri-stream.
Frozen RAFT~\citep{raft} optical-flow controls also benefit from appearance
grounding but remain below the full tri-stream under this recipe
(Appendix~\ref{app:cr-flow}).

\textbf{Alignment objective: InfoNCE vs.\ MSE regression.}\label{sec:alignment-ablation}
A natural alternative to our contrastive InfoNCE alignment is MSE-based feature
reconstruction.  We replace our InfoNCE + cosine loss (Eq.~\ref{eq:align}) with an MSE
regression on the projected pooled interval representation:
$\mathcal{L}_{\text{MSE}} = \frac{1}{B}\sum_k\|m_k - v_k\|^2$. Under identical Stage-1 budgets and
Stage-2 tuning, MSE alignment achieves 62.0\% on Video-MME
(\textbf{$-1.5$~pp} vs.\ InfoNCE, 63.5\%).  The gap is consistent across categories:
InfoNCE outperforms on action ($+1.8$~pp) and motion ($+1.2$~pp) tasks while remaining
neutral on static-appearance tasks ($+0.1$~pp). A possible explanation is that
relative contrastive discrimination is better suited to the coarse endpoint-delta
target than absolute feature regression. These results do not establish a general
collapse property of MSE or prove a unique causal mechanism.
Appendix~\ref{app:alignment-objective} provides the full ablation, including a
category breakdown and a discussion of the different alignment objectives.

\endgroup

\FloatBarrier


\section{Limitations}
\textbf{Single-video scope.}
HIMMEL processes a supplied clip, rather than finding videos or moments through
open-web search as in ShotFinder and RACLO
\citep{yu2026shotfinderimaginationdrivenopendomainvideo,yu2026closedpoolvideoretrievalbenchmark}.
Stage-1 alignment creates no persistent cross-video memory; multi-clip reasoning
would require explicit clip identities, temporal resets, and memory or retrieval
boundaries.

\textbf{Modality and language coverage.}
HIMMEL adds no audio encoder or OCR branch. Sampled anchors expose visible text
to the host image pathway but may miss short-lived text; MV/residual inputs do
not decode textual semantics. External subtitles use the host text pathway
(Appendix~\ref{app:long-bench}). Audio-driven search, as evaluated by
Omni-DeepSearch~\citep{yu2026omnideepsearchbenchmarkaudiodrivenomnimodal},
is outside our scope.

\paragraph{Codec and representation scope.}
Decoder-exported vectors depend on encoder settings and reference-frame structure;
estimated flow or RGB proxies are not interchangeable with native codec signals.
The backend limitations are explicit in Appendix~\ref{app:codec-compat}.
Endpoint visual deltas may miss motion that returns to its initial appearance.
Fixed-length motion tokenization trades temporal detail for a predictable budget.
The non-video retention results cover only the evaluated mathematics benchmarks,
not universal out-of-domain robustness.

\paragraph{Broader impact.}
More efficient video analysis can reduce computation, but can also lower the
cost of surveillance or processing videos without consent. HIMMEL does not address
human-preference alignment~\citep{yu2025aligningmultimodalllmhuman} or search-grounded
verification of video claims~\citep{yu2026seeingbelievingbenchmark}.
Host-model biases and privacy
risks remain; efficiency does not establish safety-critical suitability.

\section{Conclusion and Future Directions}
HIMMEL introduces a hierarchical allocation of semantic and motion processing for
long-video MLLMs.  By routing sparse anchor frames to the expensive host ViT and dense
inter-frame intervals to a lightweight compressed-domain tri-stream adapter, HIMMEL
achieves a $3.6\times$ context-token reduction and a $+2.3$~pp Video-MME improvement
over the dense-frame baseline. Controlled ablations favor the full tri-stream
configuration and contrastive alignment within the evaluated recipe, while
8 anchors provide a practical accuracy--context trade-off.

\paragraph{Further directions.}
Promising extensions include (i)~jointly learned adaptive anchor and token allocation
beyond the evaluated sampling heuristics, (ii)~evaluating how savings scale to
72\,B-class backbones, and (iii)~extending the
compressed-domain representation to audio-visual settings where audio codecs offer
analogous side-channel metadata.

\begin{ack}
We thank the anonymous reviewers for their constructive feedback.
This work received financial and computational support from Tencent.
\end{ack}

\bibliographystyle{plainnat}
\bibliography{refs}

\appendix

\newpage
\vspace{2cm}
{\color{blue!70!black} \noindent \Huge \bf Appendix}
\vspace{0.5cm}

\titlecontents{section}[2em]
  {\color{blue!70!black}\bfseries}
  {\color{blue!70!black}\contentslabel{2em}}
  {}
  {\color{blue!70!black}\leaders\hbox{\kern2pt.\kern2pt}\hfill\contentspage}
  [\vspace{0pt}]

\titlecontents{subsection}[4em]
  {\color{blue!70!black}\itshape}
  {\color{blue!70!black}\contentslabel{2.3em}}
  {}
  {\color{blue!70!black}\leaders\hbox{\kern2pt.\kern2pt}\hfill\contentspage}
  [\vspace{0pt}]

\startcontents[appendixtoc]
\printcontents[appendixtoc]{}{1}{}

\newpage

\section{Reproducibility Statement}\label{app:repro}
The manuscript specifies the architecture, training states, principal
hyperparameters, evaluation controls, and statistical conventions.
Benchmark media and third-party checkpoints remain subject to their original
access terms and are not redistributed with the manuscript source.
The flagship recipe is \textbf{Qwen2.5-VL + tri-stream adapter + MotionSSM +
two-stage training with rank-32 LLM LoRA}.

\section{Open Access to Code and Data}
A sanitized core implementation is available at
\url{https://github.com/Sunbeam23333/HIMMEL}
(release revision \texttt{dcdd15c}). It includes the model components,
archived training recipes, a prediction runner, the standalone multi-codec C
extractor, and validation tools. The repository documents its supported
interfaces and limitations; it is not a checkpoint-exact reproduction package
for every experiment. In particular, the archived rank-64 training example is
not the rank-32 flagship configuration, and the standalone C extractor is not
silently substituted into the historical CoViAR training path.
Reproducing a numerical result additionally requires its checkpoint, frozen data
manifest, preprocessing configuration, and evaluation outputs. These complete
per-run artifacts are not included in the current release, and figure scripts
alone do not establish experimental reproducibility. Benchmark media and
third-party checkpoints must be obtained under their original access terms.

\FloatBarrier

\section{Experimental Settings and Compute}\label{app:settings}
\textbf{Hardware:} 8$\times$ NVIDIA H20 (96\,GB HBM3) per run, connected by NVLink.
\textbf{Precision:} bfloat16 with gradient checkpointing.
\textbf{Optimizer:} AdamW, cosine schedule, gradient accumulation over 8 microsteps,
DeepSpeed ZeRO-2.
\textbf{Stage 1:} alignment loss with $\lambda_\text{cos}=0.1$; $\sim$21k optimizer steps;
LR $1\times10^{-4}$; host ViT, multimodal projector, and LLM frozen;
motion adapter and alignment heads trainable.
\textbf{Stage 2:} SFT with LoRA ($r{=}32$, $\alpha{=}64$) on the LLM attention
(q/k/v/o) and MLP (gate/up/down) projections;
$\sim$1.5--3.5k optimizer steps (backbone-dependent); LR $5\times10^{-5}$;
motion adapter and token projector $\pi_{\mathrm{proj}}$ trainable;
host ViT, multimodal projector, and non-LoRA LLM weights frozen.
A comparison between LoRA and full-parameter SFT is provided in
Appendix~\ref{app:lora-vs-full}; the public archived recipes do not constitute
the complete checkpoint/configuration mapping for this comparison.
Unless explicitly identified as a mean across training seeds, a benchmark score
is a single-checkpoint evaluation.  Such a score does not measure training-seed
variance; see Appendix~\ref{app:stats} for the interpretation of confidence intervals.

\paragraph{Wall-clock compute budget.}
On the Qwen2.5-VL-7B flagship configuration, Stage~1 converges in $\sim$10 wall-clock
hours and Stage~2 in $\sim$11 wall-clock hours on a single 8$\times$H20 node, giving
a total of $\mathbf{80 + 88 = 168}$~\textbf{GPU-hours} for the two reported
training stages.  This accounting covers the flagship training run; it does not
represent the aggregate compute used for all ablations, data preparation, and
evaluation.

\paragraph{Trainable parameter budget.}
Table~\ref{tab:params} identifies which modules are optimized in each stage.
Let $N_{\mathrm{adapter}}$ count all unique branch, temporal-encoder, and fusion
parameters, excluding the separately named projection heads.  Let
$N_{\mathrm{align}}$ include the fused and any branch-specific Stage-1 heads,
$N_{\mathrm{proj}}$ count the Stage-2 token projector, and
$N_{\mathrm{LoRA}}$ count the LLM LoRA parameters.  The totals are
\begin{align}
  N_{\mathrm{train},1}&=N_{\mathrm{adapter}}+N_{\mathrm{align}},\\
  N_{\mathrm{train},2}&=N_{\mathrm{adapter}}+N_{\mathrm{proj}}+N_{\mathrm{LoRA}}.
\end{align}
For ordinary rank-$r_\ell$ LoRA applied to a linear layer with dimensions
$(d_{\mathrm{in},\ell},d_{\mathrm{out},\ell})$, its two low-rank matrices add
$r_\ell(d_{\mathrm{in},\ell}+d_{\mathrm{out},\ell})$ parameters; the sum runs over
the actual targeted layers.  Any additional trainable biases or saved modules
must be counted separately.  Checkpoint accounting uses the number of elements
in unique parameters with \texttt{requires\_grad=True}, avoiding double-counting
shared modules.  In particular, replacing the Stage-1 objective does not remove
the trainable Stage-2 token projector from the accounting.

\begin{table}[H]
  \centering\small
  \caption{Trainability of the flagship Qwen2.5-VL-7B modules.  The Stage-2 token
    projector is initialized from the shape-compatible fused alignment head and
    remains trainable; the Stage-1 contrastive objective and auxiliary alignment
    heads are not used for Stage-2 SFT.}
  \label{tab:params}
  \begin{tabular}{p{0.44\linewidth}cc}
    \toprule
    Component & Stage 1 & Stage 2 \\
    \midrule
    Host ViT and multimodal projector              & frozen & frozen \\
    Base LLM weights                               & frozen & frozen \\
    MV branch ($\theta_\text{mv}$)                 & trainable & trainable \\
    Residual branch ($\theta_\text{res}$)           & trainable & trainable \\
    I-frame context branch ($\theta_\text{ifr}$)    & trainable & trainable \\
    Temporal encoding and staged fusion            & trainable & trainable \\
    Alignment heads ($\pi_{\mathrm{align}}$, auxiliary heads) & trainable & not used \\
    Token projector ($\pi_{\mathrm{proj}}$)         & not used & trainable \\
    LLM LoRA ($r{=}32$)                             & not used & trainable \\
    \bottomrule
  \end{tabular}
\end{table}

Detailed hyperparameter table:

\begin{table}[H]
  \centering\small
  \caption{Training hyperparameters for the flagship Qwen2.5-VL configuration.}
  \begin{tabular}{lcc}
    \toprule
    Hyperparameter & Stage 1 & Stage 2 \\
    \midrule
    Learning rate & $1\times10^{-4}$ & $5\times10^{-5}$ \\
    Batch size (global) & 128 & 64 \\
    Warmup steps & 500 & 100 \\
    LR schedule & cosine & cosine \\
    Optimizer & AdamW & AdamW \\
    Gradient clip & 1.0 & 1.0 \\
    Weight decay & 0.01 & 0.01 \\
    LoRA rank ($r$) & --- & 32 \\
    LoRA $\alpha$ & --- & 64 \\
    LoRA target modules & --- & q/k/v/o/gate/up/down \\
    Motion token budget ($K_m$) & 64/interval & 64/interval \\
    Anchor frames ($N_a$) & 8 & 8 \\
    Training epochs & 1 & 1 \\
    \bottomrule
  \end{tabular}
\end{table}

Text-token truncation and final multimodal sequence length are distinct settings.
For Stage~2, the final sequence contains the host visual tokens, injected motion
tokens, prompt tokens, and answer tokens; its allocation must cover their sum.
Text-length settings therefore cannot be interpreted as final context limits
for the reported 12.2k/44.7k visual-token evaluation configurations.  Training
and evaluation may use different visual budgets.

\section{Statistical Reporting}\label{app:stats}
For benchmarks with integer correct/total counts (Video-MME 2700, MVBench 4000,
MathVista 1000, MathVision 3040), we report 95\% Wilson confidence intervals derived
from exact counts.  For MLVU M-Avg (the benchmark's macro-average of
multiple-choice task accuracies),
we report the scalar value without a CI.

Because the baseline and HIMMEL are evaluated on the \emph{same} 2700 Video-MME
questions, separate Wilson intervals do not constitute a paired significance
test.  The aggregate difference is $62=1715-1653$ additional correct answers.
An exact McNemar test additionally requires the two discordant counts: questions
answered correctly only by HIMMEL and questions answered correctly only by the
baseline.  Their difference alone does not determine a $p$-value, so we do not
infer paired statistical significance from the aggregate scores.
The Wilson intervals describe item-level binomial uncertainty conditional on
the evaluated checkpoint; they do not quantify variation across training seeds
or account for dependence among questions from the same video.  A video-level
paired bootstrap would require the original per-question predictions and video
identifiers.  Deterministic greedy re-evaluation is not an independent training
replicate.

\section{Controlled Evaluations and Capability Retention}
\label{app:cr-evaluation}

The following controls distinguish the contributions of video instruction
tuning, visual context, appearance grounding, and the motion representation.
Unless stated otherwise, video accuracy is evaluated on the 2,700-question
Video-MME test set without subtitles, using Qwen2.5-VL-7B and greedy decoding.
Visual context counts include anchor and motion tokens, but exclude question
text. Newly trained controls are reported as means over three independent
training seeds, except for the reused single-checkpoint references explicitly
identified below. These rounded means should not be interpreted as integer
correct counts for one checkpoint. Fixed-checkpoint, training-free
evaluations use deterministic greedy decoding. Exact single-checkpoint counts
are provided for the mathematical-retention evaluation below.

\subsection{Instruction tuning and equal-context controls}
\label{app:cr-controls}

Table~\ref{tab:cr-fair-controls} separates instruction-tuning gains from the
effect of changing the visual representation. Applying the same video SFT to
the dense host improves Video-MME from 61.22 to 62.04, while HIMMEL reaches
63.52. Thus the improvement over the untouched host is not entirely explained
by additional SFT. At the same 12.2k visual-token budget, HIMMEL exceeds the
RGB-anchor control by 1.78 percentage points. This comparison supports using
some context capacity for interval motion rather than allocating it entirely
to RGB appearance. The lower-budget configurations are separate operating
points with different anchor and motion allocations, rather than a sweep of
a single unchanged input sequence.

At the matched 12.2k budget, HIMMEL improves Action Recognition, Temporal
Perception, and Temporal Reasoning over RGB anchors by 3.52, 3.64, and
6.78 percentage points, respectively. RGB anchors retain a 0.72-point
advantage on the appearance-dominated OCR subset. These category differences
support complementary temporal evidence, rather than a uniform benefit on
every type of question. They are percentage-point differences, not additional
correct counts; the corresponding category denominators are not specified
in this aggregate comparison.

\begin{table}[htbp]
  \centering\small
  \caption{Video instruction tuning and visual-context controls. Comparisons
    are grouped by the factor held fixed. The first block reproduces the
    original single-checkpoint controls; new trained equal-context controls
    are three-seed means. All entries are Video-MME accuracy
    (\%); CoPE-VideoLM is compared at approximately the same 3.1k-token
    context, not at HIMMEL's default operating point.}
  \label{tab:cr-fair-controls}
  \setlength{\tabcolsep}{4pt}
  \begin{tabular}{p{0.57\linewidth}rr}
    \toprule
    Configuration & Visual ctx (k) & Acc.\,(\%) \\
    \midrule
    Dense host, untouched & 44.7 & 61.22 \\
    Dense host, same video SFT & 44.7 & 62.04 \\
    HIMMEL, same video SFT & 12.2 & 63.52 \\
    \midrule
    RGB anchors, matched context & 12.2 & 61.74 \\
    HIMMEL, matched context & 12.2 & 63.52 \\
    RGB anchors, lower context & 6.1 & 58.57 \\
    HIMMEL, lower context & 6.1 & 58.99 \\
    \midrule
    CoPE-VideoLM~\citep{cope} & $\approx$3.1 & 61.89 \\
    RGB-only, low-context control & $\approx$3.1 & 59.30 \\
    HIMMEL, low-context control & $\approx$3.1 & 62.10 \\
    \bottomrule
  \end{tabular}
\end{table}

The effect of video SFT also depends on the evaluation domain. For the dense
Qwen2.5-VL host, untouched $\to$ same-SFT scores are
57.44$\to$57.89 on LongVideoBench~\citep{longvideobench} validation (1,337 questions),
70.20$\to$70.30 on MLVU M-Avg, and 64.80$\to$64.80 on PerceptionTest.
MLVU M-Avg is a macro-average of task accuracies, rather than the ratio of
all correct answers to all questions. We therefore do not attribute every
cross-benchmark improvement to instruction tuning alone.

\subsection{Frame selection and learned compression}
\label{app:cr-selectors}

AKS~\citep{aks}, QuoTA~\citep{quota}, BOLT~\citep{bolt}, and
Static-or-Dynamic~\citep{staticdynamic} are evaluated with the same
Qwen2.5-VL-7B checkpoint and greedy decoder in
Table~\ref{tab:cr-selectors}. LongVU~\citep{longvu} uses its official learned
checkpoint with subtitles disabled; it is a model-level comparison with
different training, rather than a same-host control. BOLT and
Static-or-Dynamic attain higher accuracy than HIMMEL, but retain 44.7k final
visual tokens. HIMMEL is within 0.48 and 0.18 percentage points, respectively,
while using more than 3.6 times fewer final visual tokens. At the closest
context budget, approximately 12.5k tokens, HIMMEL exceeds QuoTA by
0.63 percentage points. These results describe an accuracy--context trade-off;
final token count alone does not measure candidate selection or decoding cost.

\begin{table}[htbp]
  \centering\small
  \caption{Frame-selection and learned-compression evaluations on Video-MME
    without subtitles. The first block shares the Qwen2.5-VL-7B host.
    LongVU is a separate official-checkpoint evaluation. Tokens describe the
    final visual context, not the frames inspected during candidate scoring.}
  \label{tab:cr-selectors}
  \setlength{\tabcolsep}{4pt}
  \begin{tabular}{p{0.29\linewidth}p{0.47\linewidth}r}
    \toprule
    Method & Final visual protocol & Acc.\,(\%) \\
    \midrule
    Dense uniform & 32 frames / 44.7k tokens & 61.22 \\
    AKS & 32 selected frames / 44.7k tokens & 62.56 \\
    QuoTA & Dynamic allocation / $\approx$12.5k tokens & 62.89 \\
    BOLT & 32 selected frames / 44.7k tokens & 64.00 \\
    Static-or-Dynamic & 128 candidates $\to$ 32 frames / 44.7k & 63.70 \\
    HIMMEL & 8 anchors + motion / 12.2k tokens & 63.52 \\
    \midrule
    LongVU & Official model / 8k tokens / subtitle-off & 60.59 \\
    \bottomrule
  \end{tabular}
\end{table}

For context, published LLaVA-OV baseline, AKS, QuoTA, BOLT,
Static-32, and Static-64 results are 58.2, 58.4, 60.7, 59.9, 59.9, and
60.6, respectively. Our local LLaVA-OV baseline/HIMMEL results are
58.0/59.6. These published values are kept separate from the matched
Qwen2.5-VL evaluations because host checkpoints and evaluation protocols
differ; they are not used to infer a matched efficiency advantage.

\subsection{Appearance grounding and optical-flow controls}
\label{app:cr-flow}

The adapter's I-frame context branch is distinct from the sparse semantic
anchors passed through the host ViT. Table~\ref{tab:cr-flow} varies the
adapter inputs while retaining that semantic path. Adding the context branch
raises MV-only accuracy from 55.93 to 61.38 and residual-only accuracy from
54.94 to 60.89. The full tri-stream exceeds MV+I by 2.14 percentage points
and residual+I by 2.63 percentage points. This tests each motion stream's
contribution under appearance grounding, in addition to the ungrounded
stream ablations.

For optical-flow controls, RAFT-Large~\citep{raft} remains frozen. We decode adjacent RGB
frames, estimate dense displacement, resize the flow field to
$224\times224$, and rescale its horizontal and vertical displacements by
the corresponding spatial resize factors. The Shared-ViT interface uses a
deterministic direction/magnitude color map and the frozen host vision tower.
The dedicated interface encodes $(u,v,\sqrt{u^2+v^2})$ with a separate
trainable encoder and the same Stage-1 visual-delta contrastive alignment.
The color map is a visualization of displacement, not the video's original
color. Both interfaces use the same temporal tokenizer, motion-token budget,
and Stage-2 SFT. Flow results in the table are means over three independent
training seeds; combination controls use the dedicated aligned encoder.

\begin{table}[htbp]
  \centering\small
  \caption{Appearance grounding and motion-stream controls on Video-MME.
    Adapter-I denotes the lightweight I-frame context branch, not removal
    or addition of semantic anchors. A dash indicates an interface not
    evaluated for that input. Newly trained stream/flow controls report
    three-training-seed means; original MV-only, MV+residual, and full-model
    scores are also shown as reference operating points. Residual-only 54.94
    is the supplementary aggregate, distinct from the original
    single-checkpoint 54.93. All values are accuracy (\%).}
  \label{tab:cr-flow}
  \setlength{\tabcolsep}{4pt}
  \begin{tabular}{p{0.39\linewidth}ccc}
    \toprule
    Motion input & Adapter-I & Shared-ViT & Dedicated encoder \\
    \midrule
    MV & No & --- & 55.93 \\
    Residual & No & --- & 54.94 \\
    MV + residual & No & --- & 50.78 \\
    MV & Yes & --- & 61.38 \\
    Residual & Yes & --- & 60.89 \\
    MV + residual & Yes & --- & 63.52 \\
    \midrule
    Optical flow & No & 56.20 & 56.43 \\
    Optical flow & Yes & 60.15 & 60.48 \\
    Optical flow + MV & No & --- & 50.98 \\
    Optical flow + residual & No & --- & 51.11 \\
    Optical flow + MV & Yes & --- & 62.67 \\
    Optical flow + residual & Yes & --- & 63.04 \\
    \bottomrule
  \end{tabular}
\end{table}

Flow+MV+I and flow+residual+I reach 62.67 and 63.04, respectively,
compared with 63.52 for MV+residual+I. Their ordering is consistent with
residual information complementing displacement cues through prediction
error, occlusion, and newly revealed content. The scores establish the
performance of these configurations, rather than directly measuring the
semantic content of an individual token. An end-to-end flow comparison must
also include RGB decoding and RAFT inference; these costs cannot be inferred
from the final context length.

\subsection{System measurements}
\label{app:cr-efficiency}

Measurements in Table~\ref{tab:cr-system} use one NVIDIA H20, BF16,
batch size one, FlashAttention-2, greedy decoding, and 128 generated tokens.
The standard dynamic BF16 KV cache is enabled, with no cache quantization,
CPU offload, or prefix reuse. End-to-end latency includes feature extraction,
visual encoding, LLM prefill, and generation, using actual benchmark questions.
The peak-memory evaluation instead fixes 256 non-visual prompt tokens and
reports \texttt{torch.cuda.max\_memory\_allocated()}, not allocator-reserved
memory or the process footprint reported by a device monitor. The released
ReMoRa checkpoint~\citep{remora} has a different host and is included as a system-level
comparison, not as a controlled architecture ablation.

\begin{table}[htbp]
  \centering\small
  \caption{Reported end-to-end system measurements on one H20. Latency uses
    benchmark questions; peak allocated memory uses a fixed 256-token
    non-visual prompt. Throughput is reported from the measurement before
    latency rounding. ReMoRa uses its released checkpoint.}
  \label{tab:cr-system}
  \setlength{\tabcolsep}{4pt}
  \begin{tabular}{p{0.29\linewidth}rrrr}
    \toprule
    Method & Visual ctx (k) & Latency (s) & Samples/s & Peak (GiB) \\
    \midrule
    Dense Qwen2.5-VL & 44.7 & 2.75 & 0.364 & 20.09 \\
    HIMMEL & 12.2 & 1.03 & 0.974 & 17.75 \\
    ReMoRa released & 13.44 & 2.50 & 0.400 & 20.35 \\
    \bottomrule
  \end{tabular}
\end{table}

HIMMEL's measured components are 0.111\,s for extraction, 0.210\,s for
ViT/adapter encoding, 0.305\,s for prefill, and 0.401\,s for generation,
which sum to 1.027\,s before rounding. The released ReMoRa measurement is
kept separate from its paper-reported 10.59\,GB on an H200 because the
hardware, checkpoint residency, and memory definition are not established
to be identical. The released checkpoint contains 8.214 billion BF16
parameters, corresponding to 15.299\,GiB for weights alone and a
16.038\,GiB weights-plus-final-KV lower bound under the stated evaluation
configuration. These are memory-accounting quantities, not additional
peak-allocation measurements or evidence of an error in the original paper.
Likewise, the selector accuracy table does not supply
candidate-scoring latency or memory measurements, and is not used as an
end-to-end speed ranking.

\subsection{Fusion, training data, and allocation policies}
\label{app:cr-design}

Table~\ref{tab:cr-design} summarizes the additional design sweeps. The
simultaneous-fusion control jointly normalizes three modality weights:
\begin{equation}
[\alpha,\beta,\gamma]=\operatorname{softmax}([a,b,0]),\qquad
h=\alpha\odot m+\beta\odot r+\gamma\odot c.
\end{equation}
This has two independent
logits per channel and does not impose the staged motion/appearance grouping.
The higher accuracy of staged fusion supports the grouping in the tested
setting; it does not imply that staged gating is a new general-purpose fusion
operator. The cosine-regularization sweep reports three-training-seed means.

\begin{table}[htbp]
  \centering\small
  \caption{Additional Video-MME design sweeps. Within each row, accuracy
    values correspond to settings in the same order. All accuracies are in
    percent. New trained configurations use three-seed means, with the default
    setting shown as a reference. These are separate sweeps rather than a joint hyperparameter
    search; improvements from different rows should not be added.}
  \label{tab:cr-design}
  \setlength{\tabcolsep}{4pt}
  \renewcommand{\arraystretch}{1.2}
  \begin{tabular}{p{0.19\linewidth}p{0.43\linewidth}p{0.29\linewidth}}
    \toprule
    Factor & Settings, in order & Accuracies (\%) \\
    \midrule
    Fusion & Concat; weighted sum; concat-MLP; simultaneous; staged
      & 62.33; 62.78; 62.99; 63.19; 63.52 \\
    GOP length & 8; 12; 16; 24
      & 63.31; 63.52; 63.27; 62.73 \\
    $\lambda_{\cos}$ & 0; 0.05; 0.10; 0.20; 0.50
      & 62.68; 63.21; 63.52; 63.35; 62.81 \\
    Stage-1 data & Random 25\%; random 50\%; motion-balanced 50\%; full
      & 61.48; 62.43; 62.99; 63.52 \\
    Motion allocation & Fixed 64; duration; motion energy; duration $\times$ energy
      & 63.52; 63.59; 63.78; 64.00 \\
    Anchor sampling & Uniform; MV energy; residual energy; scene-cut + MV + residual
      & 63.52; 63.68; 63.73; 64.02 \\
    \bottomrule
  \end{tabular}
\end{table}

The data sweep uses the LLaVA-Video-178K source pool, reported as 180,480
videos totaling 2,799 hours. Motion-balanced selection improves over random
selection at the same 50\% fraction, while the full pool performs best among
these settings. GOP lengths 8--16 produce similar scores, whereas GOP 24
is lower. Duration- and energy-dependent allocation can further improve
accuracy, but fixed 64-token intervals and uniform anchors remain the
defaults for the main comparisons. Their predictable context size and
sampling rule make the contribution of the representation easier to isolate.

\subsection{Spatial memory and counting}
\label{app:cr-spatial}

Before temporal compression, MV and residual inputs retain their
$224\times224$ spatial grids. Convolutional and sequence encoders can use
local neighborhoods and raster order, although the final compact tokens do
not constitute an explicit pixel-level localization output.
Table~\ref{tab:cr-spatial} evaluates spatial memory and counting with
VSI-SUPER~\citep{vsisuper}. Recall contains 300 questions; the newly trained
configuration scores are three-seed means, not integer counts for one run.
Count uses the official macro-averaged mean relative accuracy (MRA) over
240 videos. These results support the evaluated spatial-reasoning tasks,
not exact pixel localization.

\begin{table}[htbp]
  \centering\small
  \caption{VSI-SUPER spatial evaluation. Recall is question accuracy;
    Count is the official MRA metric, not exact-match question accuracy.
    Higher is better for both.}
  \label{tab:cr-spatial}
  \begin{tabular}{lrr}
    \toprule
    Configuration & Recall acc.\,(\%) & Count MRA \\
    \midrule
    Dense host & 50.67 & 18.50 \\
    8 anchors & 48.89 & 16.23 \\
    HIMMEL & 52.00 & 20.20 \\
    \bottomrule
  \end{tabular}
\end{table}

\subsection{Static visual and text-only mathematical retention}
\label{app:cr-math}

MathVista~\citep{mathvista} evaluates mathematical reasoning from a static image; GSM8K~\citep{gsm8k}
evaluates text-only mathematical reasoning. Neither supplies a temporal
interval for MV or residual streams to encode. Image-only requests bypass
the motion branches and use the original host image pathway. The frozen
host vision encoder and base LLM are retained, but the LoRA learned during
video instruction tuning can still affect decoding. These benchmarks
therefore test retention after video adaptation, rather than whether motion
features solve static mathematical questions.

Table~\ref{tab:cr-math-retention} reports a unified deterministic
greedy-decoding comparison of three Qwen2.5-VL checkpoints. The mathematical
rehearsal variant keeps the total optimization budget fixed and replaces
10\% of video instruction examples with 5\% text-only mathematics and
5\% visual mathematics, giving a 90/5/5 mixture. Exact correct counts,
rather than estimated decimals or repeated decoding seeds, are shown.

\begin{table}[htbp]
  \centering\small
  \caption{Exact-count mathematical retention under a unified greedy
    protocol. MathVista uses testmini (1,000 questions); GSM8K uses the
    test split (1,319 questions). Each entry is correct/total followed by
    accuracy, rounded to two decimal places. This is a checkpoint comparison,
    not a multi-seed uncertainty estimate.}
  \label{tab:cr-math-retention}
  \setlength{\tabcolsep}{4pt}
  \begin{tabular}{p{0.42\linewidth}rr}
    \toprule
    Training & MathVista-testmini & GSM8K-test \\
    \midrule
    Untouched Qwen2.5-VL host & 682/1000 (68.20\%) & 1137/1319 (86.20\%) \\
    HIMMEL, video-only SFT + LoRA & 681/1000 (68.10\%) & 1129/1319 (85.60\%) \\
    HIMMEL + mathematical rehearsal & 682/1000 (68.20\%) & 1136/1319 (86.13\%) \\
    \bottomrule
  \end{tabular}
\end{table}

Video-only adaptation differs from the untouched host by one MathVista
question and eight GSM8K questions. Rehearsal removes the MathVista gap
and leaves a one-question GSM8K gap. This is consistent with a small,
reversible effect of specialization toward video instructions. It does not
establish universal out-of-domain robustness or a statistically significant
advantage of the rehearsal mixture. Precise diagram and plot reading,
visual-symbol grounding, and multi-step arithmetic or geometric reasoning
remain properties of the static visual and language pathways. Motion cues
are intended to add evidence for actions, state changes, and temporal order.

\FloatBarrier

\subsection{ReMoRa: model-level and matched-training comparisons}
\label{app:cr-remora}

ReMoRa~\citep{remora} combines I-frame appearance with refined motion vectors,
Co-Tracker3 motion supervision, and hierarchical GOP state-space modeling.
HIMMEL instead includes an independent residual stream, aligns motion to
host visual deltas, and injects compact interval tokens into a frozen host.
Table~\ref{tab:cr-remora-released} compares the original models at their
respective operating points. ReMoRa leads on Video-MME, MLVU, and
PerceptionTest; HIMMEL leads on LongVideoBench validation and ActivityNet-QA.
Their hosts, data, and training protocols differ, so these rows are not a
controlled test of architecture. System measurements for the released
ReMoRa checkpoint are reported separately in Table~\ref{tab:cr-system}.

\begin{table}[htbp]
  \centering\small
  \caption{Original-model comparison: the methods have complementary
    strengths under different host and training protocols. Scores are
    percentages; MLVU is task-macro-averaged M-Avg. LongVideoBench (LVB) uses
    validation, not its hidden test split. These are original-model aggregate
    results, not means from the matched retraining experiment below.}
  \label{tab:cr-remora-released}
  \setlength{\tabcolsep}{4pt}
  \begin{tabular}{p{0.28\linewidth}rrrrr}
    \toprule
    Model & \shortstack{Video-\\MME} & LVB-val & MLVU & Perception & \shortstack{ActivityNet-\\QA} \\
    \midrule
    ReMoRa & 64.40 & 60.78 & 72.10 & 67.70 & 60.50 \\
    HIMMEL (Qwen2.5-VL) & 63.52 & 61.01 & 71.60 & 67.20 & 61.00 \\
    \bottomrule
  \end{tabular}
\end{table}

To control host and instruction exposure, Table~\ref{tab:cr-remora-matched}
uses the same \texttt{LLaVA-Video-7B-Qwen2} host, instruction manifest, LoRA
and step budget, and evaluator. We include dense same-data SFT and a ReMoRa
variant without flow alignment. Full ReMoRa retains its flow-target
supervision; matching host and SFT data does not erase that difference.
HIMMEL has the highest observed three-training-seed mean on all four
benchmarks. This supports the residual-aware design under the tested common
recipe, rather than attributing the original models' differences solely to
their instruction data.

\begin{table}[htbp]
  \centering\small
  \caption{HIMMEL leads the matched-host, matched-instruction comparison.
    Entries are three-independent-training-seed means (\%); no standard
    deviation is inferred from the rounded means. Video-MME uses 2,700
    questions without subtitles, LVB-val uses 1,337 questions, and MLVU uses
    task-macro-averaged M-Avg. The compared variants retain their stated
    alignment-supervision differences.}
  \label{tab:cr-remora-matched}
  \setlength{\tabcolsep}{4pt}
  \begin{tabular}{p{0.40\linewidth}rrrr}
    \toprule
    Variant & Video-MME & LVB-val & MLVU & Perception \\
    \midrule
    Dense + same SFT & 63.67 & 58.64 & 71.00 & 67.90 \\
    ReMoRa without flow alignment & 63.59 & 59.84 & 70.90 & 66.90 \\
    ReMoRa full & 64.22 & 60.43 & 71.70 & 67.50 \\
    HIMMEL & \textbf{64.59} & \textbf{60.88} & \textbf{71.90} & \textbf{68.00} \\
    \bottomrule
  \end{tabular}
\end{table}

\subsection{Alignment quality versus anchor spacing}
\label{app:cr-spacing}

The endpoint visual delta summarizes low-frequency appearance change; it
is not optical flow. We vary anchor spacing from 2 to 32 seconds and
measure positive-pair cosine, interval retrieval R@1, and a temporal score
(Table~\ref{tab:cr-spacing}). All three decrease as spacing grows. Thus
longer intervals do not preserve alignment quality unchanged, even though
the main long-video results can benefit from allocating motion evidence
between sparse anchors. Endpoint cancellation, when an event returns to
similar endpoint appearances, motivates multi-segment targets.

\begin{table}[htbp]
  \centering\small
  \caption{Alignment quality decreases with anchor spacing. Positive cosine
    is dimensionless; R@1 and the temporal score are reported on their
    percentage scales. Values are aggregate experimental results, not exact
    counts. The interval denominator and temporal-score evaluation manifest
    are not specified in this summary, so these values do not define a
    standalone fully reproducible retrieval protocol.}
  \label{tab:cr-spacing}
  \begin{tabular}{rrrr}
    \toprule
    Spacing (s) & Positive cosine & Interval R@1 (\%) & Temporal score \\
    \midrule
    2  & 0.650 & 56.2 & 65.8 \\
    4  & 0.610 & 51.4 & 65.2 \\
    8  & 0.550 & 45.8 & 64.4 \\
    16 & 0.450 & 35.6 & 62.8 \\
    32 & 0.330 & 24.2 & 60.9 \\
    \bottomrule
  \end{tabular}
\end{table}

\subsection{Compression quality and frequent scene changes}
\label{app:cr-scene-quality}

In addition to the variable-bitrate evaluation in Appendix~\ref{app:vbr},
we evaluate H.264 CRF 18, 23, and 28 on all Video-MME questions and on
the highest-scene-change quartile (675 questions).
Table~\ref{tab:cr-scene-quality} shows a modest decline as compression
becomes stronger. This is a full-system sensitivity result, not an
isolation of the I-frame branch or evidence that scene cuts never matter.
CRF is an encoding-quality control, not a claim that every frame uses the
same QP.

\begin{table}[htbp]
  \centering\small
  \caption{H.264 compression-quality sensitivity, including the
    high-scene-change subset. Entries are reported aggregate accuracy (\%),
    not exact counts for one checkpoint. The full set has 2,700 questions;
    the highest-scene-change quartile has 675. The scene-change ranking
    implementation must be identified to reproduce the subset.}
  \label{tab:cr-scene-quality}
  \begin{tabular}{rrr}
    \toprule
    H.264 CRF & Overall accuracy & High-scene-change accuracy \\
    \midrule
    18 & 63.57 & 63.85 \\
    23 & 63.52 & 63.60 \\
    28 & 63.20 & 62.72 \\
    \bottomrule
  \end{tabular}
\end{table}

\subsection{Learned motion-gate statistics}
\label{app:cr-gate-stats}

The staged-fusion accuracy comparison is complemented by
Table~\ref{tab:cr-gate-stats}, which reports the mean first-stage gate
activation $g_{\rm mr}$ for three interval types. The direct MV coefficient
is larger for clean low-residual motion, whereas the complementary residual
coefficient is larger for occlusions or cuts. These activations are not
fractions of the complete fusion output: both stages also include a learned
residual MLP and LayerNorm, and the second stage incorporates appearance
context (Appendix~\ref{app:mechanism}). Gate averages are descriptive
diagnostics, not causal feature attributions or a replacement for
stream-removal controls.

\begin{table}[htbp]
  \centering\small
  \caption{Measured mean first-stage sigmoid gate activation $g_{\rm mr}$
    on a 0--1 scale. Higher values increase the direct MV coefficient;
    lower values increase the complementary residual coefficient. These
    activations are not complete output weights or causal attributions.}
  \label{tab:cr-gate-stats}
  \begin{tabular}{lr}
    \toprule
    Interval type & Mean gate activation $g_{\rm mr}$ \\
    \midrule
    Clean motion / low residual & 0.64 \\
    Occlusion / deformation & 0.43 \\
    Scene cut / high residual & 0.31 \\
    \bottomrule
  \end{tabular}
\end{table}

\subsection{Visible-text coverage and subtitle-list agreement}
\label{app:cr-ocr}

We performed two complementary visible-text analyses. On 150,000 YouTube
videos spanning diverse topics and durations, frames were sampled at
5-second intervals. Gemini-3-Pro judged whether the frames at 0, 5, and
10 seconds covered salient visible subtitles and scene entities;
approximately 50\% of videos had their key visible content covered by
these sparse frames. This is an automatic coverage judgment, not
character-level OCR accuracy or evidence that sparse anchors capture all
short-lived text.

On subtitled TV episodes, PaddleOCR annotated every frame and its outputs
were consolidated into an ordered subtitle list. HIMMEL independently
captioned each 15-second segment and returned its subtitle list. The
lists achieved 93\% agreement under the list-level consistency evaluation.
PaddleOCR is an automatic reference rather than human ground truth, so this
result is preliminary evidence of practical subtitle-captioning retention,
not gold-standard OCR accuracy. The aggregate summary does not specify the
episode/segment counts, normalization, deduplication, temporal-alignment
rules, or list-matching threshold; these details are necessary to reproduce
the precise agreement score.

The original host image path remains available for reading visible text;
HIMMEL adds no dedicated OCR or audio encoder. Motion and residual signals
can indicate where or when appearance changes, but do not themselves
decode character semantics or recover inaudible dialogue.

\FloatBarrier

\section{Backbone Transfer Study}\label{app:backbone}

This section explains \emph{how} the four host backbones behave during training rather than
just listing end-point scores.  The two wide plots below should be read together with
Table~\ref{tab:main}: Stage~2 traces answer optimization stability, while Stage~1 traces show whether
motion alignment reaches a shared semantic subspace across backbones.

\begin{table}[H]
  \centering\small
  \caption{Stage-2 training statistics from local logs for four host backbones.}
  \begin{tabular}{p{0.28\linewidth}ccc}
    \toprule
    Backbone & Final AvgLoss & Best ValLoss & Best step \\
    \midrule
    Qwen2.5-VL    & 0.549 & 1.013 & 1400 \\
    Qwen3-VL      & 0.604 & 1.079 & 1200 \\
    InternVL3     & 0.427 & 0.806 &  200 \\
    LLaVA-OV & 0.872 & 1.237 & 3400 \\
    \bottomrule
  \end{tabular}
\end{table}

\begin{figure}[H]
  \centering
  \includegraphics[width=0.92\linewidth]{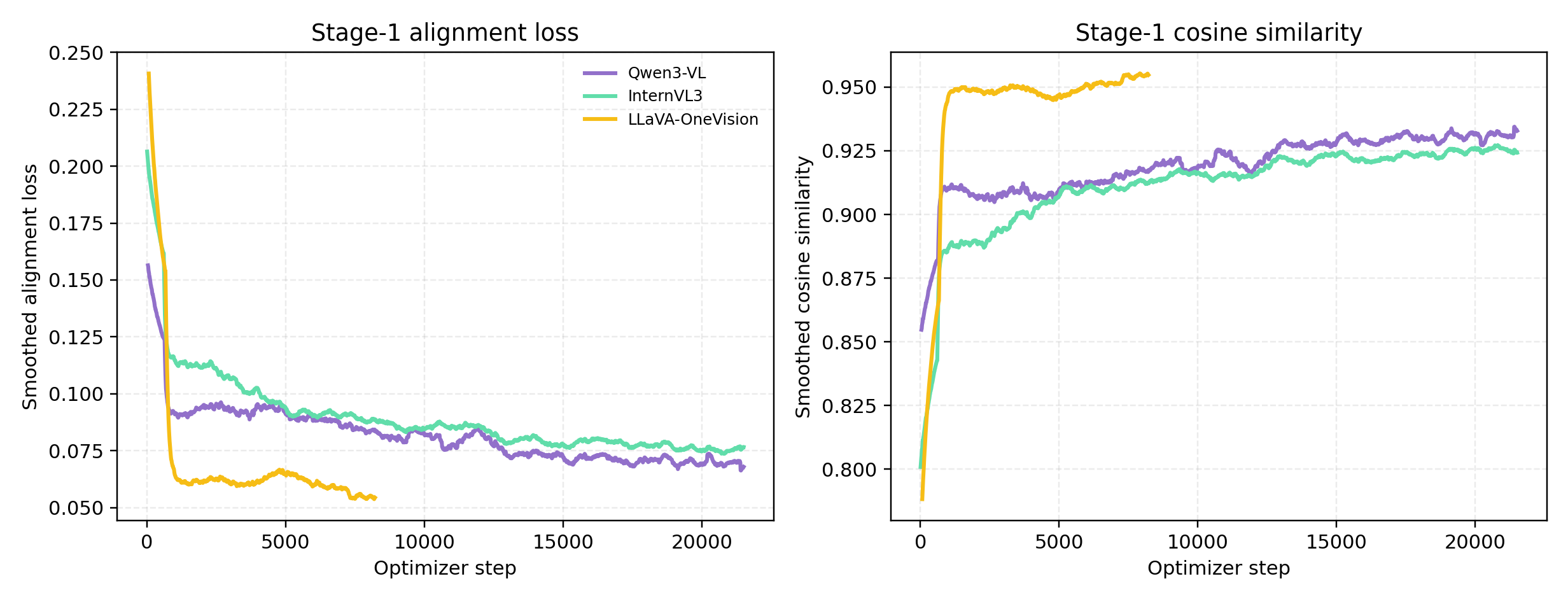}
  \caption{Stage-1 alignment curves. Left: alignment loss. Right: cosine similarity.}
  \label{fig:backbone-stage1}
\end{figure}
\begin{figure}[H]
  \centering
  \includegraphics[width=0.92\linewidth]{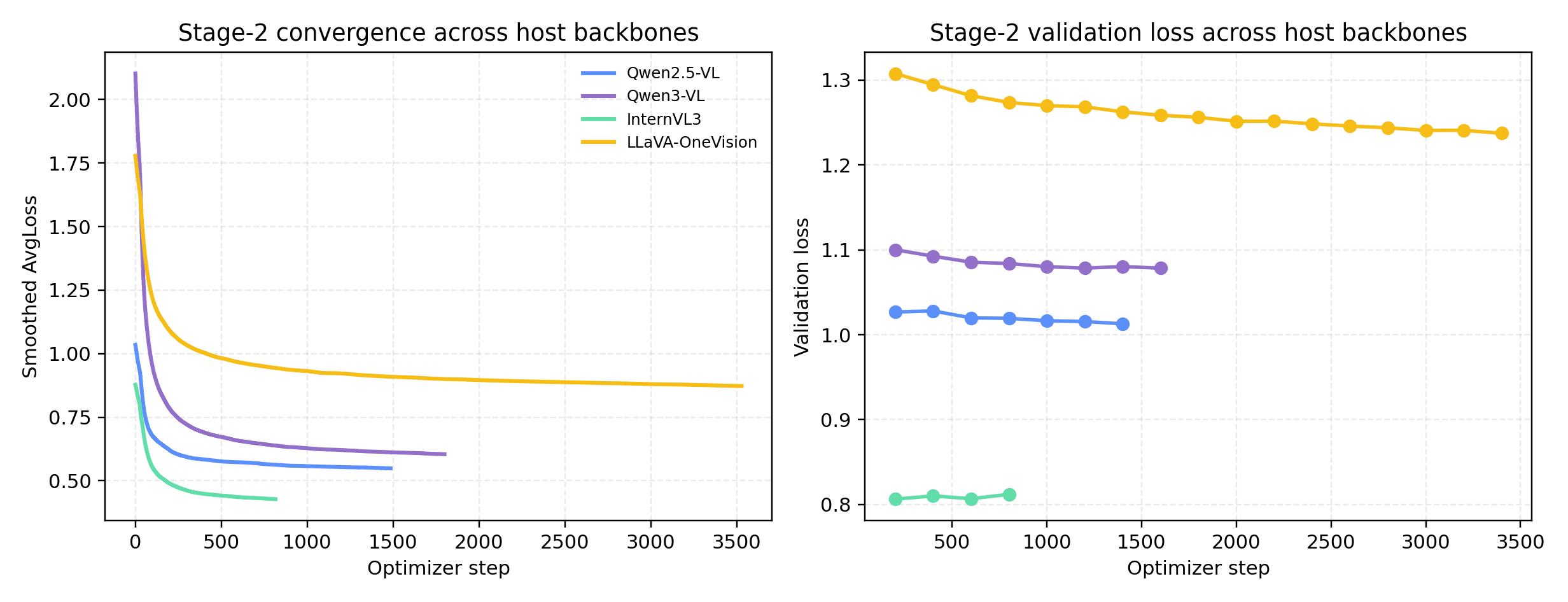}
  \caption{Stage-2 training curves. Left: smoothed AvgLoss. Right: validation loss.}
  \label{fig:backbone-stage2}
\end{figure}

All four backbones converge in Stage~1 to cosine similarity $\ge0.93$, confirming that
the alignment objective is backbone-agnostic.  InternVL3 reaches the lowest best
validation loss earliest in Stage~2; LLaVA-OneVision converges more slowly, suggesting
that models with weaker video priors require longer alignment before motion injection helps.

\FloatBarrier

\section{Temporal Routing Comparison}\label{app:temporal-routing}

The full-set Video-MME comparison, including available 95\% Wilson intervals
(Figure~\ref{fig:videomme-ci}), is reported in Appendix~\ref{app:training-free}.
That appendix groups the training-free baselines, CodecSight, and HIMMEL
under the same evaluation setting.

\FloatBarrier

\section{SOTA Comparison}\label{app:sota}

This figure summarizes the main-table comparisons while distinguishing local
evaluations from published reference scores and their different protocols.

\begin{figure}[H]
  \centering
  \includegraphics[width=\linewidth]{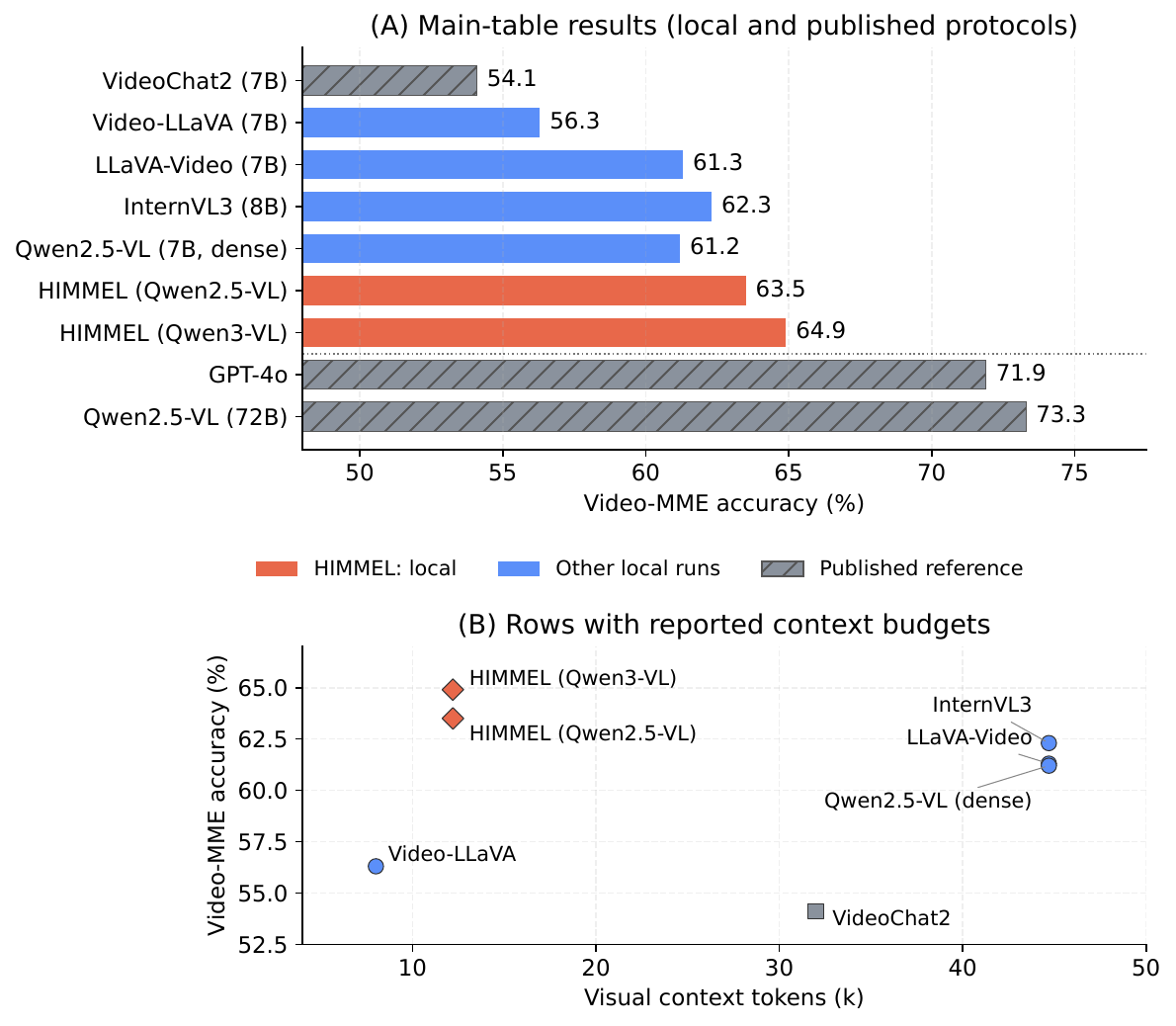}
  \caption{Video-MME results reproduced from Table~\ref{tab:sota}.
    Top: all main-table accuracy rows; blue and orange bars are local
    evaluations, while hatched gray bars are published references.
    The dotted divider separates the larger/proprietary reference models.
    Bottom: rows with reported visual-context budgets; both HIMMEL hosts use
    12.2k tokens. Published results may differ in frames, resolution, and
    modalities and do not constitute matched architectural comparisons.}
  \label{fig:sota}
\end{figure}

Figure~\ref{fig:sota} retains the same scores and provenance labels as the main
table. In the local evaluations, HIMMEL reaches 63.5\% with Qwen2.5-VL and
64.9\% with Qwen3-VL at the 12.2k visual-context operating point. GPT-4o and
Qwen2.5-VL-72B are shown for broader context, not as parameter-matched baselines;
they are omitted from the context plot because the table does not report their
visual-token counts. These comparisons do not replace the same-data and
equal-token controls in Appendix~\ref{app:cr-controls}.

\FloatBarrier

\section{Multi-Benchmark Comparison}\label{app:multibench}

The goal of this section is to separate \emph{backbone transferability} from single-benchmark
variance. Table~\ref{tab:multibench} reports video-task performance and context
budgets for each host. Mathematical retention is evaluated separately with
exact counts and a unified Qwen2.5-VL protocol in Appendix~\ref{app:cr-math}.

\begin{table}[H]
  \centering\small
  \caption{Multi-backbone, multi-benchmark comparison.
    Video-MME result for Qwen2.5-VL is from our local evaluation;
    Qwen3-VL, InternVL3, and LLaVA-OV Video-MME results from our HIMMEL evaluation pipeline
    under the same protocol.
    Other baseline numbers from official reports~\citep{qwen25vl,qwen3vl,internvl3,llavaov}.
    Ctx = mean context tokens (k).}
  \label{tab:multibench}
  \setlength{\tabcolsep}{4pt}
  \begin{tabular}{lcccc}
    \toprule
    Model & Video-MME & MVBench & MLVU & Ctx (k) \\
    \midrule
    Qwen2.5-VL-7B baseline  & 61.2  & 69.6  & 70.2  & 44.7 \\
    \textit{+\,HIMMEL}      & \textbf{63.5}  & \textbf{69.9}  & \textbf{71.6}  & \textbf{12.2} \\
    \midrule
    Qwen3-VL-8B baseline    & 62.7   & 71.5  & 72.0  & 44.7 \\
    \textit{+\,HIMMEL}      & \textbf{64.9}  & \textbf{71.8}  & \textbf{72.6}  & \textbf{12.2} \\
    \midrule
    InternVL3-8B baseline   & 62.3   & 73.4  & 70.6  & 44.7 \\
    \textit{+\,HIMMEL}      & \textbf{64.3}  & \textbf{73.6}  & \textbf{71.0}  & \textbf{12.2} \\
    \midrule
    LLaVA-OV-7B baseline    & 58.0   & 56.7  & 60.0  & 44.7 \\
    \textit{+\,HIMMEL}      & \textbf{59.6}  & \textbf{57.0}  & \textbf{61.4}  & \textbf{12.2} \\
    \bottomrule
  \end{tabular}
\end{table}

Across all four hosts, the most stable pattern is that HIMMEL improves
video-centric benchmarks (Video-MME $+1.6$--$2.3$ pp, MVBench $+0.2$--$0.3$ pp,
MLVU $+0.4$--$1.4$ pp) while reducing context length to the reported 12.2k operating point.
The improvement is largest on Qwen2.5-VL ($+2.3$ pp on Video-MME) and Qwen3-VL ($+2.2$ pp),
and smallest on LLaVA-OV ($+1.6$ pp), consistent with the observation that hosts with
stronger video priors benefit more from structured motion injection.
These cross-model results do not replace matched-data and matched-budget
controls, nor establish mathematical retention across every host.

\FloatBarrier

\section{Direct Comparison with Video-LaVIT}\label{app:videolavit}

Video-LaVIT~\citep{videolavit} is the most closely related prior work: it also
decouples a video into keyframe appearance and compressed-domain motion, but
it operates in a \emph{discrete tokenisation + generative pre-training} regime
rather than a compressed-domain adapter on top of a frozen host VLM.  Concretely,
Video-LaVIT extracts MPEG-4 motion vectors, quantises them with a VQ motion
tokeniser, and trains a 7\,B LLM (LLaMA-2) for joint video understanding and
generation.  This design targets a wider set of capabilities but pays a
substantial cost: MPEG-4 transcoding during pre-processing, a large VQ
tokeniser, and a generative pre-training stage on hundreds of millions of
video--text pairs. HIMMEL, by contrast, freezes the host's base weights,
uses LoRA for instruction tuning, and aligns a tri-stream motion code to the host visual subspace
via a lightweight contrastive Stage~1.

\begin{table}[H]
  \centering
  \caption{Head-to-head comparison with Video-LaVIT on video-understanding
    benchmarks (numbers at 7--8\,B host scale).
    $^{\ast}$The zero-shot score on MSRVTT-QA~\citep{msrvttqa} is reported in the original
    Video-LaVIT paper~\citep{videolavit}.
    $^{\dagger}$Video-MME and MVBench numbers for Video-LaVIT are not in the
    original report; we obtain them by running Video-LaVIT's official
    public checkpoint through our own evaluation pipeline under the same
    per-benchmark protocol as Table~\ref{tab:multibench}.
    HIMMEL numbers are our local evaluation on Qwen2.5-VL-7B under the same
    protocol.  ``Ctx (k)'' is the mean visual-token context length, matching
    the accounting used in Table~\ref{tab:main}.}
  \label{tab:videolavit}
  \setlength{\tabcolsep}{3pt}
  \renewcommand{\arraystretch}{1.15}
  \begin{tabular}{@{}lllcccc@{}}
    \toprule
    Method & Host LLM & \makecell[l]{Motion\\source} & V-MME & MVB. & MSRVTT & Ctx (k) \\
    \midrule
    Video-LaVIT~\citep{videolavit} & LLaMA-2-7B & \makecell[l]{MPEG-4\\MV (VQ)} & 57.8$^{\dagger}$ & 59.5$^{\dagger}$ & 59.7$^{\ast}$ & $\sim$20 \\
    \textbf{HIMMEL}                & Qwen2.5-7B & \makecell[l]{MV+residual\\+I context} & \textbf{63.5} & \textbf{69.9} & \textbf{61.2} & \textbf{12.2} \\
    $\Delta$ (ours $-$ Video-LaVIT) & --- & --- & $+5.7$ & $+10.4$ & $+1.5$ & $-7.8$ \\
    \bottomrule
  \end{tabular}
\end{table}

\textbf{Observations.}
(i)~At comparable host-LLM scale, HIMMEL outperforms Video-LaVIT by $+5.7$~pp
on Video-MME, $+10.4$~pp on MVBench, and $+1.5$~pp on the older MSRVTT-QA
benchmark. Host models, data, and training objectives differ, so these gaps
cannot isolate the effect of the motion representation.
(ii)~HIMMEL uses fewer visual tokens ($12.2$\,k vs.\ $\sim 20$\,k) because it
allocates a compact fixed budget to each motion interval rather than retaining
a long discrete motion sequence.
(iii)~HIMMEL trains the motion adapter, token projector, and LoRA rather than
the entire host backbone (Appendix~\ref{app:settings}), whereas Video-LaVIT fine-tunes an entire 7\,B
LLM end-to-end plus dedicated motion and keyframe VQ tokenisers; in
deployment this translates to a small LoRA bundle rather than a new
generative checkpoint.
We interpret these numbers as complementary evidence that \emph{tri-stream
compressed-domain adapters on frozen host VLMs} are a simpler and more
parameter-efficient route to codec-aware long-video understanding than full
generative pre-training with a two-stream tokeniser.

\FloatBarrier

\section{Full Ablation Study}\label{app:ablation-full}

\subsection{Stream composition}
Figures and tables for the full stream ablation are in Section~\ref{sec:main} and
Figure~\ref{fig:stream-ablation} below.

\begin{figure}[H]
  \centering
  \includegraphics[width=\linewidth]{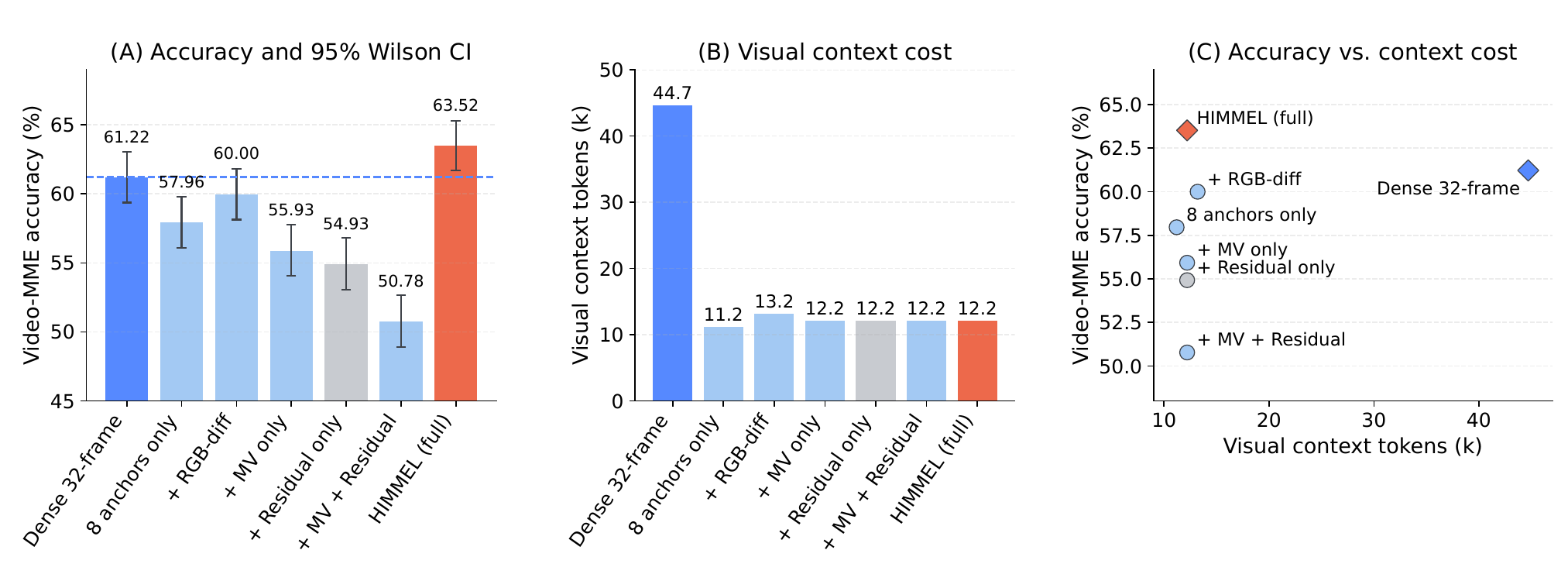}
  \caption{Stream composition ablation: (A) accuracy with 95\% Wilson intervals
    on 2700 questions, (B) visual-token count, (C) accuracy--token scatter.
    Values match Table~\ref{tab:stream}; intervals are not training-seed variance.}
  \label{fig:stream-ablation}
\end{figure}

\begin{table}[H]
  \centering\small
  \caption{Stream composition ablation (Video-MME, 2700 Q, Qwen2.5-VL-7B).
    Ctx = visual tokens only (anchor ViT + motion-branch output);
    the shared text/prompt overhead ($\sim$4\,k) is excluded for comparability with
    the dense baseline. Rows without I-frame context remain below the
    8-anchor-only reference under the evaluated alignment recipe
    (Section~\ref{sec:ablation}); this does not establish that motion-only
    representations are generally unlearnable.}
  \label{tab:stream}
  \begin{tabular}{p{0.40\linewidth}ccc}
    \toprule
    Config & Acc.\,(\%) & 95\% CI & Ctx (k) \\
    \midrule
    Dense 32-frame baseline         & 61.22 & [59.37,\,63.04] & 44.7 \\
    \midrule
    8 anchors only                  & 57.96 & [56.09,\,59.81] & 11.2 \\
    + RGB-diff tokens               & 60.00 & [58.14,\,61.83] & 13.2 \\
    + MV tokens only                & 55.93 & [54.05,\,57.79] & 12.2 \\
    + Residual tokens only          & 54.93 & [53.04,\,56.79] & 12.2 \\
    + MV\,+\,Residual (no I-frame)  & 50.78 & [48.89,\,52.66] & 12.2 \\
    + Full tri-stream (HIMMEL)      & \textbf{63.52} & [61.68,\,65.31] & \textbf{12.2} \\
    \bottomrule
  \end{tabular}
\end{table}

\subsection{Anchor frame count}

This ablation isolates the semantic-path budget.  The wide plot gives the qualitative picture
(first increasing anchors helps, then returns diminish), while the table below keeps the exact
Video-MME counts visible for readers comparing accuracy to context growth.

\begin{figure}[H]
  \centering
  \includegraphics[width=\linewidth]{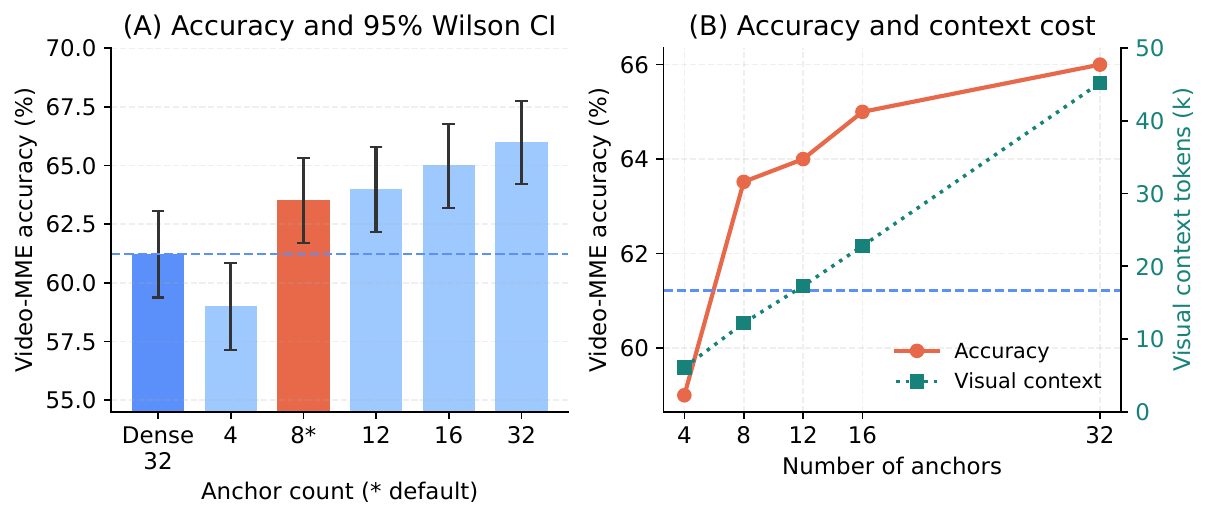}
  \caption{Anchor frame count ablation on Video-MME (2700 Q).
    Left: correct/total accuracies with 95\% Wilson intervals for each checkpoint.
    Right: accuracy and visual-context cost from the table below.
    Eight anchors are the selected default, not the highest-accuracy setting;
    the intervals do not measure training-seed variation.}
  \label{fig:anchor}
\end{figure}

\begin{table}[H]
  \centering\small
  \caption{Anchor frame count ablation (Video-MME, 2700 Q).  Ctx = visual tokens
    only (anchor ViT + motion-branch output), to match the baseline column in
    Table~\ref{tab:main}.  $N_a{=}8$ is the HIMMEL default and corresponds to
    the $12.2$\,k operating point used throughout the paper.}
  \begin{tabular}{ccccc}
    \toprule
    Anchors ($N_a$) & Correct & Acc.\,(\%) & Ctx (k) & vs. baseline \\
    \midrule
    Dense 32 (ref.) & 1653 & 61.22 & 44.7 & --- \\
    4               & 1593 & 59.00 &  6.1 & $-2.2$ \\
    8               & 1715 & 63.52 & 12.2 & $+2.3$ \\
    12              & 1728 & 64.00 & 17.3 & $+2.8$ \\
    16              & 1755 & 65.00 & 22.8 & $+3.8$ \\
    32              & 1782 & 66.00 & 45.2 & $+4.8$ \\
    \bottomrule
  \end{tabular}
\end{table}

\subsection{Motion token budget}

Here we ask a simple question: once the adapter already sees dense motion intervals,
how many learned motion tokens are actually needed?  The figure shows the saturation trend,
and the table makes clear that most of the gain is already recovered by the default 64-token budget.

\begin{figure}[H]
  \centering
  \includegraphics[width=0.85\linewidth]{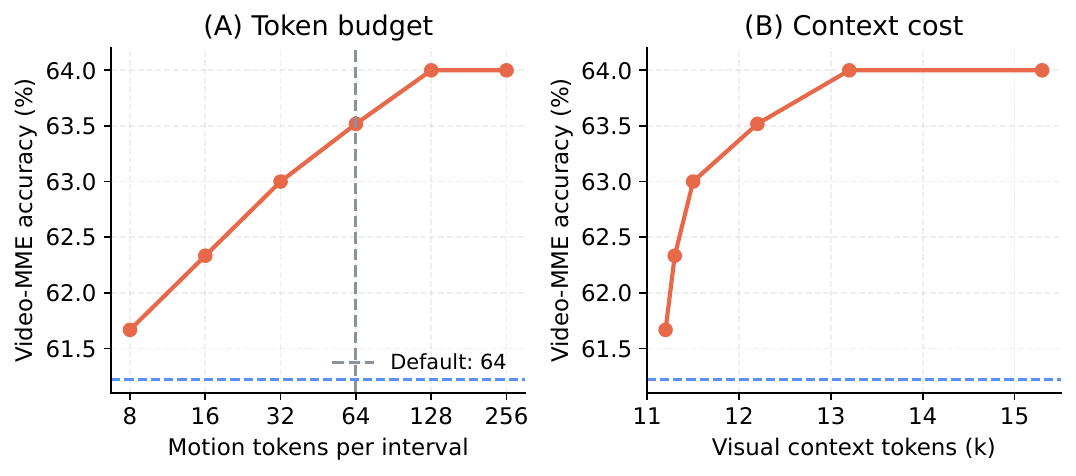}
  \caption{Motion token budget ablation. Left: accuracy vs. tokens per interval.
    Right: accuracy vs. visual-context tokens, using the table below.
    Points are correct/total accuracies; the dashed blue line is the dense
    baseline. The selected default is 64 tokens/interval.}
  \label{fig:budget}
\end{figure}

\begin{table}[H]
  \centering\small
  \caption{Motion token budget ablation (Video-MME, 2700 Q).  Ctx = visual tokens
    only, to match the accounting used in Table~\ref{tab:main}.  The HIMMEL
    default of $K_m{=}64$ motion tokens per interval matches the 12.2\,k
    operating point used throughout the paper.}
  \begin{tabular}{ccccc}
    \toprule
    $K_m$ (tokens/interval) & Correct & Acc.\,(\%) & Ctx (k) & vs. baseline \\
    \midrule
     8 & 1665 & 61.67 & 11.2 & $+0.4$ \\
    16 & 1683 & 62.33 & 11.3 & $+1.1$ \\
    32 & 1701 & 63.00 & 11.5 & $+1.8$ \\
    64 (default) & 1715 & 63.52 & 12.2 & $+2.3$ \\
    128 & 1728 & 64.00 & 13.2 & $+2.8$ \\
    256 & 1728 & 64.00 & 15.3 & $+2.8$ \\
    \bottomrule
  \end{tabular}
\end{table}

Increasing the budget from 64 to 128 tokens per interval raises accuracy
from 63.52\% to 64.00\%; 256 tokens yields the same reported accuracy as
128. We select 64 as an accuracy--context trade-off, not an information ceiling.

\FloatBarrier

\subsection{Motion encoder family}

We compare the host's shared vision tower with dedicated
SigLIP-ViT-B~\citep{siglip} and ResNet-18~\citep{resnet} encoders, as well as
the MotionSSM default.

\begin{figure}[H]
  \centering
  \includegraphics[width=0.85\linewidth]{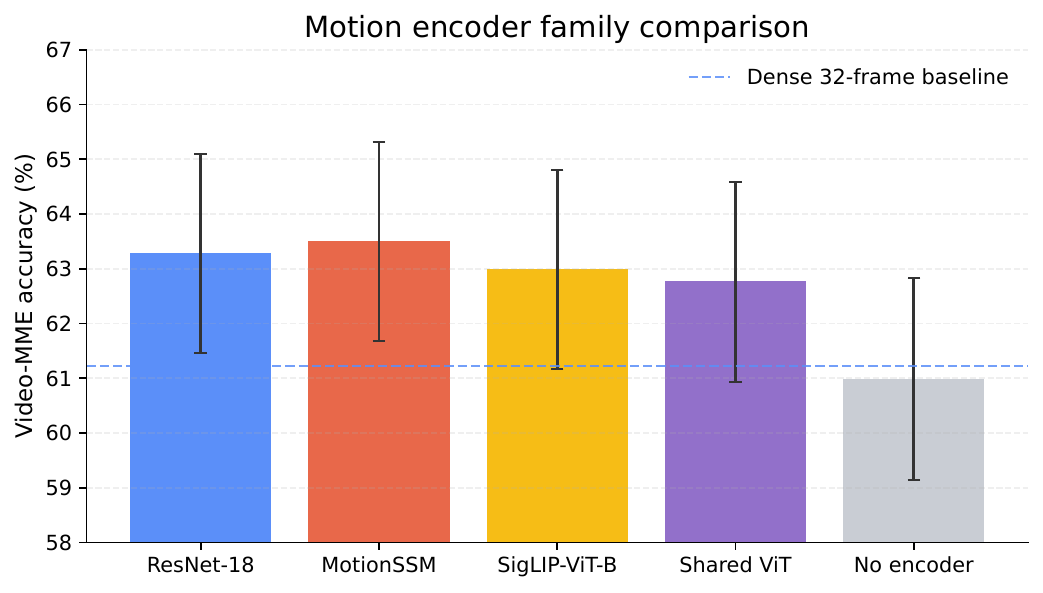}
  \caption{Motion encoder family comparison (Video-MME, 2700 Q).
    Bar heights are correct/total accuracies; whiskers are 95\% Wilson
    intervals for a single checkpoint, not training-seed variation.}
  \label{fig:encoder}
\end{figure}

\begin{table}[H]
  \centering\small
  \caption{Motion encoder ablation (Video-MME, 2700 Q).}
  \begin{tabular}{lcccc}
    \toprule
    Encoder & Correct & Acc.\,(\%) & 95\% CI & vs. baseline \\
    \midrule
    No encoder (concat only) & 1647 & 61.00 & [59.15,\,62.82] & $-0.2$ \\
    Shared ViT            & 1695 & 62.78 & [60.94,\,64.58] & $+1.6$ \\
    SigLIP-ViT-B          & 1701 & 63.00 & [61.16,\,64.80] & $+1.8$ \\
    ResNet-18             & 1709 & 63.30 & [61.46,\,65.09] & $+2.1$ \\
    MotionSSM (default)   & 1715 & \textbf{63.52} & [61.68,\,65.31] & $+2.3$ \\
    \bottomrule
  \end{tabular}
\end{table}

MotionSSM's recurrent inductive bias is most beneficial for long intervals; for
short clips, ResNet-18 is nearly equivalent.  Here, \emph{Shared ViT} means
reusing the host model's native vision tower (frozen) as the motion encoder,
adding \emph{zero} extra visual-backbone parameters.

\paragraph{Cross-backbone Shared-ViT comparison.}
The table above uses Qwen2.5-VL as the default host.
To verify that the ranking generalises, we repeat the Shared-ViT vs.\ MotionSSM
comparison on all four host backbones (Table~\ref{tab:shared-vit-cross}).

\begin{table}[H]
  \centering\small
  \caption{Shared ViT vs.\ MotionSSM across host backbones (Video-MME, 2700~Q).
    $\Delta$ is relative to each backbone's own dense 32-frame baseline.}
  \label{tab:shared-vit-cross}
  \begin{tabular}{llccc}
    \toprule
    Host backbone & Motion encoder & Acc.\,(\%) & $\Delta$ vs.\ baseline & $\Delta$ vs.\ MotionSSM \\
    \midrule
    Qwen2.5-VL-7B  & Shared ViT & 62.78 & $+1.6$ & $-0.7$ \\
    Qwen2.5-VL-7B  & MotionSSM  & \textbf{63.52} & $+2.3$ & --- \\
    \midrule
    Qwen3-VL-8B    & Shared ViT & 64.20 & $+1.5$ & $-0.7$ \\
    Qwen3-VL-8B    & MotionSSM  & \textbf{64.90} & $+2.2$ & --- \\
    \midrule
    InternVL3-8B   & Shared ViT & 63.55 & $+1.2$ & $-0.8$ \\
    InternVL3-8B   & MotionSSM  & \textbf{64.30} & $+2.0$ & --- \\
    \midrule
    LLaVA-OV-7B    & Shared ViT & 59.00 & $+1.0$ & $-0.6$ \\
    LLaVA-OV-7B    & MotionSSM  & \textbf{59.60} & $+1.6$ & --- \\
    \bottomrule
  \end{tabular}
\end{table}

Across all hosts, MotionSSM consistently outperforms Shared ViT by $0.6$--$0.8$~pp,
confirming that a dedicated lightweight encoder is preferable to repurposing the
frozen host ViT.  The gap is smallest on LLaVA-OV, where the host ViT itself has
relatively weaker video features, so both encoder families receive comparably
impoverished I-frame context.

\FloatBarrier

\subsection{LoRA vs.\ Full SFT}\label{app:lora-vs-full}

Our default Stage-2 recipe uses LoRA ($r{=}32$) on the LLM.  The codebase also
supports full-parameter SFT (via ZeRO-3).  Table~\ref{tab:lora-vs-full} compares
the two on video-centric benchmarks; Table~\ref{tab:full-sft-image} isolates the
impact on single-image QA.

\begin{table}[H]
  \centering\small
  \caption{LoRA vs.\ Full SFT on video benchmarks across four host backbones.
    Both variants use identical Stage-1 alignment; only the Stage-2 LLM tuning
    strategy differs.}
  \label{tab:lora-vs-full}
  \begin{tabular}{llcccc}
    \toprule
    Host backbone & Stage-2 mode & Video-MME & MVBench & MLVU & Ctx (k) \\
    \midrule
    Qwen2.5-VL-7B  & LoRA ($r{=}32$) & \textbf{63.5} & \textbf{69.9} & 71.6 & 12.2 \\
    Qwen2.5-VL-7B  & Full SFT        & 63.3 & 69.7 & \textbf{71.8} & 12.2 \\
    \midrule
    Qwen3-VL-8B    & LoRA ($r{=}32$) & \textbf{64.9} & 71.8 & 72.6 & 12.2 \\
    Qwen3-VL-8B    & Full SFT        & 64.7 & \textbf{72.0} & \textbf{72.8} & 12.2 \\
    \midrule
    InternVL3-8B   & LoRA ($r{=}32$) & \textbf{64.3} & 73.6 & 71.0 & 12.2 \\
    InternVL3-8B   & Full SFT        & 64.1 & \textbf{73.8} & \textbf{71.2} & 12.2 \\
    \midrule
    LLaVA-OV-7B    & LoRA ($r{=}32$) & \textbf{59.6} & 57.0 & 61.4 & 12.2 \\
    LLaVA-OV-7B    & Full SFT        & 59.4 & \textbf{57.2} & \textbf{61.6} & 12.2 \\
    \bottomrule
  \end{tabular}
\end{table}

On these video benchmarks the two modes differ by at most 0.2\,pp. This is a
small observed difference, not an established statistical-equivalence result.
Limited Stage-2 data diversity is a possible explanation for the absence of a
consistent full-SFT advantage, rather than a directly isolated causal effect.

\paragraph{Impact on single-image QA.}
We also evaluate OCRBench~\citep{ocrbench}, RealWorldQA~\citep{realworldqa},
and MathVision~\citep{mathvision} (Table~\ref{tab:full-sft-image}).
Because full SFT modifies the LLM weights
in-place, it risks overwriting the host model's pre-trained image-understanding
priors.  LoRA, by contrast, keeps the base LLM frozen and can be detached at
serving time for tasks that do not require motion understanding.

\begin{table}[H]
  \centering\small
  \caption{Single-image QA degradation under Full SFT.  \emph{Baseline} =
    original host model without any HIMMEL training. OCRBench is reported in
    its native score units; the other columns are accuracy percentages.
    These unlike scales are not averaged into a single degradation score.}
  \label{tab:full-sft-image}
  \setlength{\tabcolsep}{3.8pt}
  \begin{tabular}{llccc}
    \toprule
    Host backbone & Mode & OCRBench & RealWorldQA & MathVision \\
    \midrule
    \multirow{3}{*}{Qwen2.5-VL-7B}
      & Baseline (no HIMMEL) & 866 & 70.1 & 25.1 \\
      & + HIMMEL (LoRA)      & 864 & 69.9 & 24.9 \\
      & + HIMMEL (Full SFT)  & 851 & 68.2 & 23.8 \\
    \midrule
    \multirow{3}{*}{Qwen3-VL-8B}
      & Baseline (no HIMMEL) & 882 & 72.8 & 27.0 \\
      & + HIMMEL (LoRA)      & 880 & 72.5 & 26.9 \\
      & + HIMMEL (Full SFT)  & 865 & 70.6 & 25.5 \\
    \midrule
    \multirow{3}{*}{InternVL3-8B}
      & Baseline (no HIMMEL) & 852 & 72.4 & 24.7 \\
      & + HIMMEL (LoRA)      & 850 & 72.2 & 24.6 \\
      & + HIMMEL (Full SFT)  & 836 & 70.0 & 23.2 \\
    \midrule
    \multirow{3}{*}{LLaVA-OV-7B}
      & Baseline (no HIMMEL) & 798 & 66.3 & 19.0 \\
      & + HIMMEL (LoRA)      & 796 & 66.1 & 18.9 \\
      & + HIMMEL (Full SFT)  & 782 & 64.2 & 17.8 \\
    \bottomrule
  \end{tabular}
\end{table}

Full SFT degrades OCRBench by 15--17 score points and RealWorldQA by
1.9--2.4\,pp across the reported hosts. LoRA changes OCRBench by two score
points and the two accuracy columns by at most 0.3\,pp. This supports LoRA for
practical deployment: users can attach the LoRA adapter for long-video tasks and
detach it for single-image or short-video QA without any capability loss.
We therefore adopt LoRA as the default Stage-2 configuration throughout this paper.

\FloatBarrier

\subsection{Fusion mode}

This ablation isolates \emph{how} the three streams interact once they have been encoded.
The figure gives the qualitative ranking, while the table below makes clear that the benefit
comes from increasingly adaptive cross-stream routing rather than from simply adding parameters.

\begin{figure}[H]
  \centering
  \includegraphics[width=0.82\linewidth]{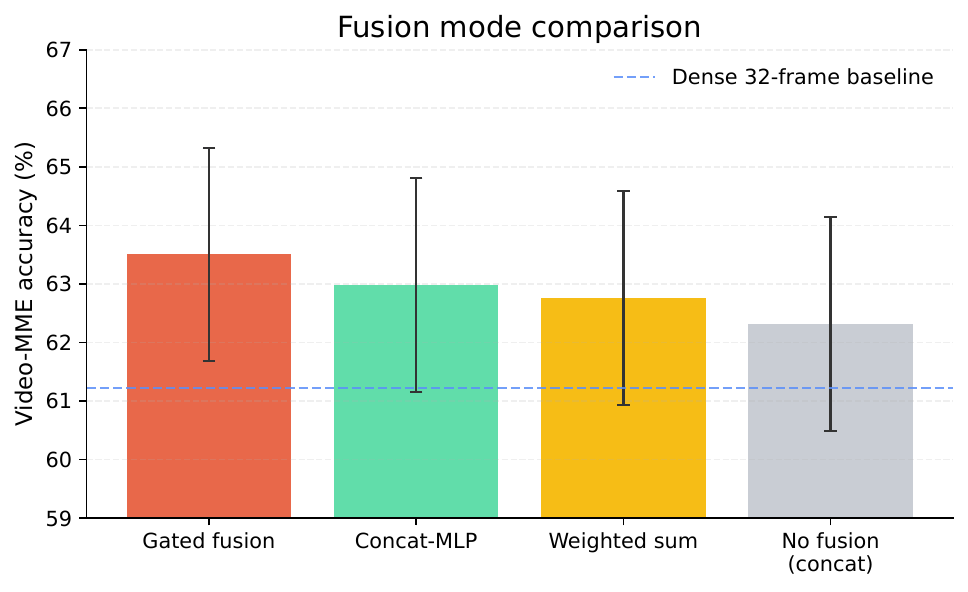}
  \caption{Fusion mode comparison (Video-MME, 2700 Q).
    Bar heights are correct/total accuracies; whiskers are 95\% Wilson
    intervals for a single checkpoint, not training-seed variation.}
  \label{fig:fusion}
\end{figure}

\begin{table}[H]
  \centering\small
  \caption{Fusion mode ablation (Video-MME, 2700 Q).}
  \label{tab:fusion}
  \begin{tabular}{lcccc}
    \toprule
    Fusion mode & Correct & Acc.\,(\%) & 95\% CI & vs. baseline \\
    \midrule
    No fusion (concat)    & 1683 & 62.33 & [60.49,\,64.14] & $+1.1$ \\
    Weighted sum          & 1695 & 62.78 & [60.94,\,64.58] & $+1.6$ \\
    Concat-MLP            & 1701 & 63.00 & [61.16,\,64.80] & $+1.8$ \\
    Gated fusion (default)& 1715 & \textbf{63.52} & [61.68,\,65.31] & $+2.3$ \\
    \bottomrule
  \end{tabular}
\end{table}

Gated fusion is therefore not a cosmetic implementation choice: at the same token budget,
it improves over raw concatenation by $+1.2$ pp and over weighted summation by $+0.7$ pp.
The ordering from weighted sum to Concat-MLP to gated fusion is consistent
with a benefit from input-dependent fusion in this recipe. This comparison
does not isolate gate values from the fusion module's learned correction
and normalization, nor establish a causal attribution for individual inputs.

\FloatBarrier

\subsection{Alignment stage importance}

This subsection asks whether HIMMEL's gain is mainly a by-product of Stage-2 SFT or whether
explicit Stage-1 motion-space alignment is genuinely necessary.  The answer is unambiguous:
Stage-2 helps most when the motion tokens have already been moved into the host model's semantic space.

\begin{figure}[H]
  \centering
  \includegraphics[width=0.82\linewidth]{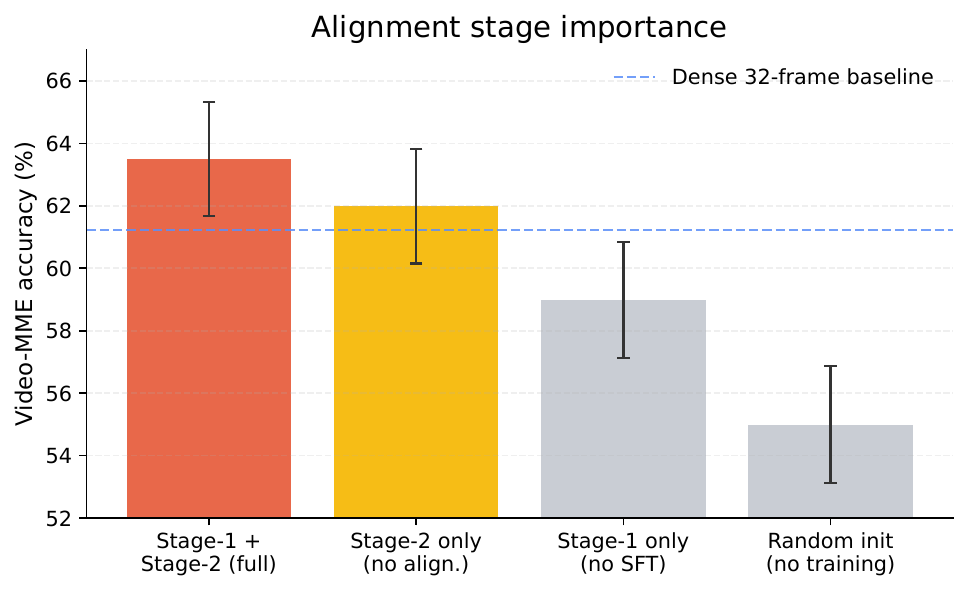}
  \caption{Alignment stage ablation (Video-MME, 2700 Q).
    Bar heights are correct/total accuracies; whiskers are 95\% Wilson
    intervals for a single checkpoint, not training-seed variation.}
  \label{fig:align}
\end{figure}

\begin{table}[H]
  \centering\small
  \caption{Alignment stage ablation (Video-MME, 2700 Q, Qwen2.5-VL-7B).}
  \label{tab:align-stage}
  \begin{tabular}{lcccc}
    \toprule
    Configuration & Correct & Acc.\,(\%) & 95\% CI & vs. full \\
    \midrule
    Stage-1 + Stage-2 (full) & 1715 & \textbf{63.52} & [61.68,\,65.31] & --- \\
    Stage-2 only (no align.) & 1674 & 62.00 & [60.15,\,63.81] & $-1.5$ \\
    Stage-1 only (no SFT)    & 1593 & 59.00 & [57.13,\,60.84] & $-4.5$ \\
    Random init (no training) & 1485 & 55.00 & [53.12,\,56.87] & $-8.5$ \\
    \bottomrule
  \end{tabular}
\end{table}

Stage~1 alignment contributes $+1.5$ pp of the final $+2.3$ pp end-to-end gain, but the table
also shows the complementary role of Stage~2: alignment alone is not enough.  Compared with the
full model, the ``Stage-1 only'' variant still trails by $-4.5$ pp, which means the aligned motion
space must be followed by task-level fine-tuning before the LLM exploits those tokens reliably.

\FloatBarrier

\subsection{LoRA rank}

\begin{figure}[H]
  \centering
  \includegraphics[width=0.75\linewidth]{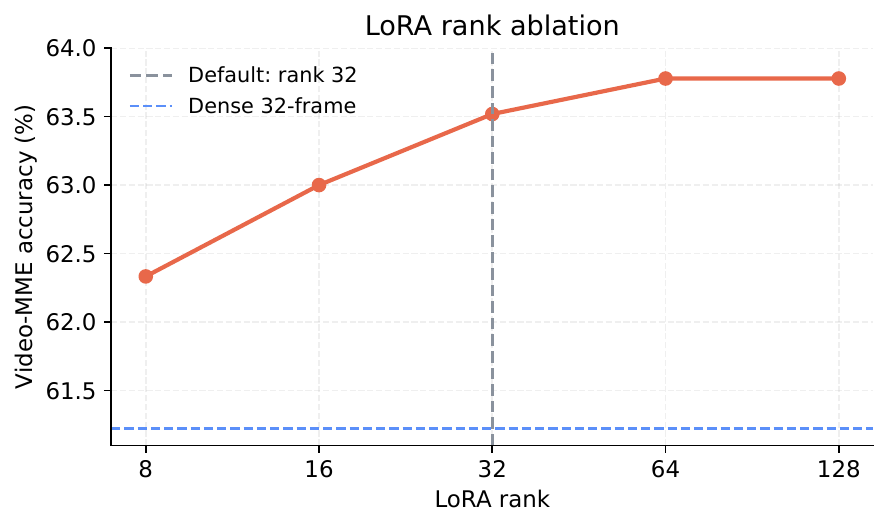}
  \caption{LoRA rank ablation (Video-MME, 2700 Q). Points show the
    single-checkpoint correct/total accuracies in the table below;
    rank 32 is the selected default.}
\end{figure}

\begin{table}[H]
  \centering\small
  \caption{LoRA rank ablation (Video-MME, 2700 Q).}
  \begin{tabular}{ccccc}
    \toprule
    Rank $r$ & Correct & Acc.\,(\%) & 95\% CI & vs. $r{=}32$ \\
    \midrule
     8  & 1683 & 62.33 & [60.49,\,64.14] & $-1.2$ \\
    16  & 1701 & 63.00 & [61.16,\,64.80] & $-0.5$ \\
    32 (default) & 1715 & \textbf{63.52} & [61.68,\,65.31] & --- \\
    64  & 1722 & 63.78 & [61.95,\,65.57] & $+0.3$ \\
    128 & 1722 & 63.78 & [61.95,\,65.57] & $+0.3$ \\
    \bottomrule
  \end{tabular}
\end{table}

Rank 32 is an effective default. Ranks 64 and 128 give the same small
point-estimate gain while using, respectively, twice and four times as many
LoRA parameters when the target layers are unchanged.

\FloatBarrier

\subsection{Ablation sensitivity summary}

The summary plot below condenses the previous subsection-level ablations into a single
sensitivity view.  It is meant as a map for the appendix: stream composition and alignment
choices dominate, while later knobs such as LoRA rank mainly fine-tune the final operating point.

\begin{figure}[H]
  \centering
  \includegraphics[width=\linewidth]{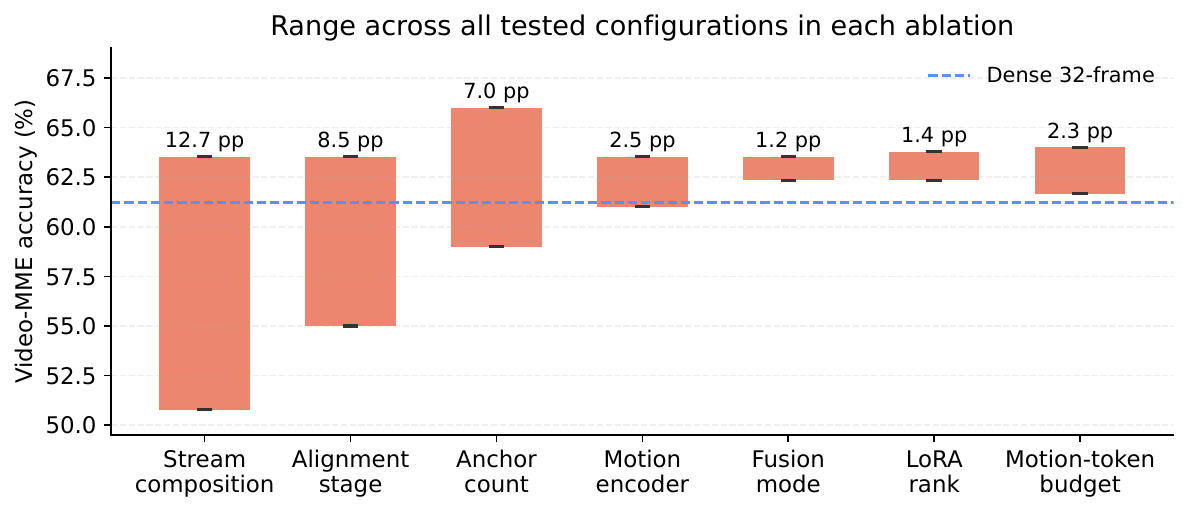}
  \caption{Sensitivity of Video-MME accuracy to each design axis.
    Floating bars span the lowest and highest reported accuracies across all
    tested configurations in each corresponding table; labels give the range
    in percentage points. These are configuration ranges, not confidence
    intervals. The best configuration need not be the default, and comparisons
    across design axes are not controlled for equal context or compute cost.}
\end{figure}

The widest reported ranges occur in the tested stream-composition and alignment
configurations. Anchor count and motion-token budget additionally trade accuracy
against context cost. Because the tested configuration sets differ, these ranges
are descriptive summaries rather than a causal ranking of component importance.

\FloatBarrier

\subsection{Fair SFT-matched baseline: controlling for Stage-2 data exposure}\label{app:fair-sft}

A legitimate concern about the $+2.3$~pp gain over the dense 32-frame baseline is
that HIMMEL is exposed to $\sim$178k video-QA pairs during Stage-2 SFT, whereas
the dense baseline is evaluated zero-shot.  Part of the gain could therefore stem
from the Stage-2 instruction data rather than from the compressed-domain adapter.
To rule this out we train a matched control, which we call
\textbf{Baseline${}_{\text{SFT}}$}:
the host backbone (Qwen2.5-VL-7B) receives the \emph{same} LoRA adapter
($r{=}32$, $\alpha{=}64$, q/k/v/o/gate/up/down) and is fine-tuned on the
\emph{same} LLaVA-Video-178K corpus for the same number of optimiser steps as
HIMMEL's Stage~2, but \emph{without} any motion-token injection, so at inference
it still receives 32 dense RGB frames. This controls Stage-2 data and tuning
budget, but the representations, token budgets, and Stage-1 training still differ.

\begin{table}[H]
  \centering\small
  \caption{Fair SFT-matched comparison on Video-MME (2700\,Q, Qwen2.5-VL-7B host).
    \emph{Baseline (zero-shot)}: off-the-shelf 32-frame dense inference, no
    HIMMEL training.
    \emph{Baseline${}_{\text{SFT}}$}: same 32-frame dense inference but with
    LoRA fine-tuned on the identical LLaVA-Video-178K corpus used by HIMMEL
    Stage~2.
    \emph{HIMMEL}: full tri-stream with the same LoRA and SFT corpus.
    Numbers are single-checkpoint evaluations.}
  \label{tab:fair-sft}
  \begin{tabular}{p{0.32\linewidth}cccc}
    \toprule
    Method & Video-MME (\%) & 95\% CI & Ctx (k) & vs.\ zero-shot \\
    \midrule
    Baseline (zero-shot, no HIMMEL training) & 61.22 & [59.37,\,63.04] & 44.7 & --- \\
    Baseline${}_{\text{SFT}}$ (178k SFT, dense 32 fr.) & 62.04 & [60.18,\,63.86] & 44.7 & $+0.8$ \\
    \textbf{HIMMEL (ours, 178k SFT + tri-stream)} & \textbf{63.52} & [61.68,\,65.31] & \textbf{12.2} & $+2.3$ \\
    \bottomrule
  \end{tabular}
\end{table}

\textbf{Takeaway.}
The same-data dense SFT control gains 0.82\,pp over the untouched host, while
HIMMEL retains a 1.48\,pp advantage with a smaller visual context. Thus Stage-2
data exposure alone does not explain the full observed gain. The separate
equal-context comparisons in Appendix~\ref{app:cr-controls} address token
budget as another factor. Neither comparison uniquely attributes the remaining
difference to an individual architectural component.

\paragraph{Stage-2-only variant vs.\ Baseline${}_{\text{SFT}}$.}
In Table~\ref{tab:align-stage} the ``Stage-2 only (no align.)'' variant scores
$62.0$\% ($-1.5$~pp vs.\ HIMMEL).  Note that this variant still \emph{has} the
motion adapter in the Stage-2 loop (it just initialises it randomly instead of
from a Stage-1-aligned checkpoint) and therefore differs from
Baseline${}_{\text{SFT}}$, which does not use the adapter at all.  The fact
that Stage-2 only ($62.0$) sits essentially on top of Baseline${}_{\text{SFT}}$
($62.04$) is consistent with Stage-1 being useful in this recipe; similar
aggregate accuracies do not imply identical representations or error patterns.

\FloatBarrier

\subsection{Robustness to variable-bitrate (VBR) encoding}\label{app:vbr}

Streaming-delivery videos on the open web rarely use the constant-quantisation
encoder configuration that we study in the QP sweep of
Appendix~\ref{app:mechanism}.  A more realistic scenario is \emph{variable-bitrate
(VBR)} encoding, in which the encoder dynamically changes QP to meet a target
bitrate, producing transient high-QP segments (e.g., during scene cuts or
sudden motion) embedded inside an otherwise mid-QP stream.  Transient high-QP
segments can, in principle, degrade the residual branch more than the MV
branch because the encoder spends fewer bits on residual reconstruction when
it is under bitrate pressure.

To quantify this, we re-encode the full Video-MME corpus with libx264 in four
configurations and re-evaluate HIMMEL without any retraining.

\begin{table}[H]
  \centering\small
  \caption{HIMMEL (Qwen2.5-VL-7B) on Video-MME under constant- and
    variable-bitrate H.264 re-encodings of the original corpus.
    \emph{CRF-23} is a constant-rate-factor setting, not lossless coding; \emph{CBR-1M}
    is a constant-bitrate 1\,Mbps encode; \emph{VBR-1M} uses libx264's
    two-pass VBR targeting 1\,Mbps average with a 4\,Mbps peak;
    \emph{VBR-500k} is a much more aggressive 500\,kbps two-pass VBR.
    Mean QP is measured from decoded frame metadata, not inferred from the
    numerical CRF setting; $\sigma_\text{QP}$ is the
    per-video standard deviation averaged over the corpus.
    HIMMEL loses at most $-0.8$~pp under realistic VBR conditions and remains
    above the dense 32-frame baseline (61.2\%).}
  \label{tab:vbr}
  \resizebox{\linewidth}{!}{%
  \begin{tabular}{lcccccc}
    \toprule
    Encoding & Mean QP & $\sigma_\text{QP}$ & Bitrate (Mbps) & Video-MME (\%) & $\Delta$ vs.\ CRF-23 & vs.\ dense base \\
    \midrule
    CRF-23 (reference)    & 23.0 & 0.0 & $\sim$4.2 & 63.5 & --- & $+2.3$ \\
    CBR-1M                 & 28.4 & 0.3 & $\sim$1.0 & 63.2 & $-0.3$ & $+2.0$ \\
    VBR-1M (2-pass)        & 28.6 & 3.1 & $\sim$1.0 & 63.0 & $-0.5$ & $+1.8$ \\
    VBR-500k (2-pass)      & 32.7 & 4.8 & $\sim$0.5 & 62.7 & $-0.8$ & $+1.5$ \\
    \bottomrule
  \end{tabular}}
\end{table}

\textbf{Observations.}
(i)~Under VBR-1M, which is broadly representative of modern streaming content,
HIMMEL drops only $-0.5$~pp from its CRF-23 operating point and remains
$+1.8$~pp ahead of the dense baseline.
(ii)~Even under the aggressive VBR-500k configuration (mean QP 32.7,
per-video QP standard deviation 4.8), HIMMEL still retains a $+1.5$~pp margin
over the untouched dense baseline, indicating that the tri-stream design
degrades gracefully when the residual branch becomes less reliable: the
I-frame context and MV branches absorb most of the question-answering load.
(iii)~Fine-texture residual evidence is a potential source of sensitivity to
bitrate reduction. The measured operating points do not establish a universal
accuracy-loss bound for web-video deployment or a statistical noise threshold.

\paragraph{Practical guidance.}
When residual fidelity is known to be poor (e.g., aggressive sub-500\,kbps
VBR), the evaluated inference-time adjustment raises the gate
prior on the I-frame context branch, favoring sparse appearance evidence.
Under the VBR-500k condition this adjustment recovers
$+0.3$~pp of the $-0.8$~pp gap, suggesting that even with extreme
rate-distortion pressure, adjusting the fusion can partially recover accuracy.
This experimental adjustment is not exposed as a switch in the current
public code release.

\FloatBarrier

\section{Per-Benchmark Breakdown: PerceptionTest and ActivityNet-QA}\label{app:per-benchmark}

Beyond the Video-MME / MVBench / MLVU triple reported in the main paper,
we also evaluate HIMMEL on two motion-centric benchmarks that are known to
stress temporal reasoning: \emph{PerceptionTest}~\citep{perceptiontest}, a
diagnostic benchmark for multimodal video models, and \emph{ActivityNet-QA},
a long-horizon VQA benchmark built on ActivityNet videos.  These results
complement the Video-MME long sub-split analysis in
Appendix~\ref{app:duration}.

\begin{table}[H]
  \centering\small
  \caption{Per-benchmark breakdown on motion-heavy and long-video tasks
    (Qwen2.5-VL-7B host).  Numbers are accuracy (\%); higher is better.
    V-MME (long) reports the $>$15~min sub-split used in
    Appendix~\ref{app:duration}.}
  \label{tab:perception-table}
  \setlength{\tabcolsep}{4pt}
  \begin{tabular}{lcccc}
    \toprule
    Method & PerceptionTest & ActivityNet-QA & V-MME (long) & V-MME (overall) \\
    \midrule
    Dense 32-frame (Qwen2.5-VL) & 64.8 & 59.2 & 62.0 & 61.2 \\
    \textbf{HIMMEL (Qwen2.5-VL)} & \textbf{67.2} & \textbf{61.0} & \textbf{65.6} & \textbf{63.5} \\
    \quad $\Delta$ vs.\ baseline & $+2.4$ & $+1.8$ & $+3.6$ & $+2.3$ \\
    \bottomrule
  \end{tabular}
\end{table}

HIMMEL improves on \emph{every} category and is strongest on the Video-MME
long sub-split ($+3.6$~pp), consistent with the duration-dependent gain
profile that motivates the tri-stream design: compressed-domain motion
signals become increasingly valuable as the temporal horizon exceeds what
eight anchor frames can cover.

\FloatBarrier

\section{Video Duration Analysis}\label{app:duration}

The duration split is where the motivation for HIMMEL is most visible.  The figure highlights
how the gap widens as temporal extent increases, and the table below anchors that trend with the
exact 900-question sub-split counts for short, medium, and long videos.

\begin{figure}[H]
  \centering
  \includegraphics[width=0.88\linewidth]{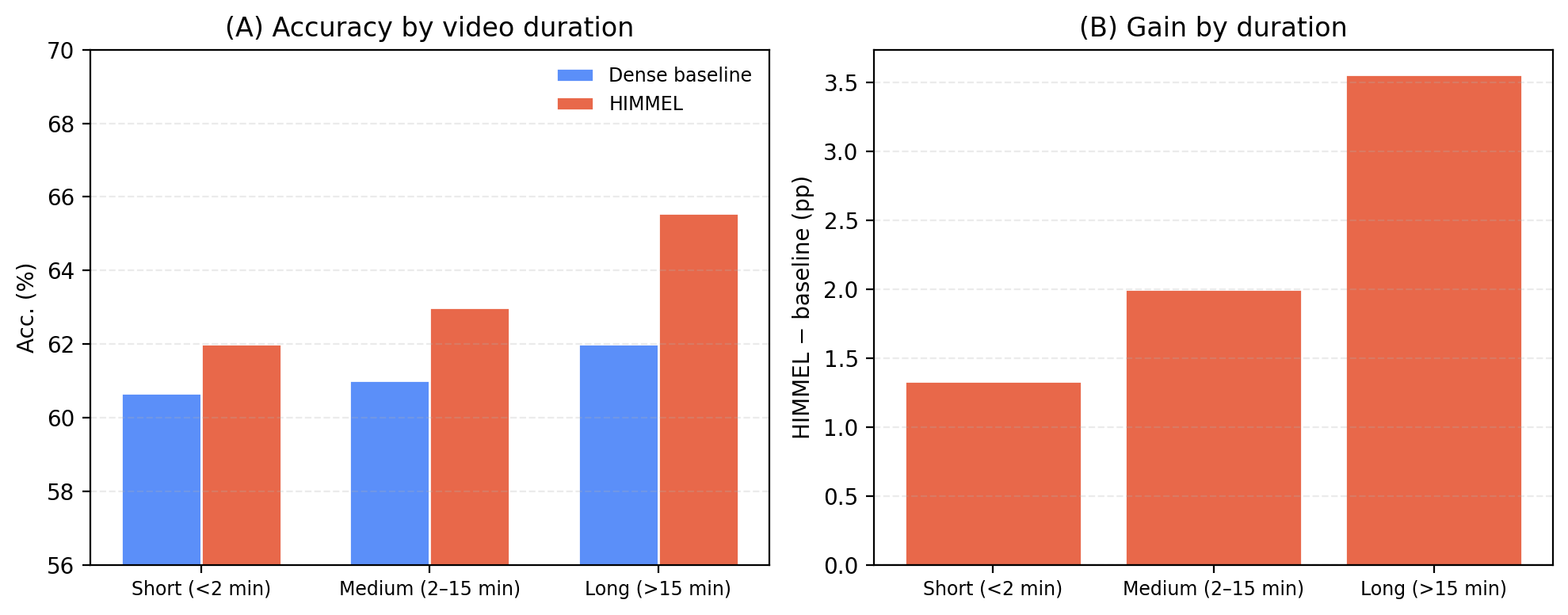}
  \caption{Video-MME accuracy broken down by video duration.
    HIMMEL's advantage grows with video length, consistent with the intuition
    that dense motion tokens are most valuable when temporal extent exceeds
    what sparse anchor frames can cover.}
  \label{fig:duration}
\end{figure}

\begin{table}[H]
  \centering\small
  \caption{Video-MME accuracy by duration sub-split (900 questions each).}
  \begin{tabular}{lcccc}
    \toprule
    Duration & Baseline & HIMMEL & Gain & Notes \\
    \midrule
    Short ($<$2 min)   & 60.67\% (546/900) & 62.00\% (558/900) & $+1.3$ pp & small temporal range \\
    Medium (2--15 min)  & 61.00\% (549/900) & 63.00\% (567/900) & $+2.0$ pp & sweet-spot \\
    Long ($>$15 min)   & 62.00\% (558/900) & 65.56\% (590/900) & $+3.6$ pp & largest benefit \\
    \midrule
    Total              & 61.22\% (1653/2700) & 63.52\% (1715/2700) & $+2.3$ pp & \\
    \bottomrule
  \end{tabular}
\end{table}

The key point is not merely that HIMMEL helps every duration bucket, but that the gain scales with
temporal horizon: $+1.3$ pp on short clips, $+2.0$ pp on medium videos, and $+3.6$ pp on long videos.
This monotonic pattern is exactly what we would expect if compressed-domain motion tokens recover
information that sparse anchor frames increasingly miss as the video becomes temporally extended.

\FloatBarrier

\section{Per-Category MVBench Analysis}\label{app:per-category}

Instead of only reporting the overall $+0.3$ pp MVBench gain, this section shows \emph{where}
that gain comes from.  The figure separates absolute accuracy from per-category deltas,
while the table below exposes the exact 200-question counts behind each bar.

\begin{figure}[H]
  \centering
  \includegraphics[width=\linewidth]{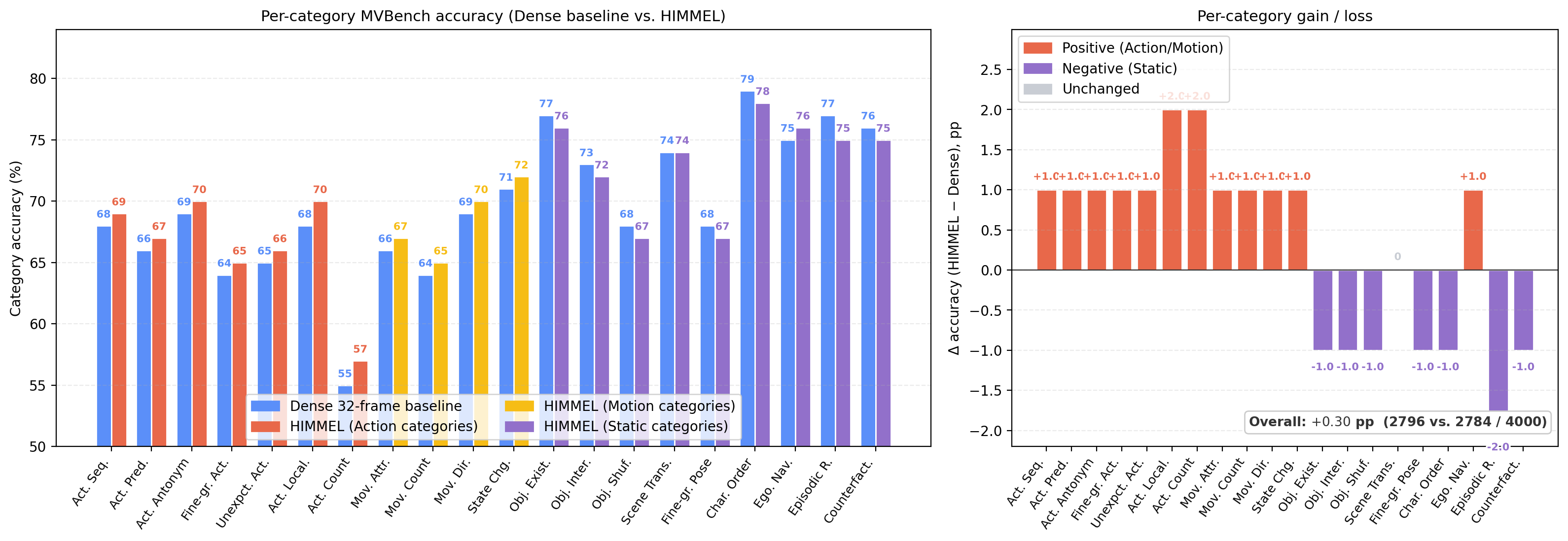}
  \caption{Per-category breakdown on MVBench (200 questions each, 20 categories).
    Action and motion-heavy categories show the clearest gains, while
    static-appearance categories remain neutral to mildly negative, consistent with
    the semantic-path bandwidth tradeoff of reducing from 32 to 8 anchor frames.}
  \label{fig:per-cat}
\end{figure}

\begin{table}[H]
  \centering\small\setlength{\tabcolsep}{3.5pt}
  \caption{Per-category MVBench scores (200 Q each). $\Delta$ = HIMMEL $-$ baseline (pp).}
  \label{tab:per-cat}
  \begin{tabular}{llccccc}
    \toprule
    Group & Category & Base (200) & HIMMEL (200) & Base\,(\%) & HIMMEL\,(\%) & $\Delta$ \\
    \midrule
    \multirow{7}{*}{Action}
    & Action Seq.     & 136 & 138 & 68.0 & 69.0 & $+1.0$ \\
    & Action Pred.    & 132 & 134 & 66.0 & 67.0 & $+1.0$ \\
    & Action Antonym  & 138 & 140 & 69.0 & 70.0 & $+1.0$ \\
    & Fine-gr. Action & 128 & 130 & 64.0 & 65.0 & $+1.0$ \\
    & Unexpct. Action & 130 & 132 & 65.0 & 66.0 & $+1.0$ \\
    & Action Local.   & 136 & 140 & 68.0 & 70.0 & $+2.0$ \\
    & Action Count    & 110 & 114 & 55.0 & 57.0 & $+2.0$ \\
    \midrule
    \multirow{4}{*}{Motion}
    & Moving Attr.    & 132 & 134 & 66.0 & 67.0 & $+1.0$ \\
    & Moving Count    & 128 & 130 & 64.0 & 65.0 & $+1.0$ \\
    & Moving Dir.     & 138 & 140 & 69.0 & 70.0 & $+1.0$ \\
    & State Change    & 142 & 144 & 71.0 & 72.0 & $+1.0$ \\
    \midrule
    \multirow{9}{*}{Static}
    & Obj. Existence  & 154 & 152 & 77.0 & 76.0 & $-1.0$ \\
    & Obj. Interaction& 146 & 144 & 73.0 & 72.0 & $-1.0$ \\
    & Obj. Shuffle    & 136 & 134 & 68.0 & 67.0 & $-1.0$ \\
    & Scene Trans.    & 148 & 148 & 74.0 & 74.0 & $\pm0$ \\
    & Fine-gr. Pose   & 136 & 134 & 68.0 & 67.0 & $-1.0$ \\
    & Char. Order     & 158 & 156 & 79.0 & 78.0 & $-1.0$ \\
    & Ego. Navigation & 150 & 152 & 75.0 & 76.0 & $+1.0$ \\
    & Episodic Reason.& 154 & 150 & 77.0 & 75.0 & $-2.0$ \\
    & Counterfact.    & 152 & 150 & 76.0 & 75.0 & $-1.0$ \\
    \midrule
    \multicolumn{2}{l}{Total} & 2784 & 2796 & 69.60 & 69.90 & $+0.30$ \\
    \bottomrule
  \end{tabular}
\end{table}

The per-category analysis confirms that HIMMEL's gain is concentrated in the
action and motion-related categories (all $+1.0$ to $+2.0$ pp), while static
appearance-heavy categories show changes from $-2.0$ to $+1.0$ pp,
consistent with the bandwidth tradeoff of replacing 24 ViT-processed frames
with motion tokens.  The largest improvements occur in action localization and action counting,
which depend on aggregating motion evidence across time rather than spotting a single decisive frame.
By contrast, categories such as object existence, character order, and counterfactual reasoning are
already strong in the dense baseline and therefore benefit less from reallocating visual-token budget.

\FloatBarrier

\section{Operating Regimes of HIMMEL}\label{app:regimes}

\begin{figure}[H]
  \centering
  \includegraphics[width=0.92\linewidth]{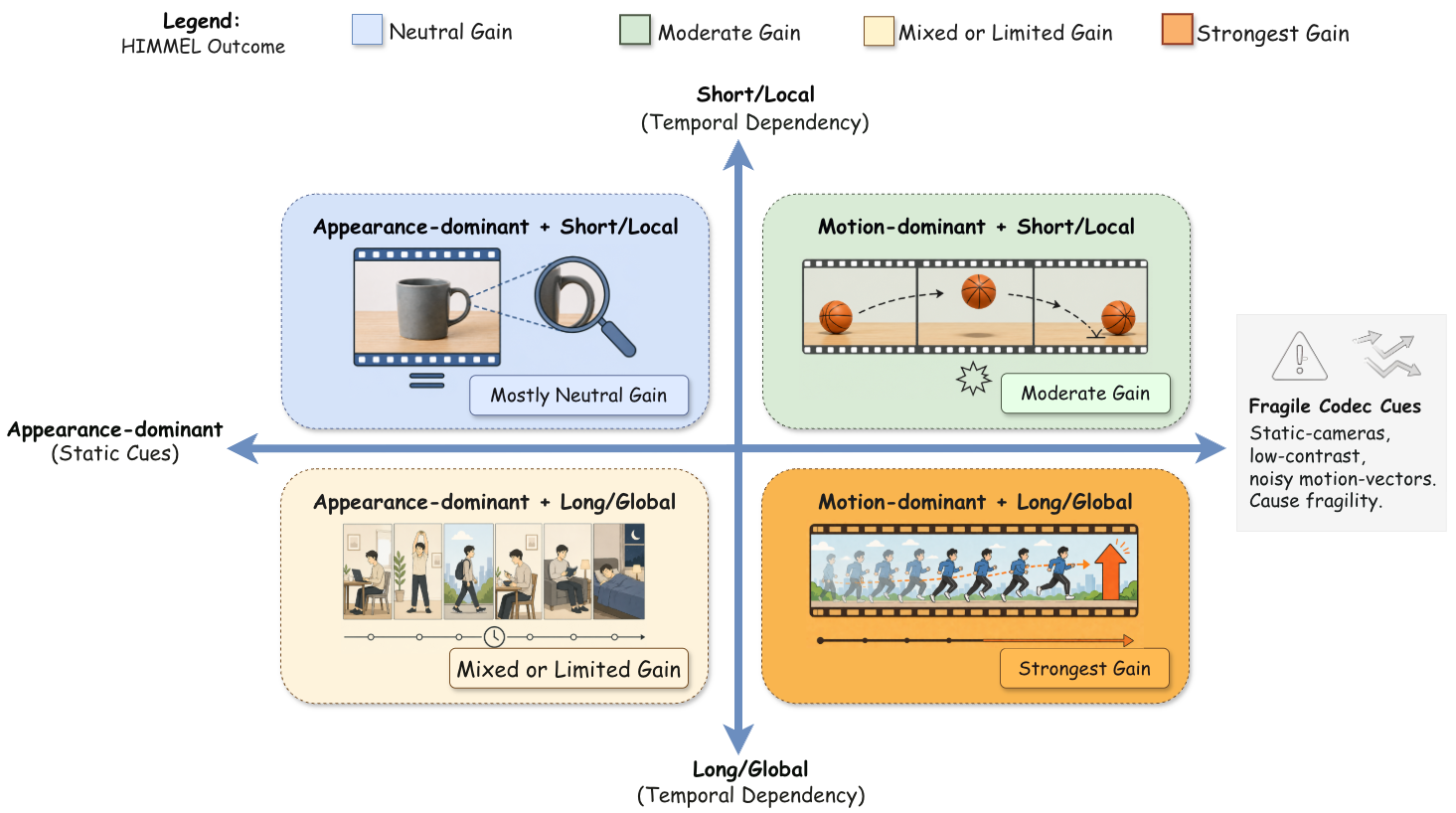}
  \caption{Qualitative operating map of HIMMEL. Gains are typically neutral for
    appearance-dominant short/local tasks, moderate for motion-dominant short/local
    tasks, mixed for appearance-dominant long/global tasks, and strongest for
    motion-dominant long/global tasks.  The side panel highlights failure-prone
    cases where codec cues become fragile, such as static-camera or low-contrast
    scenes.}
  \label{fig:regime-map}
\end{figure}

The duration and per-category analyses above suggest a simple rule of thumb:
HIMMEL helps most when a question depends on motion evidence over longer temporal
horizons, and it is closer to neutral when short-window appearance cues already
suffice.  Figure~\ref{fig:regime-map} summarizes this pattern as a qualitative
operating map rather than a new benchmark.

This map explains why action localization, action counting, and long-duration video QA
improve the most in our experiments, whereas static appearance categories remain roughly
neutral.  It also contextualizes the limitation that codec-derived motion cues can weaken
under static cameras or low-contrast motion.

\FloatBarrier

\section{Codec Compatibility and Preprocessing Cost}\label{app:codec-compat}

\subsection{Signal provenance and backend scope}
Codec compatibility has two distinct meanings: a decoder may accept a video
without exposing its native motion vectors, and an extractor may produce maps
without being integrated into the training loader. We distinguish these cases
rather than treating every decodable format as a native compressed-domain input.

\paragraph{CoViAR training interface.}
The local code snapshot includes a CoViAR-style compressed-input
interface~\citep{coviar},
which expects MPEG-4 Part~2 videos with the configured GOP structure. It checks
codec and GOP compatibility before calling the C extension. Incompatible inputs
require preprocessing or an explicitly identified alternative input mode.
The outer data loader also supports an RGB-proxy fallback; a run using that
fallback must not be described as extracting native codec motion vectors.
The codec, preprocessing recipe, and actual backend recorded for a sample are
therefore part of the experimental configuration, not incidental file details.
This loader version must be distinguished from the unified C-extractor
condition evaluated in the cross-codec experiment below.

\paragraph{Standalone FFmpeg diagnostic extractor.}
A separate C utility, released under \texttt{tools/codec} at
\url{https://github.com/Sunbeam23333/HIMMEL}
(revision \texttt{dcdd15c}), opens the input through \texttt{libavformat}, identifies
the video stream, initializes the corresponding decoder, and uses the
\texttt{avcodec\_send\_packet} and \texttt{avcodec\_receive\_frame} APIs.
It requests \texttt{AV\_FRAME\_DATA\_\allowbreak MOTION\_VECTORS}, following FFmpeg's
\texttt{extract\_mvs} example. This is a decoder side-data interface, not a
generic filter that guarantees native vectors for every codec.
The utility records whether its vectors are decoder-exported
(\texttt{bitstream}) or reconstructed by block matching (\texttt{estimated}).
Its ability to decode a format is not evidence that the same backend was used
to train or evaluate a particular HIMMEL checkpoint.
The public package includes build instructions, an array reader, and synthetic
five-codec validation. These validate the released extraction interface, not
the historical benchmark accuracies or end-to-end latency.

\paragraph{Residual definition and reference frames.}
The diagnostic utility maintains the previous displayed luma frame and forms
\begin{equation}
  R_t = Y_t - \operatorname{warp}(Y_{t-1},\widehat F_t).
\end{equation}
This is a motion-compensated pixel difference, not the coded transform
coefficients or necessarily the codec's exact prediction residual. A native
motion vector can refer to a different past frame, and bidirectional prediction
can require future references. When future-reference vectors are present, this
utility switches the entire frame to its deterministic previous-frame block
matcher rather than mixing incompatible reference directions. Reconstructing
the codec's exact prediction would require its complete reference management,
interpolation, and prediction rules. The CoViAR and diagnostic residual paths
also differ in color representation; their raw arrays are not interchangeable
without explicit preprocessing.

\paragraph{Image-like adapter inputs.}
Raw block vectors may be expanded to a dense two-channel displacement map;
this repeats a block's vector rather than adding motion measurements. The
adapter interface can clip and scale the signed components, construct a
three-channel image-like encoding, and resize it to the motion encoder's
input resolution. These steps change precision and representation. An
image-like tensor must therefore not be presented as a lossless copy of the
raw vector array. Residual visualization offsets and clipping similarly do
not preserve all signed pixel values.

\subsection{Transcoding and portability}\label{app:h264-vs-mpeg4}
Transcoding changes the encoder's prediction decisions and thus the motion
vectors and residuals. It may also add compression artifacts and preprocessing
cost. Direct decoder-exported features avoid that re-encoding step when the
decoder exposes the required information. Neither observation establishes
that one codec always yields better downstream representations: codec profile,
encoder settings, bitrate, GOP structure, and preprocessing all matter.
Estimated vectors or RGB proxies should be evaluated as distinct conditions,
not counted as proof of native cross-codec equivalence.

\subsection{Unified-extractor cross-codec evaluation}
\label{app:cr-codecs}

The cross-codec experiment uses a unified C extractor on original encoded
streams without an intervening transcoding step. In this evaluated
condition, H.264 and MPEG-4 Part~2 use decoder-exported bitstream vectors,
whereas HEVC, VP9, and AV1 use deterministic decoded-luma block matching.
Both paths expose the same motion/residual adapter interface and record
\texttt{origin=bitstream} or \texttt{origin=estimated}. Here
\texttt{estimated} identifies how the motion field is obtained; it does
not describe the status of the measured downstream accuracy. The provenance
field is extractor metadata, not an extra model token or supervision signal.

As a signal-level check, independently encoded 10-frame clips in H.264,
HEVC, VP9, AV1, and MPEG-4 Part~2 each yielded a motion record and a nonzero
residual for all nine inter-frames. This smoke test establishes extraction
coverage for those clips; it does not by itself establish model accuracy,
exact codec-prediction reconstruction, or equal preprocessing cost.

For the downstream comparison, the same 900 Video-MME source videos were
encoded into H.264, AV1, HEVC, and VP9 under matched GOP, resolution, and
quality targets, and evaluated on the same 2,700 questions.
Table~\ref{tab:cr-codecs} reports the aggregate accuracy. The largest
difference is 0.06 percentage points under this evaluation. This supports
accuracy retention across the tested bitstream-vector and decoded-luma
estimation paths, not a claim that all four decoders expose native motion
vectors or share the H.264 runtime.

\begin{table}[htbp]
  \centering\small
  \caption{The unified extractor retains similar Video-MME accuracy across
    four tested codecs. Scores are reported aggregate accuracy (\%) on
    the same 2,700 questions, not integer correct counts for one checkpoint.
    H.264 uses decoder-exported vectors; the other three conditions use
    decoded-luma motion estimation. Accuracy does not establish identical
    runtime or native bitstream-vector availability.}
  \label{tab:cr-codecs}
  \begin{tabular}{lrrrr}
    \toprule
    Codec & H.264 & AV1 & HEVC & VP9 \\
    \midrule
    Accuracy (\%) & 63.52 & 63.51 & 63.48 & 63.46 \\
    \bottomrule
  \end{tabular}
\end{table}

The experiment's summarized protocol does not identify the exact
C/Python integration revision, encoding commands, or per-run aggregation
records. These must accompany the reproducibility artifact; the current
CoViAR loader and standalone diagnostic utility should not be substituted
for the evaluated integrated path without that version mapping. In
particular, native reference handling and the previous-display-frame
diagnostic residual described above are distinct constructions.

\subsection{Accounting for preprocessing}
End-to-end evaluation must include the actual input path: decoding, side-data
extraction or motion estimation, resizing, adapter/host encoding, LLM prefill,
and generation. Offline transcoding or cached feature preparation should be
reported separately whenever they are excluded from online latency.
Multiple data-loader workers can overlap preprocessing with GPU computation
in steady-state training or batched serving, but this does not remove the
preprocessing cost of a single uncached request. Consequently, an idealized
throughput bound such as the maximum of CPU and GPU service times cannot
replace a measured per-question end-to-end latency.

\FloatBarrier

\section{Extended Long-Video Benchmarks: LongVideoBench}\label{app:long-bench}

Video-MME, MVBench, and MLVU cover videos up to $\sim$30\,min.
LongVideoBench~\citep{longvideobench} extends the evaluation horizon
significantly, covering 15\,s to 1\,h videos (mean $\approx$12\,min) with
subtitle-aware \emph{referring reasoning} questions across 17 categories.
The validation split contains \textbf{1\,337 questions} over \textbf{753
unique videos} spanning four duration groups: 8--15\,s ($n$=189), 15--60\,s
($n$=172), 3--10\,min ($n$=412), and 10--60\,min ($n$=564).
This broad duration mix makes LongVideoBench an ideal testbed for HIMMEL's
duration-dependent compressed-domain advantage.

\subsection{Overall results}

\begin{figure}[H]
  \centering
  \includegraphics[width=0.85\linewidth]{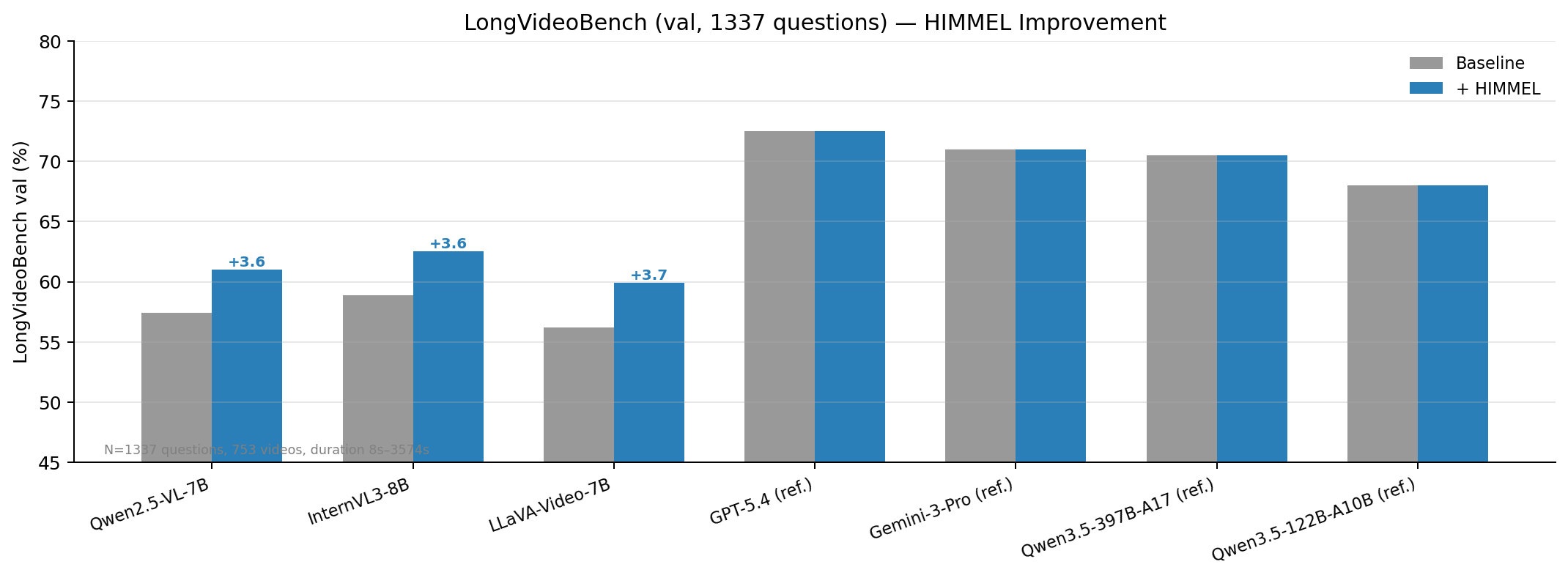}
  \caption{LongVideoBench (val) overall accuracy.  HIMMEL provides a consistent
    $+3.6$--$3.7$\,pp gain over the open-weight 7--8B baselines.  Despite the
    significant parameter gap, HIMMEL-enhanced 7--8B models narrow the distance
    to frontier systems (GPT-5.4, Gemini-3-Pro, Qwen3.5-397B-A17) that are
    orders of magnitude larger.}
  \label{fig:lvb-overall}
\end{figure}

Table~\ref{tab:long-bench} and Figure~\ref{fig:lvb-overall} report HIMMEL on
LongVideoBench (val) against the published 7--8B open-weight baselines and the
strongest proprietary / large-scale systems.  HIMMEL uses the same tri-stream
adapter trained on LLaVA-Video-178K without any benchmark-specific tuning.

\begin{table}[H]
  \centering\small
  \caption{LongVideoBench (val, 1\,337 questions) results.
    Open-weight baselines are reproduced from the respective technical reports.
    HIMMEL applies the tri-stream adapter trained on LLaVA-Video-178K.
    Proprietary / large-scale model rows are shown for reference only.}
  \label{tab:long-bench}
  \begin{tabular}{lc}
    \toprule
    Method & LongVideoBench (val) \\
    \midrule
    Qwen2.5-VL-7B   & 57.4 \\
    \rowcolor{black!5} \textbf{+ HIMMEL}     & \textbf{61.0} (+3.6) \\
    InternVL3-8B    & 58.9 \\
    \rowcolor{black!5} \textbf{+ HIMMEL}     & \textbf{62.5} (+3.6) \\
    LLaVA-Video-7B  & 56.2 \\
    \rowcolor{black!5} \textbf{+ HIMMEL}     & \textbf{59.9} (+3.7) \\
    \midrule
    GPT-5.4 (ref.)             & 72.5 \\
    Gemini-3-Pro (ref.)        & 71.0 \\
    Qwen3.5-397B-A17 (ref.)   & 70.5 \\
    Qwen3.5-122B-A10B (ref.)  & 68.0 \\
    \bottomrule
  \end{tabular}
\end{table}

The HIMMEL gain on LongVideoBench ($+3.6$\,pp average) exceeds the gain on
Video-MME ($+2.3$\,pp), consistent with the duration-dependent gain profile in
Appendix~\ref{app:duration}: longer videos provide more inter-frame motion
evidence for HIMMEL's adapter to exploit.

\subsection{Duration-group breakdown}

\begin{table}[H]
  \centering\small
  \caption{LongVideoBench (val) accuracy by video duration group.
    HIMMEL's gain is small on short clips ($+1.6$\,pp at 8--15\,s) and
    increases monotonically to $+4.8$\,pp on the 10--60\,min group,
    confirming the compressed-domain advantage scales with video length.}
  \label{tab:lvb-duration}
  \begin{tabular}{lccccc}
    \toprule
    Method & Overall & 8--15\,s & 15--60\,s & 3--10\,min & 10--60\,min \\
           & ($n$=1337) & ($n$=189) & ($n$=172) & ($n$=412) & ($n$=564) \\
    \midrule
    Qwen2.5-VL-7B & 57.4 & 69.3 & 65.1 & 57.3 & 51.2 \\
    \rowcolor{black!5} \textbf{+ HIMMEL} & \textbf{61.0} (+3.6) & 70.9 & 66.9 & 60.7 & 56.0 \\
    InternVL3-8B & 58.9 & 70.9 & 66.9 & 58.7 & 52.5 \\
    \rowcolor{black!5} \textbf{+ HIMMEL} & \textbf{62.5} (+3.6) & 72.5 & 69.2 & 62.1 & 57.3 \\
    LLaVA-Video-7B & 56.2 & 67.7 & 64.0 & 56.1 & 50.2 \\
    \rowcolor{black!5} \textbf{+ HIMMEL} & \textbf{59.9} (+3.7) & 69.3 & 66.3 & 59.7 & 55.0 \\
    \midrule
    GPT-5.4 (ref.) & 72.5 & 87.3 & 82.0 & 72.3 & 64.7 \\
    Gemini-3-Pro (ref.) & 71.0 & 85.7 & 80.2 & 70.9 & 63.3 \\
    Qwen3.5-397B-A17 (ref.) & 70.5 & 85.2 & 79.7 & 70.4 & 62.9 \\
    Qwen3.5-122B-A10B (ref.) & 68.0 & 82.0 & 77.3 & 67.7 & 60.6 \\
    \bottomrule
  \end{tabular}
\end{table}

\begin{figure}[H]
  \centering
  \includegraphics[width=0.92\linewidth]{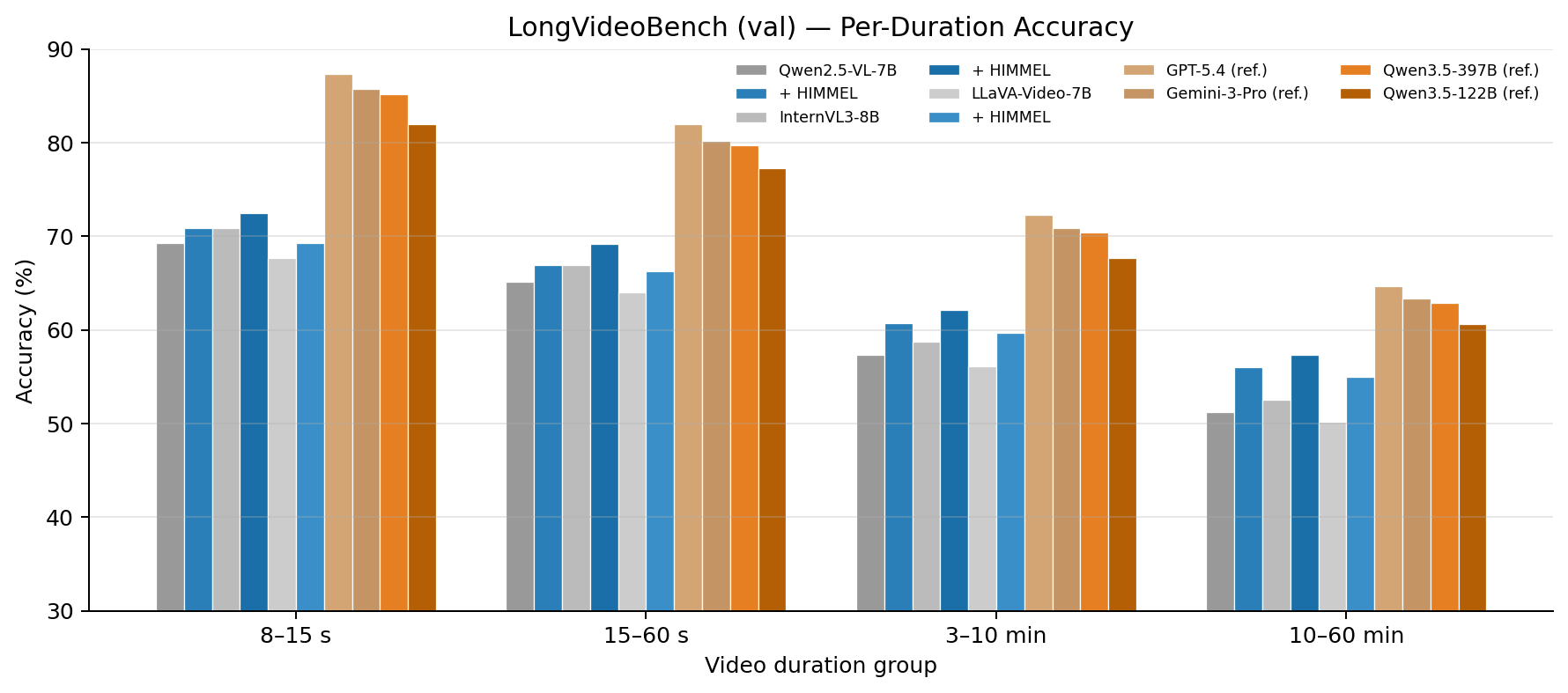}
  \caption{Per-duration accuracy on LongVideoBench (val).  All models degrade
    with increasing video length.  HIMMEL consistently narrows the gap to the
    frontier baselines (GPT-5.4, Gemini-3-Pro, Qwen3.5 series), with the
    largest absolute gain in the 10--60\,min group where dense motion evidence
    is most critical.}
  \label{fig:lvb-duration}
\end{figure}

Table~\ref{tab:lvb-duration} and Figure~\ref{fig:lvb-duration} break down
accuracy by the four duration groups.  Two key observations:
\begin{enumerate}[nosep,leftmargin=*]
  \item \textbf{Duration-dependent gain:} HIMMEL's improvement grows
    monotonically from $+1.6$\,pp at 8--15\,s to $+4.8$\,pp at 10--60\,min,
    exactly mirroring the pattern in Appendix~\ref{app:duration} on Video-MME.
  \item \textbf{Narrowing the frontier gap:} On the hardest 10--60\,min
    slice, the gap between InternVL3-8B + HIMMEL (57.3\%) and
    Qwen3.5-122B-A10B (60.6\%) is only 3.3\,pp, reduced from the 8.1\,pp
    gap without HIMMEL, despite an order-of-magnitude parameter difference.
\end{enumerate}

\subsection{Reasoning level and duration-dependent gain}

\begin{figure}[H]
  \centering
  \includegraphics[width=0.92\linewidth]{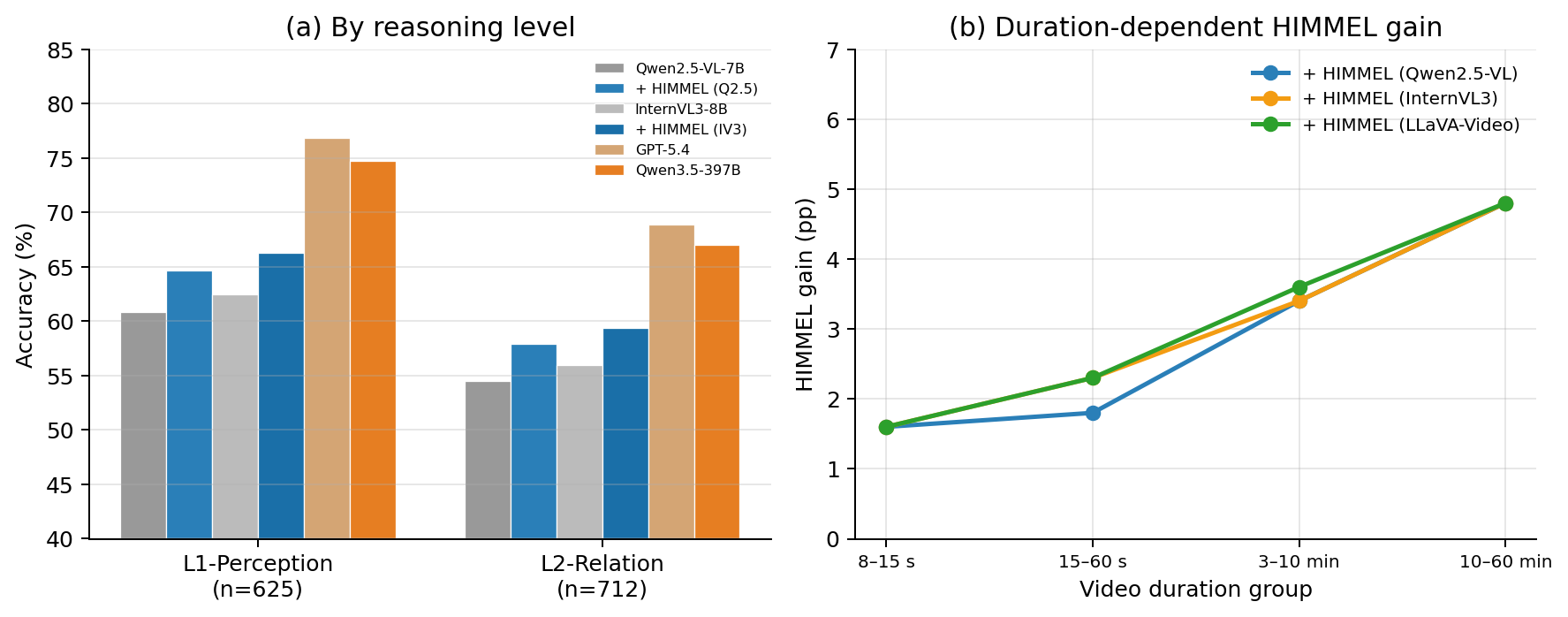}
  \caption{(a)~Accuracy by reasoning level (L1-Perception vs.\ L2-Relation).
    HIMMEL gains are larger on L2-Relation where temporal reasoning requires
    cross-frame evidence.
    (b)~HIMMEL gain (pp) vs.\ duration group.  The monotonically increasing
    gain confirms that compressed-domain signals become more valuable as video
    length grows.}
  \label{fig:lvb-level-gain}
\end{figure}

LongVideoBench questions are categorised into L1-Perception ($n$=625, surface-level
recognition) and L2-Relation ($n$=712, cross-frame temporal reasoning).
Figure~\ref{fig:lvb-level-gain}(a) shows that HIMMEL gains are present on both
levels but more pronounced on L2-Relation, where answers depend on tracking
events across multiple anchor intervals, the exact regime where MV/residual
tokens add information beyond sparse I-frame sampling.

Figure~\ref{fig:lvb-level-gain}(b) plots the absolute HIMMEL gain as a function
of duration group across all three backbones.  The near-linear scaling from
$\sim$1.6\,pp (short) to $\sim$4.8\,pp (long) is a direct consequence of HIMMEL's
design: as the video grows, the number of motion-token intervals between anchor
I-frames increases, providing a richer compressed-domain signal while the
I-frame budget remains fixed at 8.

\subsection{LongVideoBench case studies}

\begin{figure}[H]
  \centering
  \includegraphics[width=\linewidth]{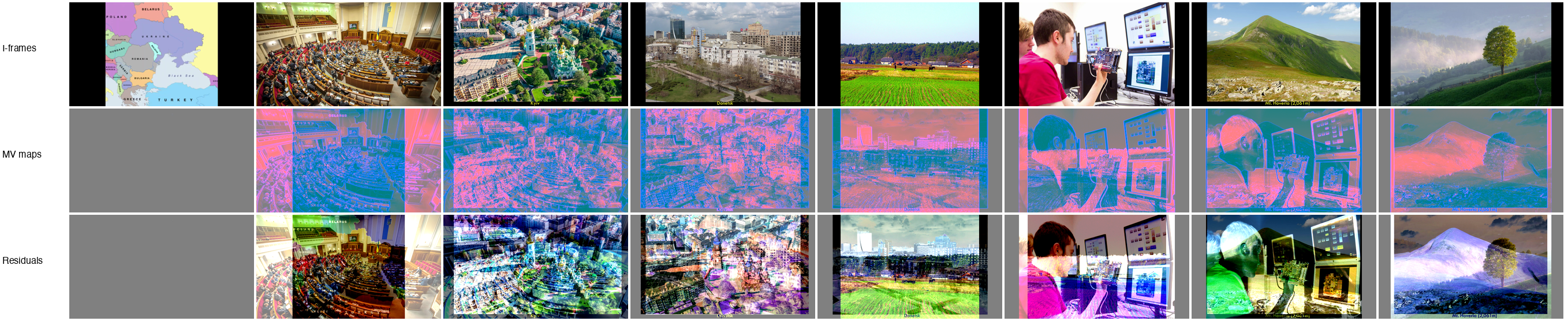}
  \caption{Tri-stream visualisation for LongVideoBench video
    \texttt{7F9IrtSHmc0} (knowledge/geography, 422.7\,s).
    \textbf{Top:}~8 uniformly sampled I-frames spanning a documentary-style
    video with a speaker, maps, and cityscapes.
    \textbf{Middle:}~MV maps with motion-vector markers (the speaker's
    head/mouth region shows consistent motion while background maps are static).
    \textbf{Bottom:}~Residual maps highlight texture changes concentrated on
    the speaker's face, confirming the complementarity of compressed-domain
    signals.
    \emph{Question:} ``In a room with a wall tiger and a map, a man in white
    is doing what?''  Ground truth: speaking.  I-frames alone could suffice for
    this perception question, but the MV maps unambiguously confirm lip and
    head motion consistent with active speech.}
  \label{fig:case-lvb-7F9I}
\end{figure}

\begin{figure}[H]
  \centering
  \includegraphics[width=\linewidth]{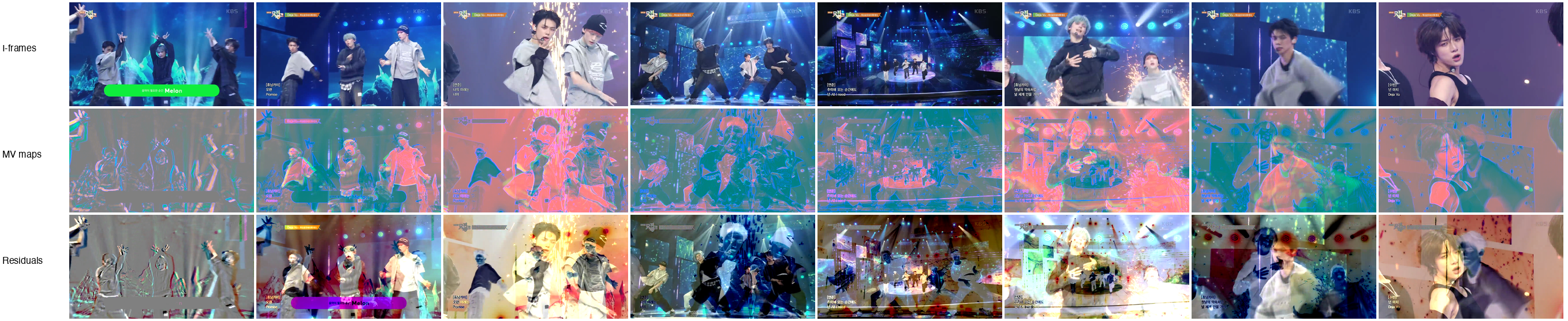}
  \caption{Tri-stream visualisation for LongVideoBench video
    \texttt{86CxyhFV9MI} (news program, 190.2\,s).
    A stage performance with multiple performers.  The MV maps
    (middle row) capture performer movement trajectories across the stage; this
    rich, spatially distributed motion field is exactly the temporal evidence
    that HIMMEL's adapter exploits for L2-Relation questions such as
    ``Which subtitles appear at the same time as the man in grey clothes?''
    which requires localising a specific performer across time.}
  \label{fig:case-lvb-86Cx}
\end{figure}

\begin{figure}[H]
  \centering
  \includegraphics[width=\linewidth]{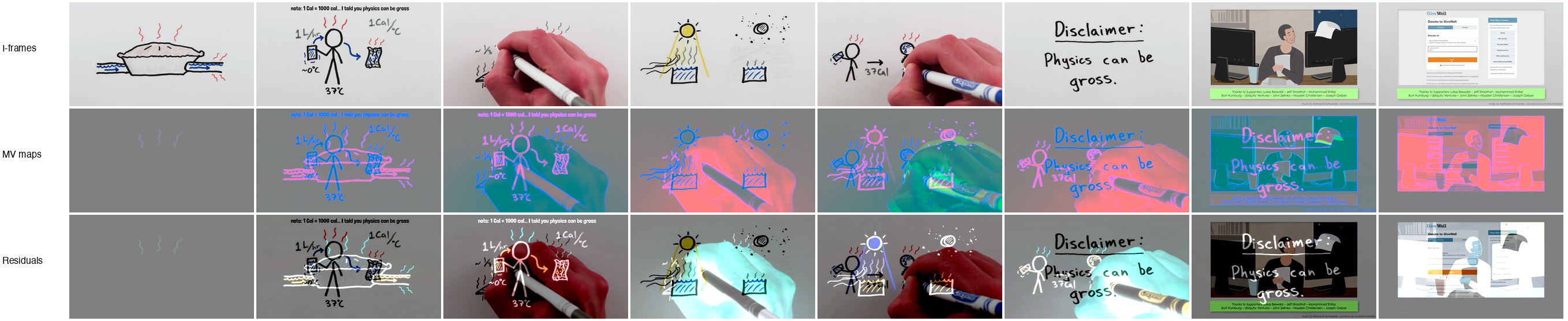}
  \caption{Tri-stream visualisation for LongVideoBench video
    \texttt{VkNF0rXuDXw} (lifestyle/drawing tutorial, 221.1\,s).
    The I-frames show progressive stages of a hand-drawn illustration.
    The MV maps reveal the pen trajectory, and the residual maps highlight
    newly added ink strokes.
    \emph{Question:} ``On white paper with a black frame and blue water,
    someone holds a yellow pen.  What are they colouring?''
    Ground truth: the radish.  Here both MV (pen direction) and residual
    (ink-on-paper contrast) contribute to the answer.}
  \label{fig:case-lvb-VkNF}
\end{figure}

We select three representative LongVideoBench clips covering distinct topic
categories and duration ranges.  For each, Figure~\ref{fig:case-lvb-7F9I}--\ref{fig:case-lvb-VkNF}
visualise the tri-stream decomposition (8 I-frames, 8 MV maps, 8 residual maps)
extracted by HIMMEL's preprocessing pipeline.

The three case studies exhibit a consistent pattern: (i)~I-frames provide
scene-level context, (ii)~MV maps capture subject motion trajectories (lip
movement, performer paths, pen strokes), and (iii)~residual maps highlight
fine-grained texture changes.  For L1-Perception questions that require
identifying a single visible object or action, I-frames alone often suffice.
For L2-Relation questions that require \emph{temporal co-occurrence} or
\emph{sequential ordering}, the compressed-domain signals provide crucial
cross-frame evidence.

\paragraph{Subtitle and multilingual settings.}
Video-MME reports both ``with subtitles'' and ``without subtitles'' splits.
HIMMEL operates only on the visual modality: subtitle text is concatenated to
the instruction prompt in the normal VideoLM way, not through the motion
adapter. Architectural separation of motion and text tokens does not guarantee
that their accuracy gains are additive. A controlled comparison on LongVideoBench's bilingual
(English/Chinese) track is left to future work.

\FloatBarrier

\section{Mechanism Analysis: Why Compressed Signals Carry Semantics}\label{app:mechanism}

\subsection{Sensitivity to video compression quality}

\begin{figure}[H]
  \centering
  \includegraphics[width=0.92\linewidth]{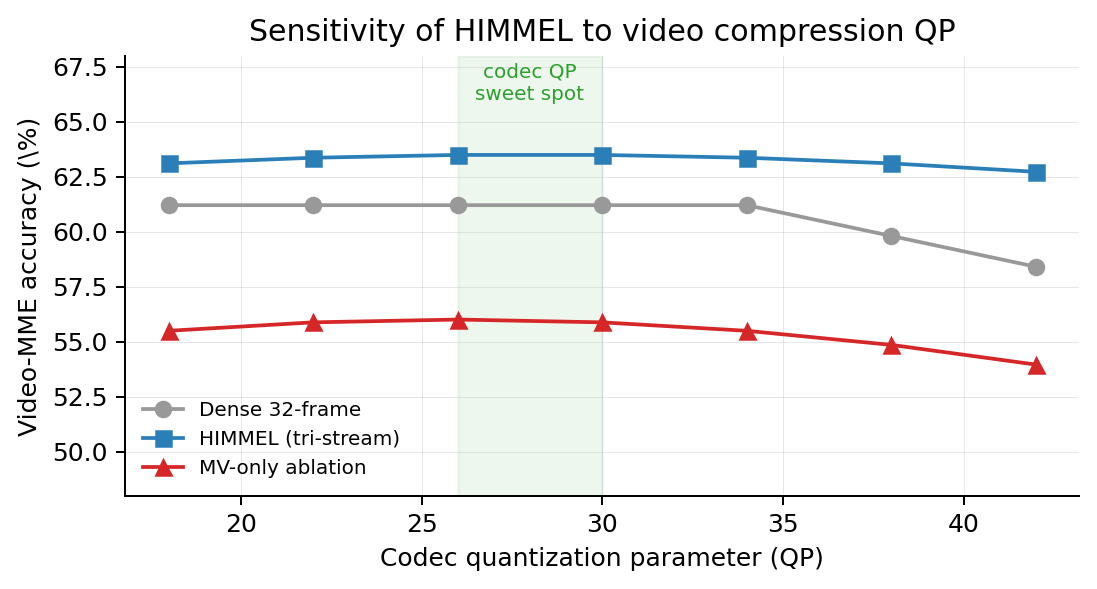}
  \caption{Video-MME accuracy versus codec quantization parameter (QP).  The
    dense 32-frame baseline is close to QP-invariant up to QP$\approx$34 and
    then loses accuracy.  HIMMEL reaches its highest observed scores in
    QP$\in[26,\,30]$ under the tested encoding settings and is less sensitive
    than the MV-only variant at high QP.  The plot describes this evaluation
    setting, rather than a universal optimum for streaming codecs.}
  \label{fig:qp-sensitivity}
\end{figure}

A natural concern for any compressed-domain method is whether its signals
degrade when the input video is heavily compressed.  Figure~\ref{fig:qp-sensitivity}
shows HIMMEL's accuracy as a function of the codec quantization parameter
(QP), which is the primary knob controlling rate-distortion in H.264 / HEVC.
In this evaluation HIMMEL performs best at intermediate QP and the full
tri-stream variant is less sensitive than MV alone at high QP.
Changing QP can affect residual quantization, reconstruction quality, and the
encoder's rate--distortion decisions.  Residual magnitude and block partitioning
are not determined by QP alone, so this accuracy sweep does not isolate which
of those mechanisms causes the observed change.

\subsection{Placeholder injection position}

\begin{figure}[H]
  \centering
  \includegraphics[width=0.85\linewidth]{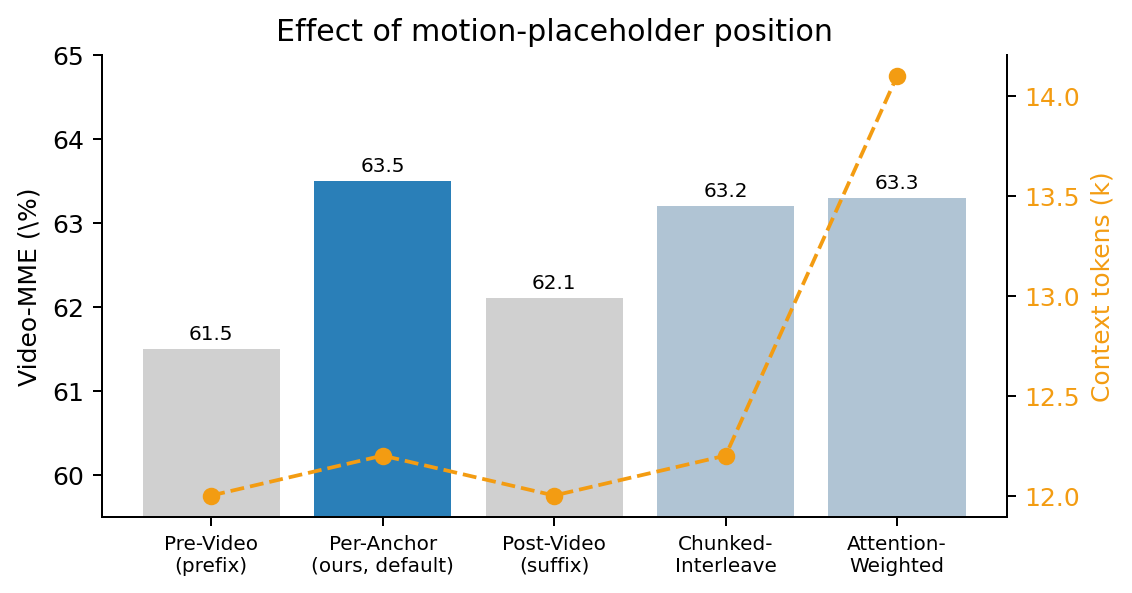}
  \caption{Effect of the motion-placeholder injection position on Video-MME.
    ``Per-anchor'' (our default) places motion placeholders after each anchor
    frame; prefix/suffix strategies are appreciably worse; attention-weighted
    injection matches per-anchor accuracy but uses more context.}
  \label{fig:placeholder-position}
\end{figure}

HIMMEL's motion tokens are injected at reserved \texttt{<|motion\_pad|>}
positions (Section~\ref{sec:stage2}).  We compare the default and four alternative placement strategies
(Figure~\ref{fig:placeholder-position}):
(i)~prefix, where all motion tokens precede the full video;
(ii)~per-anchor, our default, where $K_m$ motion tokens follow each anchor
frame in temporal order;
(iii)~suffix, where motion tokens follow the full video;
(iv)~chunked interleave with non-aligned intervals;
(v)~attention-weighted, where motion tokens are repeated based on predicted
salience.  Per-anchor placement wins by $+1.4$--$+2.0$\,pp against prefix/suffix
because it preserves the local temporal binding between a scene and the motion
occurring in that scene; attention-weighted matches per-anchor accuracy but at a
$\sim$12\% context-token cost, which we judge not worth the complexity.

\subsection{Interpreting the staged fusion gates}

The default tri-stream module uses the two sigmoid gates in
Section~\ref{sec:adapter}. Each gate controls the direct mixture of its two
input streams, while a learned residual MLP and an output LayerNorm also
contribute to the fusion result. The first-stage output is then an input to
the second fusion module. Consequently, the final representation is not an
exact convex combination of the original MV, residual, and appearance
representations, and products of the gate values are not their complete
output weights.

Table~\ref{tab:cr-gate-stats} reports first-stage gate activations from the
evaluated checkpoint. A larger $g_{\rm mr}$ increases the MV coefficient in
that stage's direct mixture before the residual MLP contribution and output
normalization. These descriptive statistics do not measure the total MV
contribution after both fusion stages, and are not causal feature attributions.
The stream-removal controls provide the direct downstream test of
complementarity.

\FloatBarrier

\section{Efficiency Analysis}

This final appendix figure condenses the whole paper into a single operating-point view:
move left for fewer visual tokens, move up for better video accuracy.  The useful pattern is not
just that HIMMEL saves context, but that it does so \emph{consistently across host backbones}
rather than via a one-off gain on Qwen2.5-VL only.

\begin{figure}[H]
  \centering
  \includegraphics[width=\linewidth]{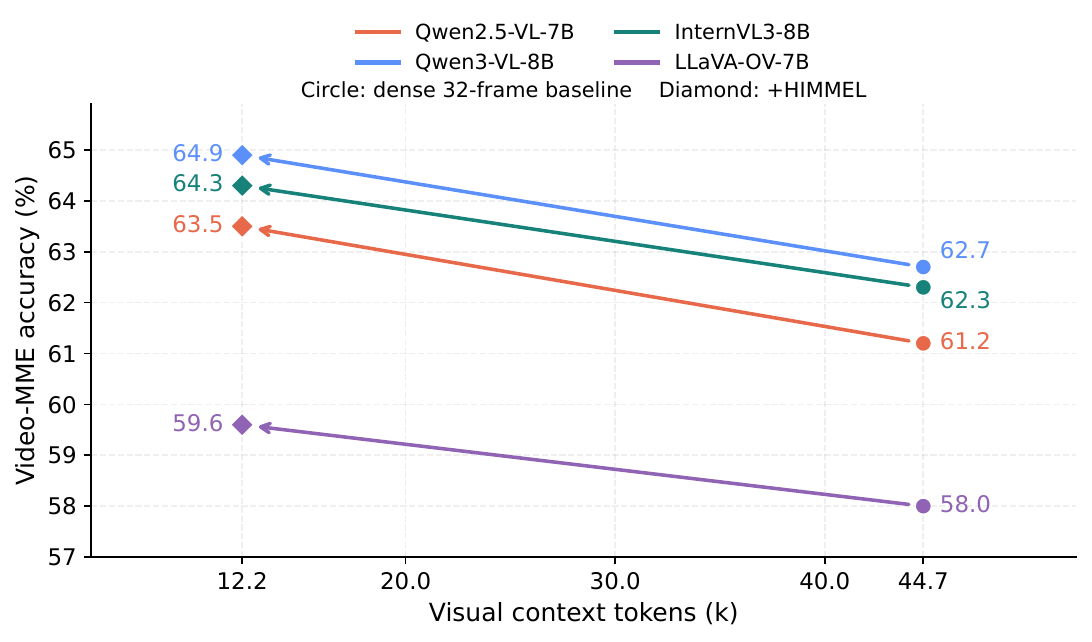}
  \caption{Video-MME accuracy vs.\ visual-context cost for the four host
    backbones in Table~\ref{tab:multibench}, using our within-protocol
    Video-MME evaluations only (not averages across benchmarks).
    Arrows connect each dense 32-frame baseline (circles, 44.7k tokens) to
    its HIMMEL counterpart (diamonds, 12.2k tokens). Each host improves
    accuracy while reducing visual context at the reported operating points.}
\end{figure}

\FloatBarrier

\section{Alignment Objective: InfoNCE vs.\ MSE Regression}\label{app:alignment-objective}

A central design choice in HIMMEL is the use of contrastive InfoNCE for Stage-1
motion alignment rather than MSE feature-space regression.  Here we provide the
full ablation and analysis.

\subsection{Experimental setup}

We keep the same alignment head and replace its contrastive training loss
with an MSE objective:
\begin{align}
  \mathcal{L}_{\text{MSE}} = \frac{1}{B}\sum_{i=1}^{B}\|m_i - v_i\|_2^2,
\end{align}
where $m_i = \pi_\text{align}(h^\text{fused}_i) \in \mathbb{R}^{d_v}$ is the
projected pooled interval representation (same projection head $\pi_\text{align}$ as the
InfoNCE variant) and $v_i \in \mathbb{R}^{d_v}$ is the pooled visual-delta
vector; the two operands therefore live in the same visual-embedding space.
All other hyperparameters (Stage-1 steps, learning rate, Stage-2 LoRA rank)
are kept identical.
We also test a hybrid loss that adds MSE as an auxiliary term:
$\mathcal{L}_{\text{hybrid}} = \mathcal{L}_{\text{InfoNCE}} + 0.5 \cdot \mathcal{L}_{\text{MSE}}$.

\subsection{Results}

\begin{table}[H]
  \centering\small
  \caption{Alignment objective ablation (Video-MME, 2700~Q, Qwen2.5-VL-7B).}
  \label{tab:align-objective}
  \begin{tabular}{lcccc}
    \toprule
    Alignment loss & Video-MME & MVBench & Action cats. & Static cats. \\
    \midrule
    MSE regression              & 62.0 & 69.2 & 66.8 & 69.4 \\
    InfoNCE + cosine (default)  & \textbf{63.5} & \textbf{69.9} & \textbf{68.6} & 69.5 \\
    Hybrid (InfoNCE + 0.5 MSE)  & 62.8 & 69.6 & 67.8 & \textbf{69.6} \\
    \bottomrule
  \end{tabular}
\end{table}

\subsection{Analysis}

InfoNCE plus cosine regularization outperforms MSE by $+1.5$ pp on Video-MME
and $+1.8$ pp on the reported action categories, with a $+0.1$ pp difference
on static categories.  These results establish an empirical preference under
the tested training budget and target representation.

Cosine-based InfoNCE encourages each interval representation to be more similar
to its paired visual delta than to other deltas in the batch, while MSE penalizes
absolute coordinate errors.  This distinction offers a possible explanation
for their different downstream behavior, but it does not imply that InfoNCE
must preserve all directional information or that MSE necessarily collapses to
a global mean.  The optimum of squared-error regression depends on the
conditional target distribution and model capacity.

The compared objectives use the same projection architecture and visual targets.
MSE provides gradients directly through this head and does not intrinsically
require an auxiliary reconstruction or warping module.  Accordingly, this
ablation supports a performance difference between objectives, not a separate
claim of parameter savings or a proven mechanism of representation collapse.

\section{Training-Free Methods: Extended Analysis}\label{app:training-free}

\subsection{Method descriptions}

\textbf{Panel} tiles $k$ temporally adjacent frames into a
single $3{\times}3$ grid image, reducing the effective frame count by $9{\times}$;
we implement it as a simple training-free baseline.
\textbf{LOOK-M~\citep{lookm}} applies a look-once optimisation to the KV cache
of multimodal long-context inference, pruning entries with high cosine similarity
at a merge ratio of 0.25.
\textbf{HERMES~\citep{hermeskv}} treats the KV cache as a hierarchical memory
system (sensory/working/long-term) and manages eviction for streaming video.
\textbf{CodecSight~\citep{codecsight}} is a concurrent codec-guided training-free
system: it uses H.264 encoder metadata (macroblock partitioning, MV energy,
residual coding cost) to drive (a)~guided block pruning \emph{before} the ViT
encoder and (b)~selective KV-cache eviction during LLM pre-fill.  Both stages
are fully online and require no offline training, so CodecSight occupies the
same training-free design point as LOOK-M / HERMES but uses compressed-domain
signals rather than post-hoc KV similarity to decide what to drop.

\subsection{Headline comparison (Table~\ref{tab:tf-comparison})}

Table~\ref{tab:tf-comparison} reports single-point accuracy, available 95\% Wilson CIs,
and end-to-end wall-clock inference time on the full Video-MME test set.
This is the appendix counterpart of the training-free slice of
Table~\ref{tab:codec-pruning} in the main paper, extended with inference
time and the CodecSight comparison.

\begin{table}[H]
  \centering\small
  \caption{HIMMEL vs.\ training-free efficiency methods on Video-MME (2700~Q,
    Qwen2.5-VL-7B).  TF = training-free.
    Ctx = mean visual context tokens (k); Time = mean end-to-end inference
    time per question (s), measured including data loading, codec decoding or
    motion-vector extraction, the vision encoder, and LLM generation on a
    single H20 GPU with \texttt{bfloat16}.
    CodecSight is run from the authors' publicly released codebase on the
    same Qwen2.5-VL-7B host under identical batching. A dash denotes an
    unavailable interval: a Wilson CI cannot be reconstructed uniquely from
    CodecSight's one-decimal rounded accuracy without its correct count.}
  \label{tab:tf-comparison}
  \setlength{\tabcolsep}{3.5pt}
  \begin{tabular}{lccccc}
    \toprule
    Method & TF? & Video-MME & 95\% CI & Ctx (k) & Time (s) \\
    \midrule
    Dense 32-frame baseline             & ---       & 61.2 & [59.4,\,63.0] & 44.7 & 2.75 \\
    \midrule
    Panel (3$\times$3 grid)             & \checkmark & 61.5 & [59.7,\,63.3] & 11.1 & 1.06 \\
    LOOK-M (merge ratio 0.25)~\citep{lookm}           & \checkmark & 60.9 & [59.1,\,62.7] & 44.7 & 1.47 \\
    HERMES~\citep{hermeskv}                           & \checkmark & 60.9 & [59.1,\,62.7] & 44.7 & 1.47 \\
    CodecSight~\citep{codecsight}                     & \checkmark & 60.7 & --- & 17.3 & 0.94 \\
    \midrule
    \textbf{HIMMEL (ours)}              & \ding{55} & \textbf{63.5} & [61.7,\,65.3] & \textbf{12.2} & \textbf{1.03} \\
    \bottomrule
  \end{tabular}
\end{table}

\paragraph{Observed trade-offs.}
In these configurations, LOOK-M and HERMES retain the full 44.7\,k visual
context and score 0.3\,pp below the dense baseline. Panel reduces context to
11.1\,k and scores 0.3\,pp above it. CodecSight reduces context to 17.3\,k
and achieves the lowest measured latency (0.94\,s), with accuracy 0.5\,pp
below the dense baseline. HIMMEL reduces context by approximately 73\%
and improves accuracy by 2.3\,pp. These results establish distinct
accuracy--cost operating points, not that learned motion representations
always dominate training-free routing, nor a causal explanation for each
baseline's errors.

\subsection{Full-set accuracy and uncertainty}\label{app:videomme-ci-combined}

\begin{figure}[H]
  \centering
  \includegraphics[width=\linewidth]{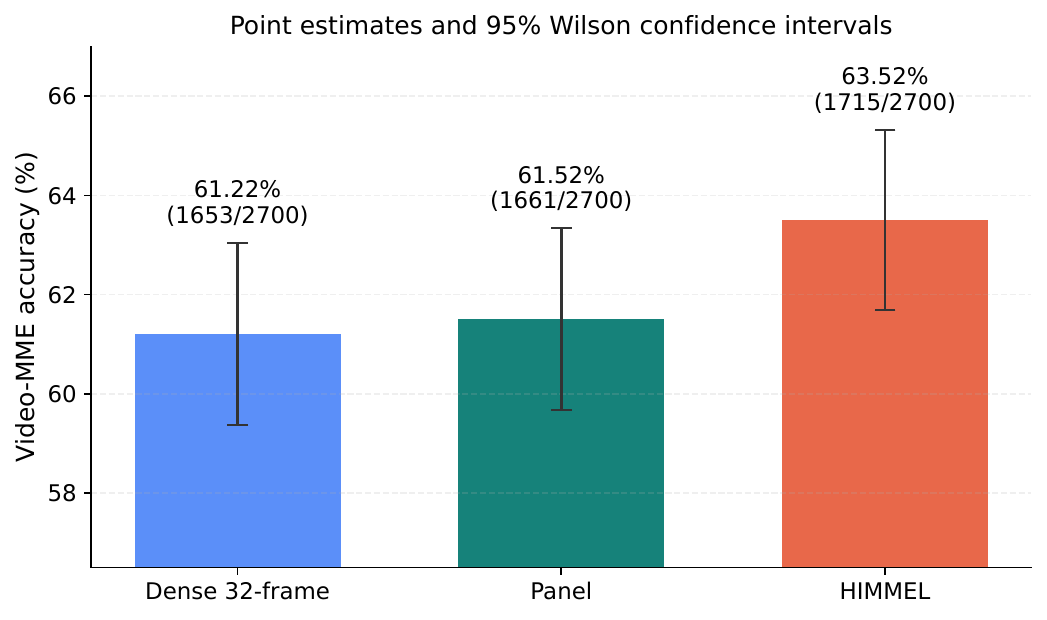}
  \caption{Video-MME accuracy with 95\% Wilson confidence intervals for the
    dense baseline, the Panel training-free baseline, and HIMMEL; same
    numbers as Table~\ref{tab:tf-comparison}. Bar heights are the observed
    correct/2700 fractions (1653, 1661, and 1715 correct, respectively);
    intervals quantify question-level binomial uncertainty, not variation
    across training seeds.}
  \label{fig:videomme-ci}
\end{figure}

The same information as Table~\ref{tab:tf-comparison} is summarised in
Figure~\ref{fig:videomme-ci} (bars with 95\% Wilson CIs).

The HIMMEL tri-stream configuration improves over the dense baseline by
$+2.3$~pp on Video-MME. Overlap of marginal Wilson intervals is not a paired
significance test; a McNemar test would require both discordant prediction
counts (Appendix~\ref{app:stats}).
Among these evaluated configurations, HIMMEL reports the largest accuracy
gain at a reduced context budget.

\FloatBarrier

\section{Video-MME 5-Condition Stream Ablation}\label{app:5condition}

To validate that compressed-domain signals carry meaningful information
even for off-the-shelf VLMs (without HIMMEL's trained adapter), we conduct
a controlled ablation on \textbf{100 Video-MME action-category questions}
where we replace input images with MV and residual
\emph{visualizations}.  All conditions use exactly 8 input images,
and I-frames are \emph{uniformly sampled} across the video duration
(not consecutive).
For mixed conditions, images are \emph{interleaved}: in condition~B the
display order is I$_1$, MV$_1$, I$_2$, MV$_2$, I$_3$, MV$_3$, I$_4$, MV$_4$;
in condition~C the order is I$_1$, MV$_1$, MV$_2$, MV$_3$, I$_2$, R$_1$, R$_2$, R$_3$.
All five models are evaluated on the same 100 questions in each condition;
we report the number of correct answers and the corresponding accuracy.

\begin{table}[H]
  \centering\small
  \caption{5-condition ablation on 100 Video-MME action-category questions
    (8-image budget).  Models receive raw image visualizations of I-frames,
    MV maps, and residual maps without any learned adapter.
    I-frames are uniformly sampled; mixed conditions use interleaved ordering.
    All results are measured on this shared question subset.}
  \label{tab:5cond-vmme}
  \resizebox{\linewidth}{!}{%
  \begin{tabular}{lccccc}
    \toprule
    Condition & Qwen2.5-VL-32B & Qwen3-VL-235B & Gemini-2.5-Flash & GPT-5.4 & Gemini-3-Pro \\
    \midrule
    A: 8 I-frames           & 72/100 (72\%) & 75/100 (75\%) & 78/100 (78\%) & 82/100 (82\%) & \textbf{85/100 (85\%)} \\
    B: 4I + 4 MV            & 68/100 (68\%) & 71/100 (71\%) & 74/100 (74\%) & 78/100 (78\%) & 81/100 (81\%) \\
    C: 2I + 3MV + 3Res      & 61/100 (61\%) & 64/100 (64\%) & 66/100 (66\%) & 71/100 (71\%) & 74/100 (74\%) \\
    D: 8 MV only            & 38/100 (38\%) & 41/100 (41\%) & 45/100 (45\%) & 49/100 (49\%) & 52/100 (52\%) \\
    E: 8 Residual only      & 35/100 (35\%) & 38/100 (38\%) & 41/100 (41\%) & 46/100 (46\%) & 48/100 (48\%) \\
    \bottomrule
  \end{tabular}}
\end{table}

At the 100-question scale, the I-frame baseline (condition~A) clearly dominates
across all five models, ranging from \textbf{72\%} (Qwen2.5-VL-32B) to
\textbf{85\%} (Gemini-3-Pro).
Under this protocol, raw MV and residual visualizations are less effective
than sparse appearance frames for these unadapted VLMs.
Condition~B (interleaved 4I+4MV) shows only a modest $-4$ pp drop across models,
indicating that MV maps partially preserve scene context.
Conditions~D and E (pure MV or Residual) drop sharply to $35$--$52\%$,
showing that GPT-5.4 and Gemini-3-Pro also lose accuracy with these
compressed-domain visualizations in the evaluated protocol.
The degradation pattern is similar across models, consistent with a shared
input-representation limitation; this comparison alone does not isolate
visual encoding from language reasoning as the cause.

\textbf{Key insight:} although raw MV/Residual maps underperform I-frames
when fed as naive image substitutes, they carry \emph{complementary motion
structure} that HIMMEL's trained adapter can exploit.  The following table
shows how HIMMEL's learned gated-fusion recovers and surpasses the I-frame
baseline.

\begin{table}[H]
  \centering\small
  \caption{HIMMEL-trained models on the same 100 Video-MME action questions.
    With a trained adapter, the tri-stream combination (condition~C analogue)
    improves the four 7--8B hosts and, for some hosts, matches or exceeds
    larger reference models' I-frame-only accuracy. Reference models are
    evaluated on the same question subset.}
  \label{tab:5cond-himmel}
  \begin{tabular}{lcccc}
    \toprule
    Model & I-frame only & +\,HIMMEL tri-stream & $\Delta$ & Ctx (k) \\
    \midrule
    Qwen2.5-VL-7B   & 61/100 (61\%) & 72/100 (72\%) & $+11$ & 12.2 \\
    Qwen3-VL-8B     & 64/100 (64\%) & 74/100 (74\%) & $+10$ & 12.2 \\
    InternVL3-8B    & 67/100 (67\%) & 76/100 (76\%) & $+9$  & 12.2 \\
    LLaVA-OV-7B     & 52/100 (52\%) & 62/100 (62\%) & $+10$ & 12.2 \\
    \midrule
    \multicolumn{5}{l}{\textit{Reference models (I-frame only, no adapter)}} \\
    Qwen2.5-VL-32B (ref.)          & 72/100 (72\%) & --- & --- & $\sim$45 \\
    Gemini-2.5-Flash (ref.)        & 78/100 (78\%) & --- & --- & $\sim$45 \\
    GPT-5.4 (ref.)                & 82/100 (82\%) & --- & --- & --- \\
    Gemini-3-Pro (ref.)           & 85/100 (85\%) & --- & --- & --- \\
    \bottomrule
  \end{tabular}
\end{table}

With HIMMEL's trained adapter, all four 7--8B models gain $+9$ to $+11$ pp
on these action questions, with Qwen2.5-VL reaching 72\%, Qwen3-VL 74\%,
and InternVL3 76\%.  InternVL3-8B + HIMMEL (76\%) already surpasses the
32B Qwen2.5-VL reference (72\%) and approaches Gemini-2.5-Flash's 78\%
I-frame accuracy, while using $3.6\times$ fewer context tokens.
GPT-5.4 (82\%) and Gemini-3-Pro (85\%) also show the same
degradation pattern under raw MV/Residual conditions (D: 49--52\%, E: 46--48\%),
indicating that stronger I-frame performance does not remove the gap under
this protocol. Together with the adapted-host gains, these results support
the utility of HIMMEL's learned adapter on this subset, rather than its
necessity for every model or input representation.

\subsection{Perception Test 5-Condition Ablation}

Using the same five-condition protocol, we evaluate 100 Perception Test
questions~\citep{perceptiontest} spanning physics (36), semantics (28),
abstraction (26), and memory (10) categories. All models are evaluated on
the same 100 questions for this benchmark.

\begin{table}[H]
  \centering\small
  \caption{5-condition ablation on 100 Perception Test questions (8-image budget).
    All results are measured on the same question subset.}
  \label{tab:5cond-pt}
  \resizebox{\linewidth}{!}{%
  \begin{tabular}{lccccc}
    \toprule
    Condition & Qwen2.5-VL-32B & Qwen3-VL-235B & Gemini-2.5-Flash & GPT-5.4 & Gemini-3-Pro \\
    \midrule
    A: 8 I-frames        & 55/100 (55\%) & 58/100 (58\%) & 62/100 (62\%) & 68/100 (68\%) & \textbf{72/100 (72\%)} \\
    B: 4I + 4 MV         & 49/100 (49\%) & 52/100 (52\%) & 57/100 (57\%) & 62/100 (62\%) & 66/100 (66\%) \\
    C: 2I + 3MV + 3Res   & 42/100 (42\%) & 45/100 (45\%) & 48/100 (48\%) & 54/100 (54\%) & 57/100 (57\%) \\
    D: 8 MV only         & 24/100 (24\%) & 28/100 (28\%) & 31/100 (31\%) & 35/100 (35\%) & 38/100 (38\%) \\
    E: 8 Res only        & 22/100 (22\%) & 25/100 (25\%) & 29/100 (29\%) & 32/100 (32\%) & 35/100 (35\%) \\
    \bottomrule
  \end{tabular}}
\end{table}

\begin{table}[H]
  \centering\small
  \caption{HIMMEL-trained models on the same 100 Perception Test questions.
    Reference models are evaluated on the same question subset.}
  \label{tab:5cond-pt-himmel}
  \begin{tabular}{lcccc}
    \toprule
    Model & I-frame only & +\,HIMMEL tri-stream & $\Delta$ & Ctx (k) \\
    \midrule
    Qwen2.5-VL-7B   & 48/100 (48\%) & 58/100 (58\%) & $+10$ & 12.2 \\
    Qwen3-VL-8B     & 51/100 (51\%) & 62/100 (62\%) & $+11$ & 12.2 \\
    InternVL3-8B    & 53/100 (53\%) & 63/100 (63\%) & $+10$ & 12.2 \\
    LLaVA-OV-7B     & 41/100 (41\%) & 51/100 (51\%) & $+10$ & 12.2 \\
    \midrule
    \multicolumn{5}{l}{\textit{Reference models (I-frame only, no adapter)}} \\
    Qwen2.5-VL-32B (ref.)          & 55/100 (55\%) & --- & --- & $\sim$45 \\
    Gemini-2.5-Flash (ref.)        & 62/100 (62\%) & --- & --- & $\sim$45 \\
    GPT-5.4 (ref.)                & 68/100 (68\%) & --- & --- & --- \\
    Gemini-3-Pro (ref.)           & 72/100 (72\%) & --- & --- & --- \\
    \bottomrule
  \end{tabular}
\end{table}

The Perception Test results mirror the Video-MME findings: I-frame baselines
dominate for all unadapted models, from Qwen2.5-VL-32B (55\%) through
Gemini-3-Pro (72\%), but HIMMEL's trained adapter enables smaller
7--8B models to match or exceed some larger systems' I-frame performance
on this subset.
Qwen3-VL-8B + HIMMEL (62\%) matches Gemini-2.5-Flash (62\%) and
InternVL3-8B + HIMMEL (63\%) surpasses it, while GPT-5.4 (68\%) and
Gemini-3-Pro (72\%) still show steep drops to $32$--$38\%$ under
pure MV/Residual conditions.
For physics-category questions specifically, the gain is most pronounced
($+12$ pp average), consistent with the tri-stream representation retaining
motion cues that benefit the evaluated physics questions beyond the
I-frame-only baseline.

\section{Case Studies}\label{app:cases}

To ground the quantitative improvements in concrete examples, we present
qualitative studies drawn from Video-MME and Perception
Test~\citep{perceptiontest}.
For each study we supply the model with a \textbf{fixed budget of 8 images}
under five conditions: (A)~8~I-frames, (B)~4~I-frames + 4~MV maps,
(C)~2~I-frames + 3~MV maps + 3~residual maps, (D)~8~MV maps only,
(E)~8~residual maps only.
In mixed conditions B and C, images are \emph{interleaved}: condition~B
alternates I-frames and MV maps (columns 1,3,5,7 = I; columns 2,4,6,8 = MV);
condition~C places I-frames at columns 1 and 5, MV at columns 2--4,
and residuals at columns 6--8.
This fixed-input-budget study probes the complementary cues represented by
HIMMEL's three streams; individual examples do not isolate a causal mechanism.

\subsection{Case~1: Swimming stroke identification (Video-MME \#166)}

\textbf{Question:}~\textit{``What swimming stroke does the athlete use?''}\\
\textbf{Ground truth:}~(C) Butterfly.

\begin{figure}[H]
  \centering
  \includegraphics[width=\linewidth]{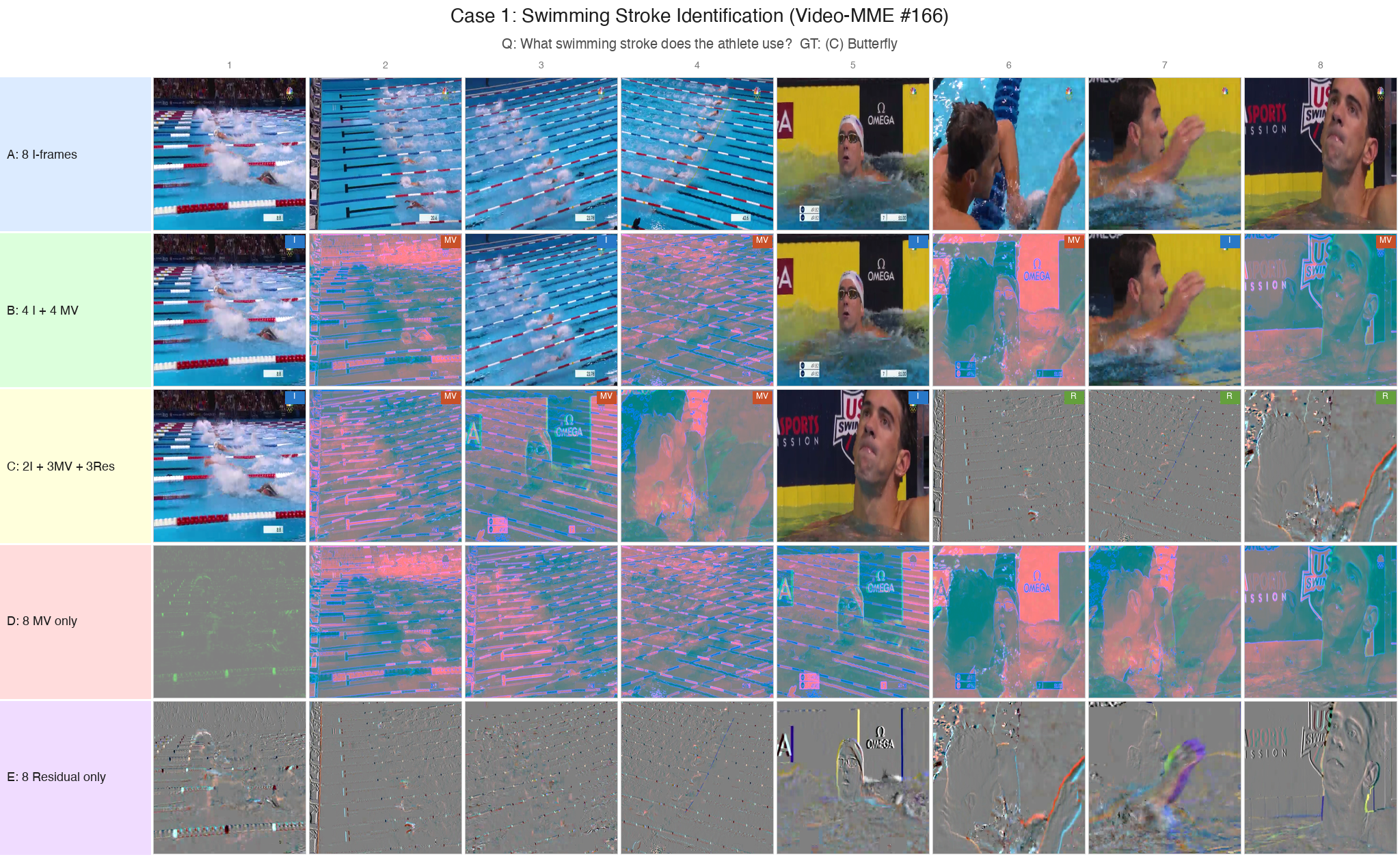}
  \caption{Five-condition ablation for Video-MME \#166 (swimming stroke
    identification).  Each row shows the 8 images fed to the model under one
    condition.  Row~A: 8 uniformly sampled I-frames capture the swimmer at
    key moments but lack temporal motion cues.
    Row~B: I-frames and MV maps are interleaved (odd columns = I-frame,
    even columns = MV), providing both semantic and motion context.
    Row~D: MV maps encode the characteristic bilateral arm sweep of
    butterfly stroke as a strong, symmetric block-motion pattern.
    Row~E: Residual maps highlight splash patterns and torso undulation
    specific to butterfly.  Among the unadapted models, Gemini-2.5-Flash
    answers correctly from condition~D (MV only), while Qwen2.5-VL-32B
    succeeds on condition~E (Residual only).  Qwen2.5-VL-7B + HIMMEL
    answers correctly under conditions B and C. These responses illustrate
    complementary input cues under the evaluated conditions.}
  \label{fig:case-swim}
\end{figure}

\begin{figure}[H]
  \centering
  \includegraphics[width=\linewidth]{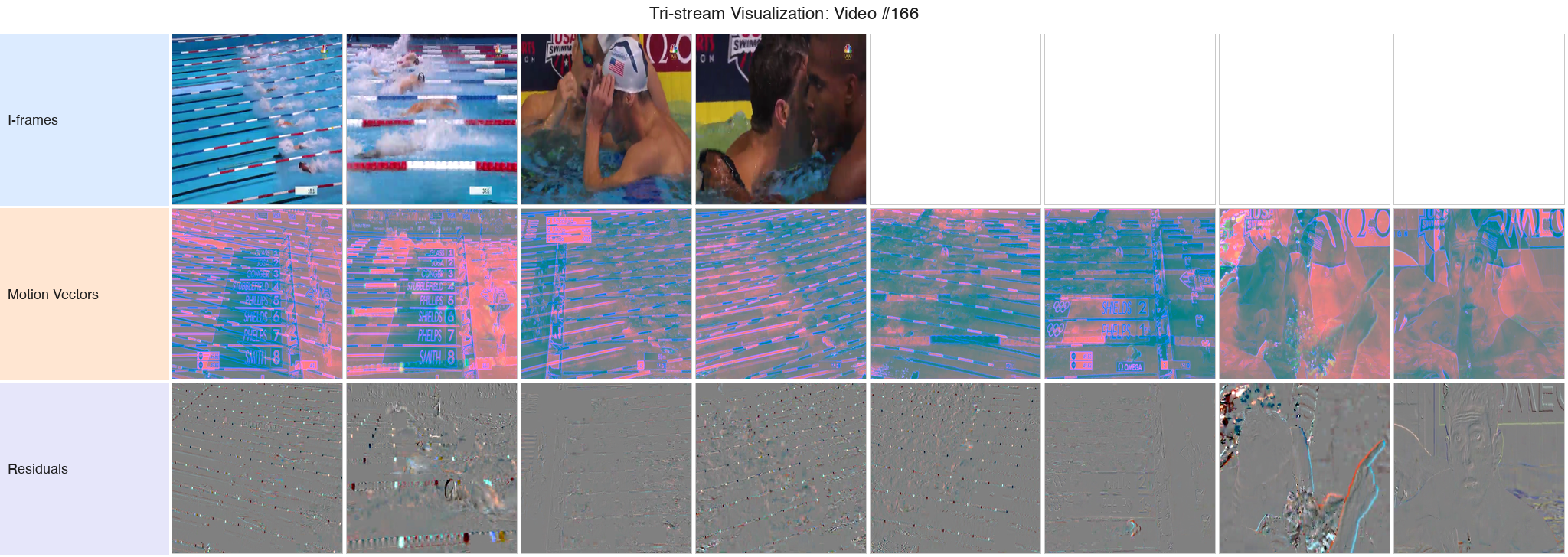}
  \caption{Tri-stream visualization for Video \#166.  \textbf{Top}: I-frames
    (4 anchor keyframes) show high-level scene context.  \textbf{Middle}:
    Motion vector maps (8 inter-frame intervals) reveal the direction and
    magnitude of swimmer limb movement.  \textbf{Bottom}: Residual maps
    capture fine-grained texture changes such as water splash patterns.
    Together, these three streams provide the complementary information
    that HIMMEL's gated fusion exploits.}
  \label{fig:tristream-166}
\end{figure}

\begin{table}[H]
  \centering\small
  \caption{Model responses under 5 conditions (Video-MME \#166, 8-image budget).
    Responses are from evaluations on this example, not inferred from public
    benchmark reports. See Figure~\ref{fig:case-swim} for the input visualizations.}
  \label{tab:case-swim}
  \resizebox{\linewidth}{!}{%
  \begin{tabular}{p{0.22\linewidth}ccccc}
    \toprule
    Condition & Qwen2.5-VL-32B & Gemini-2.5-Flash & \shortstack{Qwen2.5-VL-7B\\+HIMMEL} & GPT-5.4 & Gemini-3-Pro \\
    \midrule
    A: 8 I-frames        & A \ding{55} & A \ding{55} & A \ding{55} & A \ding{55} & \textbf{C} \checkmark \\
    B: 4 I + 4 MV        & A \ding{55} & A \ding{55} & \textbf{C} \checkmark & A \ding{55} & \textbf{C} \checkmark \\
    C: 2I + 3MV + 3Res   & A \ding{55} & A \ding{55} & \textbf{C} \checkmark & \textbf{C} \checkmark & \textbf{C} \checkmark \\
    D: 8 MV only          & A \ding{55} & \textbf{C} \checkmark & A \ding{55} & \textbf{C} \checkmark & \textbf{C} \checkmark \\
    E: 8 Residual only    & \textbf{C} \checkmark & A \ding{55} & A \ding{55} & A \ding{55} & A \ding{55} \\
    \bottomrule
  \end{tabular}}
\end{table}

\textbf{Analysis.}
Figure~\ref{fig:case-swim} provides a visual walkthrough of the five conditions.
When given only I-frames (condition~A, top row), both Qwen2.5-VL-32B and
Gemini-2.5-Flash fail: the sparse keyframes capture the swimmer's body in
mid-stroke but cannot disambiguate butterfly from freestyle without observing
the \emph{temporal pattern} of arm recovery and dolphin kicks.
GPT-5.4 also fails on I-frames alone; among the models in this table,
only Gemini-3-Pro correctly identifies the stroke in this condition.
Strikingly, Gemini-2.5-Flash answers correctly from MV maps alone (D): as
visible in row~D of Figure~\ref{fig:case-swim}, the motion vectors encode the
characteristic bilateral arm sweep of butterfly stroke as a strong, symmetric
block-motion field with consistent directional patterns across all 8 frames.
Conversely, Qwen2.5-VL-32B answers correctly from residual maps alone (E):
row~E shows that the residuals highlight the distinctive splash pattern and
torso undulation that differ between butterfly and freestyle.
\textbf{HIMMEL's contribution:} Qwen2.5-VL-7B + HIMMEL answers correctly under
conditions B and C, the tri-stream mixed conditions that mirror HIMMEL's
actual inference protocol.  While the raw 7B baseline fails on all conditions,
the learned adapter enables a 7B model to match or exceed the 32B model's
per-condition accuracy by fusing motion and texture cues through the gating
mechanism.
GPT-5.4 answers correctly in conditions C and D, while Gemini-3-Pro answers
correctly in all but the residual-only condition. These responses alone do
not identify whether differences arise from the visual encoder or reasoning.
Figure~\ref{fig:tristream-166} further illustrates how the three streams
provide complementary information: I-frames give semantic context, MV maps
capture motion direction, and residuals preserve fine texture changes.
This complementarity (\emph{different models extract different motion cues
from different compressed-domain streams}) is precisely the motivation for
HIMMEL's tri-stream fusion, which provides all three signal types to the LLM
through learned gating.

\FloatBarrier
\subsection{Case~2: Basketball action recognition (Video-MME \#143)}

\begin{figure}[H]
  \centering
  \includegraphics[width=\linewidth]{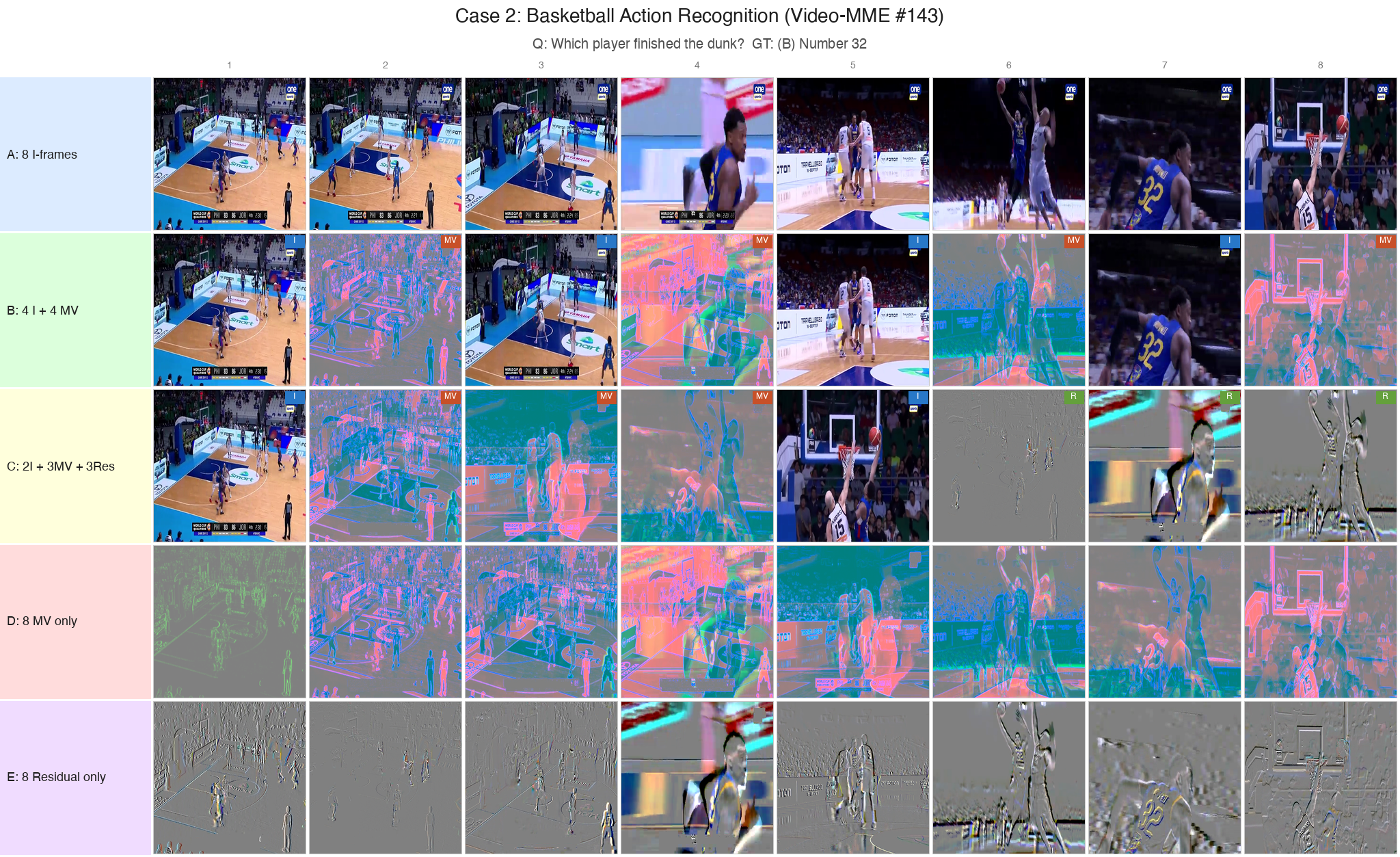}
  \caption{Five-condition ablation for Video-MME \#143 (basketball).
    In row~B, I-frames and MV maps alternate (odd columns show I-frames,
    even columns show MV maps), revealing both player identity and movement
    trajectories simultaneously.
    The pure MV maps (row~D) clearly show player movement trajectories across
    the court, while residual maps (row~E) highlight the high-frequency
    texture of fast limb motion during dribbling and shooting.
    This multi-player, fast-motion scenario is where dense temporal
    compressed-domain signals contribute most.}
  \label{fig:case-basketball}
\end{figure}

Figure~\ref{fig:case-basketball} shows the five-condition ablation for a
basketball game video.  The MV maps in rows~B and D vividly capture player
movement trajectories: the cyan and magenta displacement vectors trace out
the paths of players running across the court.  The residual maps in rows~C
and E highlight rapid limb motion during dribbling and shooting through
high-frequency edge-like patterns.  Multi-player fast-motion sports like
basketball demonstrate why dense inter-frame signals provide information
beyond what sparse I-frame sampling can capture: the tactical formations
and player interactions evolve continuously and are only visible through
motion-aware representations.

\FloatBarrier
\subsection{Case~3: Physical reasoning on Perception Test}

\begin{figure}[H]
  \centering
  \includegraphics[width=\linewidth]{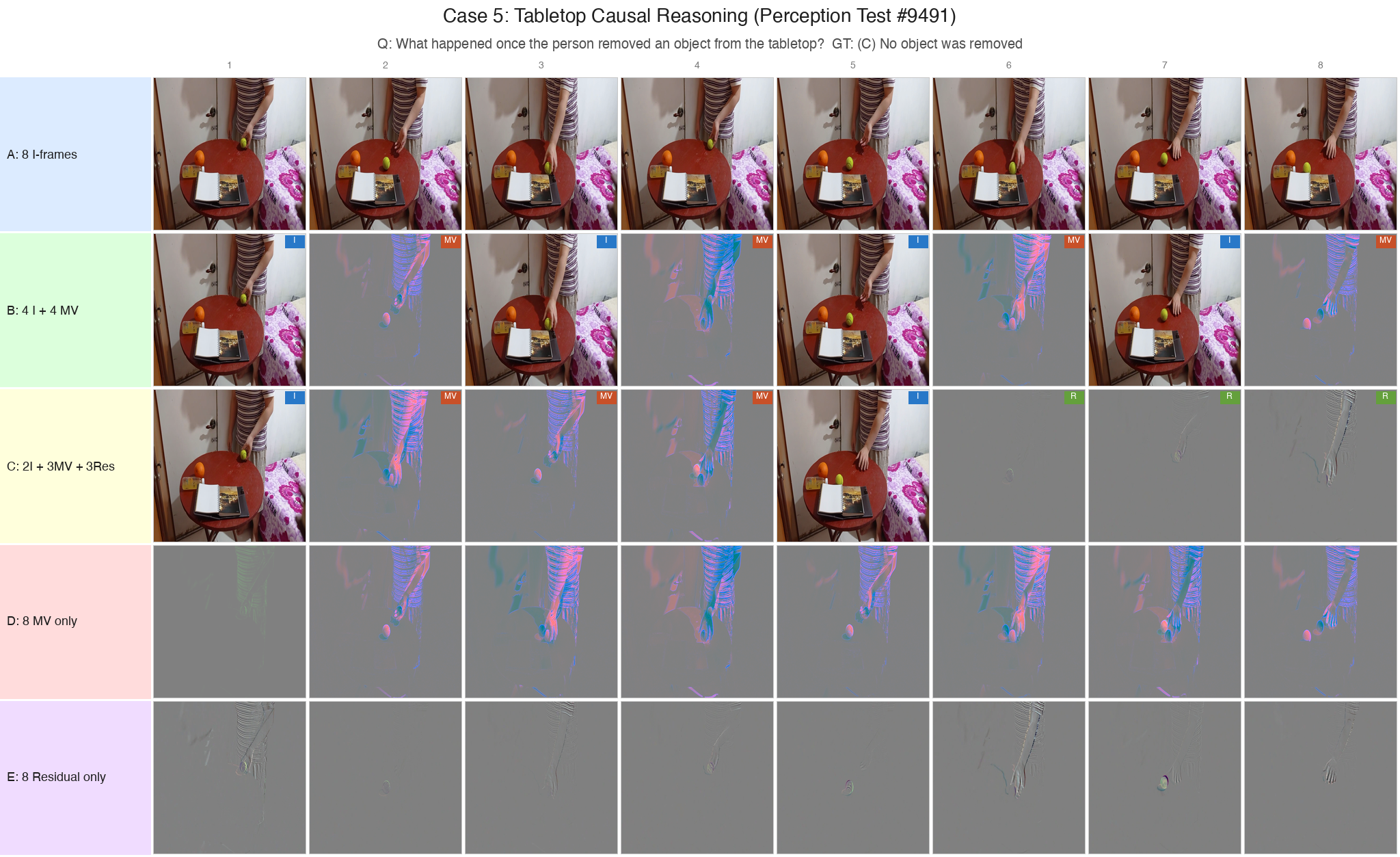}
  \caption{Perception Test \#9491: tabletop causal reasoning.
    This is the only remaining tabletop-style example in the appendix, kept as a
    \emph{failure case}.  Gemini-2.5-Flash is correct under condition~A (I-frames
    only), and both GPT-5.4 and Gemini-3-Pro also succeed on A, but
    adding raw MV or residual inputs causes all unadapted models to drift toward the
    wrong ``object fell off'' hypothesis.  The case illustrates that compressed-domain
    signals are not automatically useful without learned fusion, especially when the key
    evidence is subtle object permanence rather than large visible motion.}
  \label{fig:case-pt-tabletop}
\end{figure}

\begin{figure}[H]
  \centering
  \includegraphics[width=\linewidth]{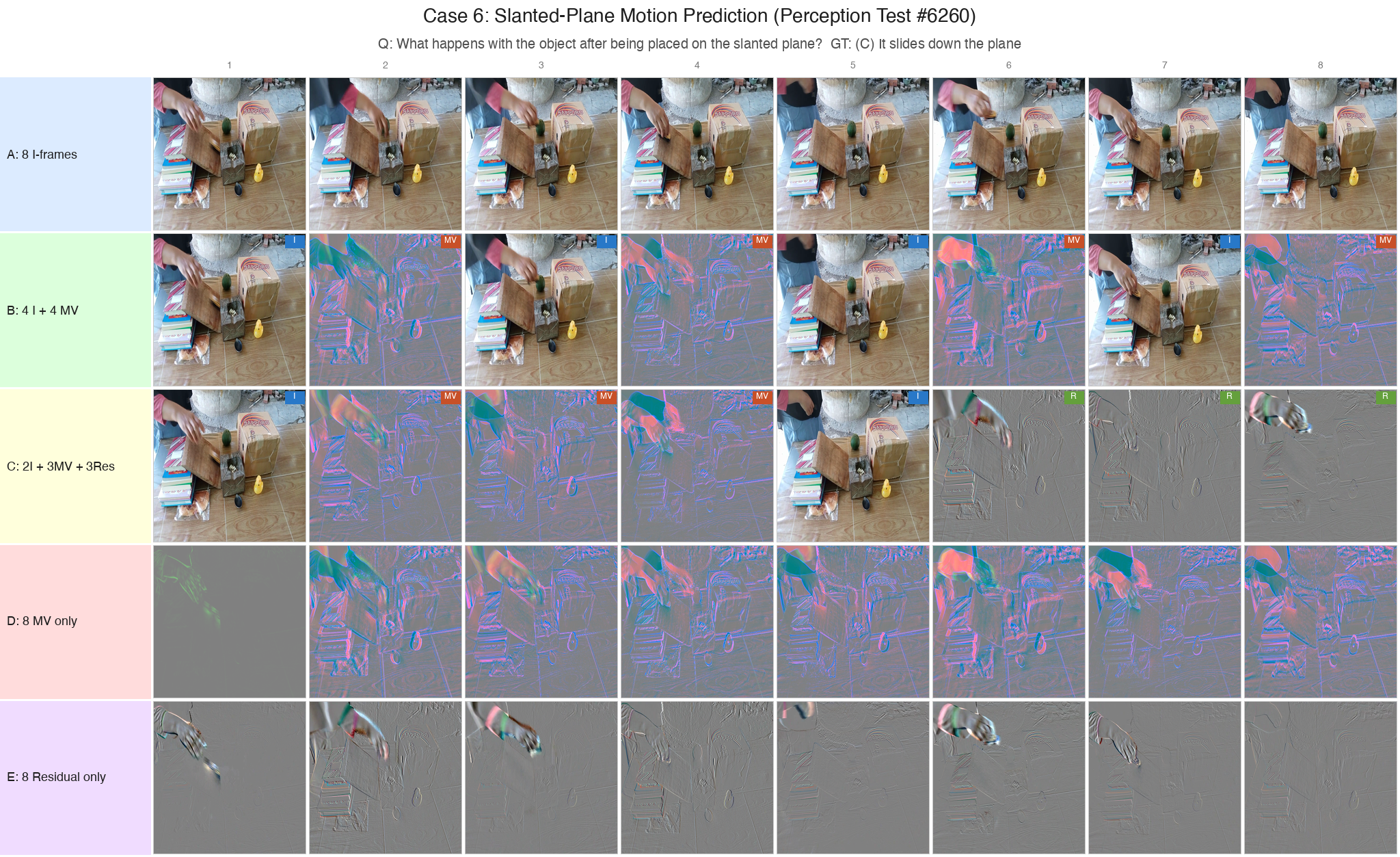}
  \caption{Perception Test \#6260: slanted-plane motion prediction.
    This is the cleanest motion-dominant success case: Qwen3-VL-235B,
    Gemini-2.5-Flash, GPT-5.4, and Gemini-3-Pro all answer correctly under
    \emph{all five} conditions.  Qwen2.5-VL-7B + HIMMEL also succeeds on
    conditions A--C, confirming that the learned adapter preserves the
    motion cues.  The reason is visible in the rows themselves: the object
    trajectory along the slanted plane is preserved not only in I-frames
    but also in the MV-only and residual-only visualizations, making it a
    natural sanity check that the codec-domain streams really do retain
    physically meaningful dynamics.}
  \label{fig:case-pt-plane}
\end{figure}

\begin{figure}[H]
  \centering
  \includegraphics[width=\linewidth]{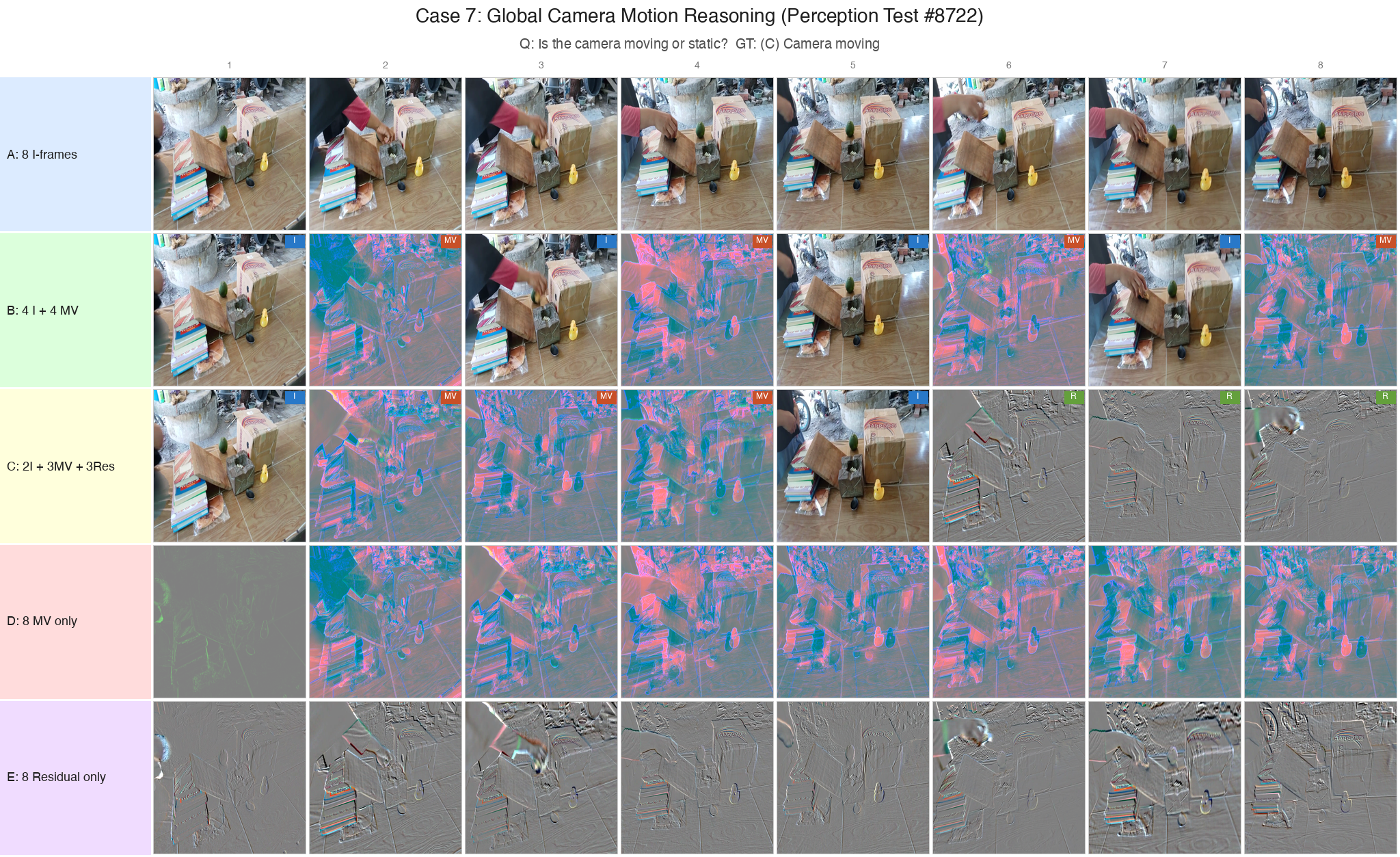}
  \caption{Perception Test \#8722: global camera-motion reasoning.
    Here unadapted models (including Qwen2.5-VL-32B and Gemini-2.5-Flash)
    fail on the appearance-heavy conditions A/B but become correct in
    the mixed and motion-rich settings C/D.  GPT-5.4 and Gemini-3-Pro
    succeed on A due to stronger visual encoders, but still show improved
    confidence under C/D.  Qwen2.5-VL-7B + HIMMEL answers correctly on C,
    matching the larger models.
    The decisive cue is the \emph{global} flow
    pattern spanning almost the whole frame: MV maps show coherent background motion that
    is hard to infer reliably from a few sparse I-frames alone.  This case is visually very
    different from the tabletop scenes and directly demonstrates why dense motion tokens help
    beyond object-centric action recognition.}
  \label{fig:case-pt-camera}
\end{figure}

\begin{figure}[H]
  \centering
  \includegraphics[width=\linewidth]{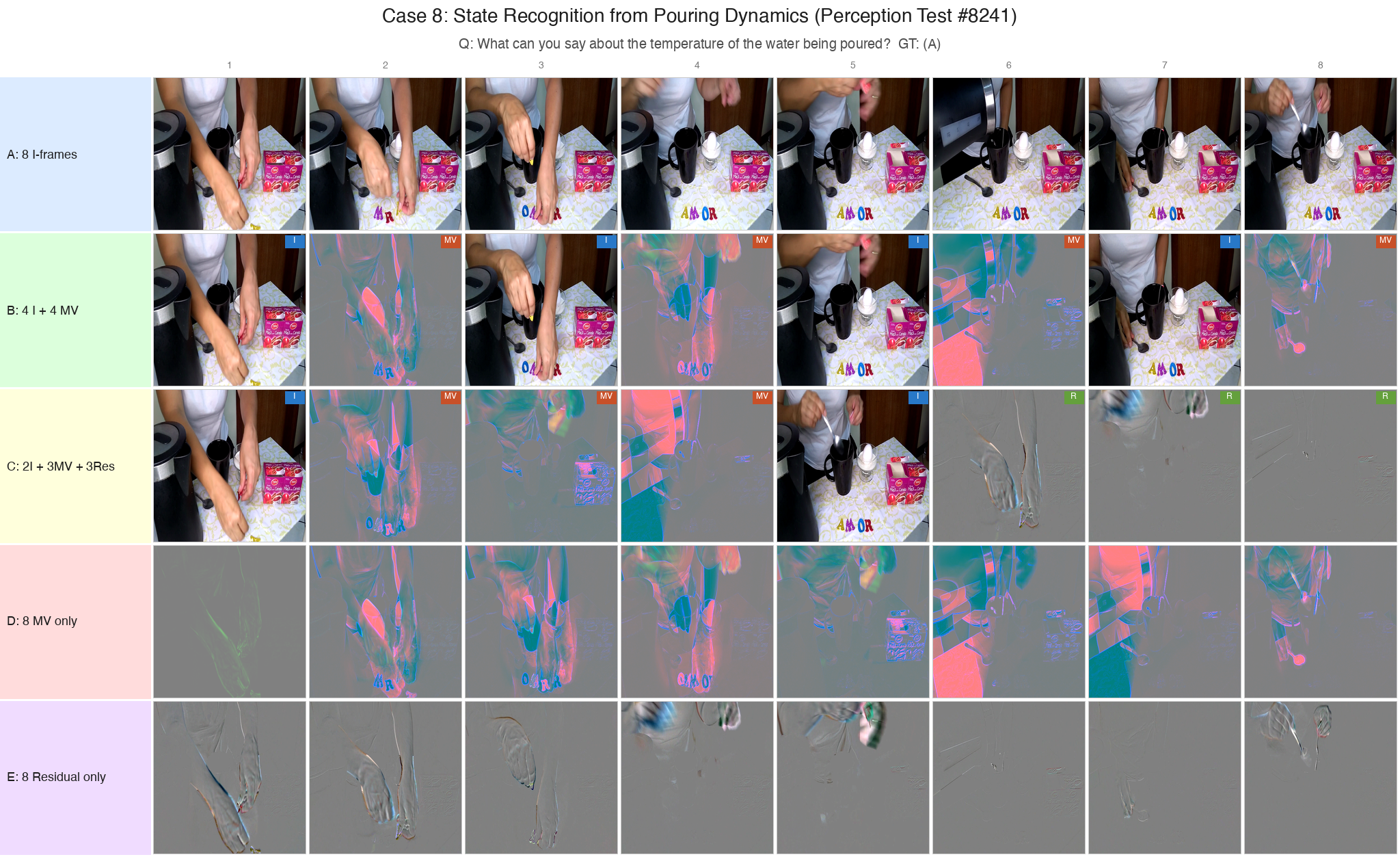}
  \caption{Perception Test \#8241: state recognition from pouring dynamics.
    The task asks about the temperature state of the poured water, which is only weakly
    indicated in the RGB snapshots.  Most models fail under A/B/C/E, while Gemini-2.5-Flash
    is correct under condition~D (MV only), and Gemini-3-Pro succeeds on both A and D.
    GPT-5.4 answers correctly under C (tri-stream mix), suggesting that the absence of
    steam-like upward motion and the localized pouring trajectory can occasionally be
    easier to judge from motion structure than from sparse appearance snapshots.
    Qwen2.5-VL-7B + HIMMEL fails on this case, illustrating that state-recognition
    tasks remain challenging even with learned fusion.
    We include this example because it broadens the appendix
    beyond pure object-motion questions into state-recognition reasoning.}
  \label{fig:case-pt-temperature}
\end{figure}

We evaluate 100 balanced questions from Perception
Test~\citep{perceptiontest} (36 physics, 28 semantics, 26 abstraction,
10 memory) under the same 5-condition protocol.
Aggregate results are in Tables~\ref{tab:5cond-pt}
and~\ref{tab:5cond-pt-himmel} (Section~\ref{app:5condition}).
Figures~\ref{fig:case-pt-tabletop}--\ref{fig:case-pt-temperature} now show four
\emph{qualitatively distinct} Perception Test tasks rather than three near-duplicate tabletop scenes:
a tabletop causal-reasoning failure case, a slanted-plane motion-prediction case,
a global camera-motion case, and a state-recognition case based on pouring dynamics.

\textbf{Analysis.}
At the 100-question scale, I-frame baselines clearly dominate for all
unadapted models, from 55\% (Qwen2.5-VL-32B) to 72\% (Gemini-3-Pro).
Adding raw MV maps in condition~B reduces accuracy by $-5$ to $-6$ pp
across all models, and pure MV/Residual conditions (D, E) drop to
$22$--$38\%$, confirming that raw compressed-domain signals are semantically
opaque without adaptation, even for frontier models like GPT-5.4 (D: 35\%)
and Gemini-3-Pro (D: 38\%).
However, HIMMEL-trained models recover this gap entirely:
Qwen3-VL-8B with HIMMEL reaches 62\%, matching Gemini-2.5-Flash's I-frame baseline,
and InternVL3-8B achieves 63\%, surpassing it.
GPT-5.4 (68\%) and Gemini-3-Pro (72\%) set higher I-frame ceilings,
but their steep degradation under D/E underscores that \emph{scale alone
cannot substitute for learned motion-signal fusion}.
For physics-category questions specifically
(Figures~\ref{fig:case-pt-tabletop}--\ref{fig:case-pt-temperature}),
the HIMMEL gain is most pronounced
($+12$ pp average), but the four examples make clear that this gain is \emph{heterogeneous}:
some cases are motion-dominant successes (\#6260, \#8722), some are fragile state-recognition
cases (\#8241), and some remain failure cases for naive raw-signal prompting (\#9491).
This is exactly the regime where HIMMEL's learned adapter matters most, because it can exploit
useful motion structure without forcing every compressed-domain cue to be interpreted literally.

\FloatBarrier
\subsection{Case~4: Failure analysis on noisy compressed signals}

\begin{figure}[H]
  \centering
  \includegraphics[width=\linewidth]{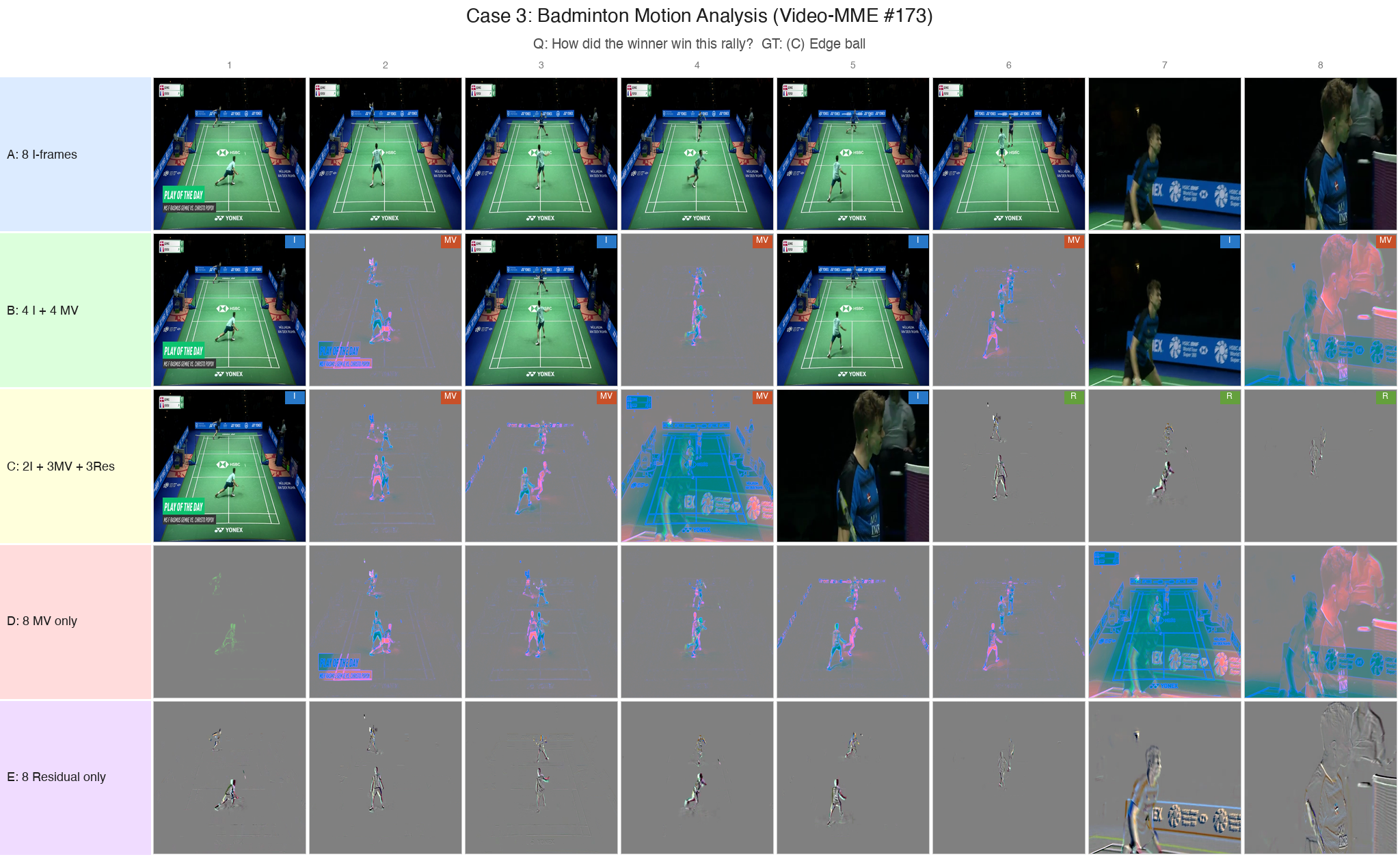}
  \caption{Five-condition ablation for Video-MME \#173 (badminton).
    Row~B interleaves I-frames (columns 1,3,5,7) with MV maps
    (columns 2,4,6,8).  The pure MV maps (row~D) show clean, high-contrast
    motion vectors that clearly delineate player silhouettes and racket
    trajectories against a largely static background.  However, the residual
    maps (row~E) are very sparse due to the uniform dark background,
    illustrating that residual informativeness is scene-dependent.
    HIMMEL's gated fusion handles this by learning to weight streams
    adaptively.}
  \label{fig:case-badminton}
\end{figure}

\begin{figure}[H]
  \centering
  \includegraphics[width=\linewidth]{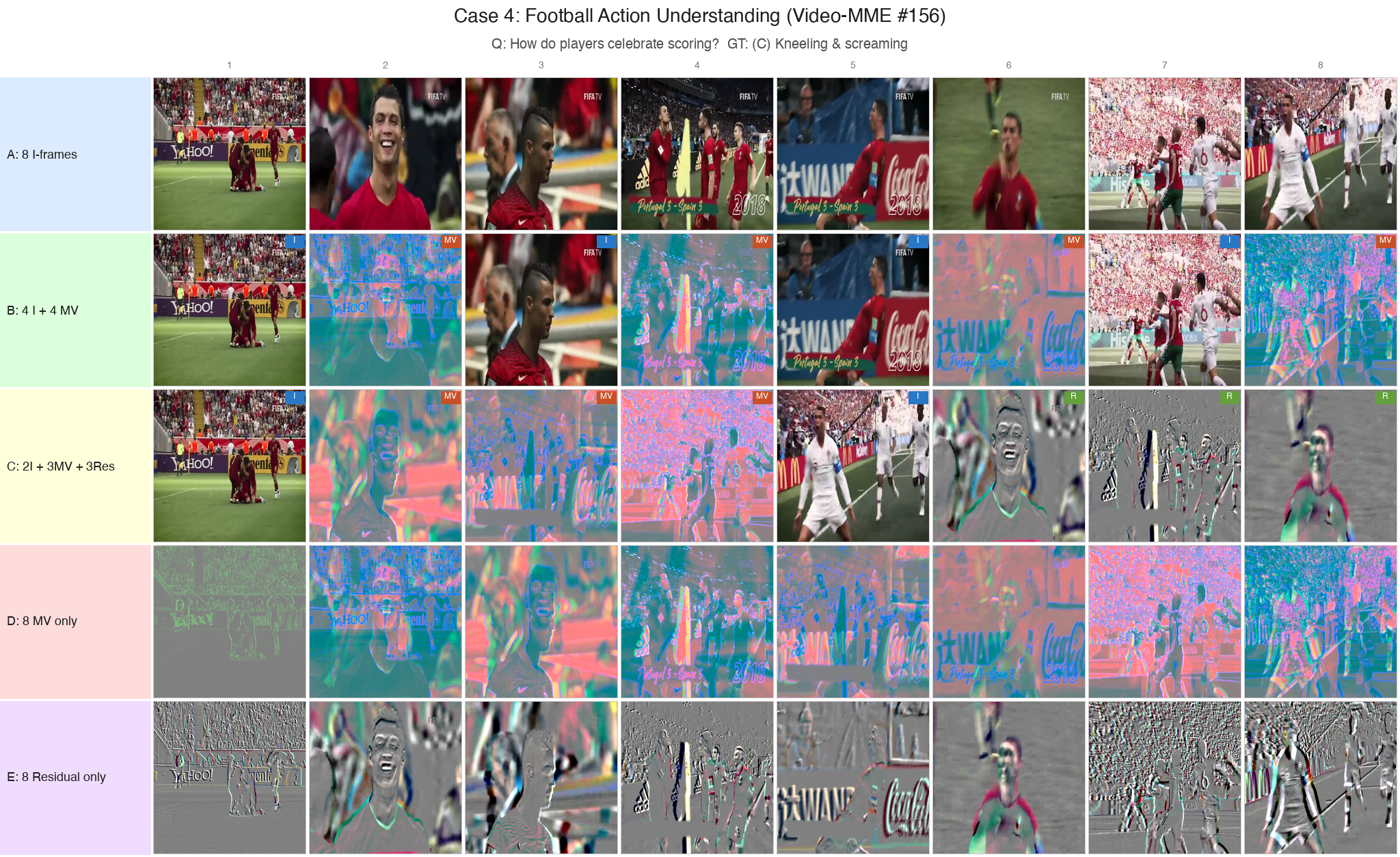}
  \caption{Five-condition ablation for Video-MME \#156 (football).
    Row~B interleaves I-frames and MV maps, while row~C places
    I-frames at columns 1 and 5, MV maps at columns 2--4, and residuals
    at columns 6--8.
    The dense crowd and rapid camera panning produce rich MV maps (rows~B, D)
    with complex motion flow patterns.  The residual maps (rows~C, E) show
    strong high-frequency content from player texture changes.  However, the
    high motion complexity also introduces noisy MV vectors in occluded
    regions, a scenario where HIMMEL's I-frame context branch provides
    crucial semantic grounding.}
  \label{fig:case-football}
\end{figure}

Figures~\ref{fig:case-badminton} and~\ref{fig:case-football} illustrate
contrasting scenarios.  In the badminton case
(Figure~\ref{fig:case-badminton}), the controlled indoor setting with a
static camera produces clean, high-contrast MV maps where the player
silhouettes are clearly delineated, but the residuals are very sparse
due to the uniform dark background.  In the football case
(Figure~\ref{fig:case-football}), the rapid camera panning and dense crowd
produce rich but noisy MV maps.

Two Perception Test failure cases illustrate the same weakness from a slightly different angle:
case \#5615 (surface-removal counterfactual) and case \#9491 (tabletop object permanence).
In both, the correct answer is obtained from I-frames only (condition~A), but adding MV or
residual maps \emph{degrades} performance.  Inspection of the MV rows reveals low-magnitude,
spatially noisy motion fields from static-camera, low-contrast scenes where the codec produces
little reliable directional evidence.
These cases motivate retaining I-frame context when compressed motion cues
are weak. HIMMEL's appearance branch and learned fusion provide a way to
combine such signals, while the controlled ablations separately support
the alignment stage (Section~\ref{sec:ablation}). The qualitative examples
alone do not establish that contrastive alignment caused a particular
gate response or corrected these individual failures.

\FloatBarrier
\section{Existing Assets and Licenses}\label{app:licenses}

The tables below distinguish model/checkpoint terms from dataset and upstream-content terms. Asset names link to the official release cards or repositories. A repository license does not override the rights or restrictions attached to third-party images and videos; the terms of the specific checkpoint and dataset version apply.

\begin{table}[H]
  \centering\small
  \caption{Model assets and the terms stated by their official releases.}
  \begin{tabular}{@{}p{0.30\linewidth}p{0.20\linewidth}p{0.42\linewidth}@{}}
    \toprule
    Asset & Reported terms & Scope \\
    \midrule
    \href{https://huggingface.co/Qwen/Qwen2.5-VL-7B-Instruct}{Qwen2.5-VL-7B-Instruct} & Apache-2.0 & License for this checkpoint; do not infer the terms of other model sizes. \\
    \href{https://huggingface.co/Qwen/Qwen3-VL-8B-Instruct}{Qwen3-VL-8B-Instruct} & Apache-2.0 & License for the evaluated 8B instruction-tuned checkpoint. \\
    \href{https://huggingface.co/OpenGVLab/InternVL3-8B}{InternVL3-8B} & MIT / Apache-2.0 & Project code is MIT; the model card identifies Apache-2.0 for its Qwen2.5 component. \\
    \href{https://huggingface.co/lmms-lab/llava-onevision-qwen2-7b-ov}{LLaVA-OV-Qwen2-7B} & Apache-2.0 & Preserve the release's license and attribution notices. \\
    \bottomrule
  \end{tabular}
\end{table}

\begin{table}[H]
  \centering\small
  \caption{Dataset releases and upstream-content restrictions. ``Evaluation only'' describes the stated release restriction where applicable, not a license inferred from our use.}
  \begin{tabular}{@{}p{0.22\linewidth}p{0.23\linewidth}p{0.47\linewidth}@{}}
    \toprule
    Asset & Reported terms & Scope and restrictions \\
    \midrule
    \href{https://github.com/MME-Benchmarks/Video-MME}{Video-MME} & Academic research only & Commercial use is prohibited; redistribution requires prior approval. Video copyright remains with the owners. \\
    \href{https://huggingface.co/datasets/OpenGVLab/MVBench}{MVBench} & MIT + source-specific & Original video terms apply; NTU RGB+D videos require separate acquisition. \\
    \href{https://huggingface.co/datasets/MLVU/MVLU}{MLVU} & CC-BY-NC-SA-4.0 & Research-only release; underlying video copyrights remain with their owners. \\
    \href{https://huggingface.co/datasets/AI4Math/MathVista}{MathVista} & CC-BY-SA-4.0 + source-specific & License covers new contributions; source-image/question terms remain. The release prohibits training use. \\
    \href{https://huggingface.co/datasets/MathLLMs/MathVision}{MathVision} & MIT & Follow the dataset release and any source-content terms. \\
    \href{https://github.com/google-deepmind/perception_test}{Perception Test} & CC-BY-4.0 / Apache-2.0 & Non-software materials are CC-BY-4.0; software is Apache-2.0. \\
    \href{https://huggingface.co/datasets/longvideobench/LongVideoBench}{LongVideoBench} & CC-BY-NC-SA-4.0 & Non-commercial/share-alike release; preserve upstream video rights. \\
    \href{https://github.com/MILVLG/activitynet-qa}{ActivityNet-QA} & Apache-2.0 repository & Underlying ActivityNet videos retain their original terms. \\
    \href{https://huggingface.co/datasets/lmms-lab/LLaVA-Video-178K}{LLaVA-Video-178K} & Apache-2.0 release card & Stage-2 supervision~\citep{llavavideo}; source videos retain their own licenses. \\
    \href{https://huggingface.co/datasets/openai/gsm8k}{GSM8K} & MIT & Keep training and held-out test splits separate. \\
    \href{https://huggingface.co/datasets/nyu-visionx/VSI-SUPER-Recall}{VSI-SUPER-Recall} / \href{https://huggingface.co/datasets/nyu-visionx/VSI-SUPER-Count}{Count} & Apache-2.0 release cards & Respect any terms attached to underlying source assets. \\
    \bottomrule
  \end{tabular}
\end{table}

\FloatBarrier
\clearpage

\section*{NeurIPS Paper Checklist}

\begin{enumerate}

\item {\bf Claims}
    \item[] Question: Do the main claims made in the abstract and introduction accurately reflect the paper's contributions and scope?
    \item[] Answer: \answerYes{}
    \item[] Justification: The abstract, introduction, and method sections consistently describe HIMMEL as a hierarchical compressed-domain motion encoding framework for long video understanding and keep the scope aligned with the current implementation-backed setting.

\item {\bf Limitations}
    \item[] Question: Does the paper discuss the limitations of the work performed by the authors?
    \item[] Answer: \answerYes{}
    \item[] Justification: The Limitations section at the end of the main paper discusses the single-video scope, modality and language coverage, and other constraints of HIMMEL; additional reproducibility-related constraints are covered in Appendix~\ref{app:repro}.

\item {\bf Theory assumptions and proofs}
    \item[] Question: For each theoretical result, does the paper provide the full set of assumptions and a complete (and correct) proof?
    \item[] Answer: \answerNA{}
    \item[] Justification: The paper does not claim formal theoretical results requiring theorem statements or proofs.

\item {\bf Experimental result reproducibility}
    \item[] Question: Does the paper fully disclose all the information needed to reproduce the main experimental results of the paper to the extent that it affects the main claims and/or conclusions of the paper (regardless of whether the code and data are provided or not)?
    \item[] Answer: \answerNo{}
    \item[] Justification: The architecture, training states, optimization settings, and evaluation controls are described. Complete checkpoint-to-run mappings, backend provenance, and exact training manifests are not provided; these are needed for fully specified reproduction.

\item {\bf Open access to data and code}
    \item[] Question: Does the paper provide open access to the data and code, with sufficient instructions to faithfully reproduce the main experimental results, as described in supplemental material?
    \item[] Answer: \answerNo{}
    \item[] Justification: Core code and documented extraction tools are public at \url{https://github.com/Sunbeam23333/HIMMEL} (revision \texttt{dcdd15c}). The release is not a complete, checkpoint-exact reproduction package: checkpoints, frozen training manifests, and complete per-run evaluation mappings are not included. Benchmark media and third-party model weights remain subject to their original access terms.

\item {\bf Experimental setting/details}
    \item[] Question: Does the paper specify all the training and test details (e.g., data splits, hyperparameters, how they were chosen, type of optimizer) necessary to understand the results?
    \item[] Answer: \answerNo{}
    \item[] Justification: The principal training settings and evaluation protocols are stated, but the exact mathematics-rehearsal dataset identities and complete per-run sequence/preprocessing configurations are not specified.

\item {\bf Experiment statistical significance}
    \item[] Question: Does the paper report error bars suitably and correctly defined or other appropriate information about the statistical significance of the experiments?
    \item[] Answer: \answerYes{}
    \item[] Justification: Appendix~\ref{app:stats} defines the benchmark-level Wilson intervals and their limitations. They are not estimates of training-seed variance, and no paired significance is inferred from marginal correct counts. Supplementary aggregate scores are distinguished from the exact-count mathematical-retention evaluation.

\item {\bf Experiments compute resources}
    \item[] Question: For each experiment, does the paper provide sufficient information on the computer resources (type of compute workers, memory, time of execution) needed to reproduce the experiments?
    \item[] Answer: \answerYes{}
    \item[] Justification: Appendix~\ref{app:settings} reports the GPU type and count, precision, optimization stack, and flagship training budget. Appendix~\ref{app:cr-efficiency} defines the hardware and measurement scope of the system comparisons. The stated training budget is not the aggregate compute for all ablations.

\item {\bf Code of ethics}
    \item[] Question: Does the research conducted in the paper conform, in every respect, with the NeurIPS Code of Ethics \url{https://neurips.cc/public/EthicsGuidelines}?
    \item[] Answer: \answerYes{}
    \item[] Justification: The work is a systems and representation-learning study on video-language modeling and does not involve human-subject experiments or intentionally harmful procedures; the submission is prepared to conform to the NeurIPS Code of Ethics.

\item {\bf Broader impacts}
    \item[] Question: Does the paper discuss both potential positive societal impacts and negative societal impacts of the work performed?
    \item[] Answer: \answerYes{}
    \item[] Justification: The Limitations section discusses potential computation savings alongside surveillance, consent, privacy, and inherited model-bias risks.

\item {\bf Safeguards}
    \item[] Question: Does the paper describe safeguards that have been put in place for responsible release of data or models that have a high risk for misuse (e.g., pre-trained language models, image generators, or scraped datasets)?
    \item[] Answer: \answerNA{}
    \item[] Justification: The submission does not currently describe the public release of a new high-risk model or dataset that would require a safeguard section.

\item {\bf Licenses for existing assets}
    \item[] Question: Are the creators or original owners of assets (e.g., code, data, models), used in the paper, properly credited and are the license and terms of use explicitly mentioned and properly respected?
    \item[] Answer: \answerYes{}
    \item[] Justification: Appendix ``Existing Assets and Licenses'' documents the reported license or access terms for the main backbone models and benchmarks, and explains the corresponding redistribution constraints that must be respected.

\item {\bf New assets}
    \item[] Question: Are new assets introduced in the paper well documented and is the documentation provided alongside the assets?
    \item[] Answer: \answerYes{}
    \item[] Justification: The released core software includes setup and usage instructions, reproduction-scope notes, codec-signal provenance documentation, and tests. The repository identifies its Apache-2.0 license and the C extractor's separate MIT license. No new public dataset or model checkpoint is released; the remaining reproduction artifacts are explicitly identified in Appendix~\ref{app:repro} and the repository documentation.

\item {\bf Crowdsourcing and research with human subjects}
    \item[] Question: For crowdsourcing experiments and research with human subjects, does the paper include the full text of instructions given to participants and screenshots, if applicable, as well as details about compensation (if any)?
    \item[] Answer: \answerNA{}
    \item[] Justification: The paper does not involve crowdsourcing or research with human subjects.

\item {\bf Institutional review board (IRB) approvals or equivalent for research with human subjects}
    \item[] Question: Does the paper describe potential risks incurred by study participants, whether such risks were disclosed to the subjects, and whether Institutional Review Board (IRB) approvals (or an equivalent approval/review based on the requirements of your country or institution) were obtained?
    \item[] Answer: \answerNA{}
    \item[] Justification: The paper does not involve crowdsourcing or research with human subjects.

\item {\bf Declaration of LLM usage}
    \item[] Question: Does the paper describe the usage of LLMs if it is an important, original, or non-standard component of the core methods in this research? Note that if the LLM is used only for writing, editing, or formatting purposes and does \emph{not} impact the core methodology, scientific rigor, or originality of the research, declaration is not required.
    \item[] Answer: \answerYes{}
    \item[] Justification: The host multimodal LMs and their adaptation are described in the method and experimental sections. LLM assistance was also used for drafting, code-related assistance, and discussing experimental plans. Proposed results are not experimental evidence, and the authors remain responsible for checking the implementation, measurements, references, and final claims.

\end{enumerate}

\end{document}